\documentclass{article}

\usepackage{PRIMEarxiv}

\usepackage[utf8]{inputenc} 
\usepackage[T1]{fontenc}    
\usepackage{hyperref}       
\usepackage{url}            
\usepackage{booktabs}       
\usepackage{amsfonts}       
\usepackage{nicefrac}       
\usepackage{microtype}      
\usepackage{lipsum}
\usepackage{fancyhdr}       
\usepackage{graphicx}       
\graphicspath{{media/}}     

\usepackage{hyperref}
\usepackage{url}

\usepackage{color}

\usepackage{algorithm}
\usepackage{algorithmic}
\usepackage{tabularx}
\usepackage{amsmath}
\usepackage{amssymb}
\usepackage{bm}
\usepackage{graphicx}
\usepackage{booktabs}  
\usepackage{threeparttable} \usepackage{multirow}
\usepackage{cite}
\usepackage{subfigure}

\newtheorem{theorem}{Theorem}
\newtheorem{proposition}{Proposition}

\newtheorem{definition}{Definition}
\newtheorem{assumption}{Assumption}

\title{Unlocking TriLevel Learning with Level-Wise Zeroth Order Constraints: Distributed Algorithms and Provable Non-Asymptotic Convergence
}

\author{
  Yang Jiao \\
  Tongji University \\
  \texttt{yangjiao@tongji.edu.cn} \\
   \And
  Kai Yang\thanks{Corresponding author.} \\
  Tongji University \\
  \texttt{kaiyang@tongji.edu.cn} \\
     \And
  Chengtao Jian \\
  Tongji University \\
  \texttt{jct@tongji.edu.cn} \\
}

\begin{document}
\maketitle

\begin{abstract}
Trilevel learning (TLL) found diverse applications in numerous machine learning applications, ranging from robust hyperparameter optimization to domain adaptation. However, existing researches primarily focus on scenarios where TLL can be addressed with first order information available at each level, which is inadequate in many situations involving zeroth order constraints, such as when black-box models are employed. Moreover, in trilevel learning, data may be distributed across various nodes, necessitating strategies to address TLL problems without centralizing data on servers to uphold data privacy. To this end, an effective distributed trilevel zeroth order learning framework DTZO is proposed in this work to address the TLL problems with level-wise zeroth order constraints in a distributed manner. The proposed DTZO is versatile and can be adapted to a wide range of (grey-box) TLL problems with partial zeroth order constraints. In DTZO, the cascaded polynomial approximation can be constructed without relying on gradients or sub-gradients, leveraging a novel cut, i.e., zeroth order cut. Furthermore, we theoretically carry out the non-asymptotic convergence rate analysis for the proposed DTZO in achieving the $\epsilon$-stationary point. Extensive experiments have been conducted to demonstrate and validate the superior performance of the proposed DTZO, e.g., it approximately achieves up to a 40$\%$ improvement in performance.
\end{abstract}

\section{Introduction}

Trilevel learning (TLL), also known as trilevel optimization, pertains to nested optimization problems involving three levels of optimization, thus exhibiting a trilevel hierarchical structure. Trilevel learning has been widely used in many machine learning applications, such as robust hyperparameter optimization \cite{sato2021gradient}, domain adaptation \cite{choe2022betty}, robust neural architecture search \cite{guo2020meets,jiao2024provably}, and so on. 
The general form of a trilevel learning problem can be expressed as,
\begin{equation}
\label{eq:5_13_1}
    \begin{array}{l}
\min \quad {{f_{1}}({\boldsymbol{x}_1},{\boldsymbol{x}_2},{\boldsymbol{x}_3})} \vspace{1mm}\\
{\rm{s}}{\rm{.t}}{\rm{.}}\;{\boldsymbol{x}_2} = \mathop {\arg \min }\limits_{{\boldsymbol{x}_2}'} {{f_{2}}({\boldsymbol{x}_1},{\boldsymbol{x}_2}',{\boldsymbol{x}_3})} \\
\quad \;\; {\rm{s}}{\rm{.t}}{\rm{.}}\;{\boldsymbol{x}_3}=\mathop {\arg \min }\limits_{{\boldsymbol{x}_3}'}  {{f_{3}}({\boldsymbol{x}_1},{\boldsymbol{x}_2}',{\boldsymbol{x}_3}')} \\
{\mathop{\rm var}} .\qquad \quad {\boldsymbol{x}_1},{\boldsymbol{x}_2},{\boldsymbol{x}_3},
\end{array}
\end{equation}
where $f_1,f_2,f_3$ denote the first, second, and third level objectives, and  ${\boldsymbol{x}_1}  \in  \mathbb{R}^{d_1},{\boldsymbol{x}_2}  \in \!\mathbb{R}^{d_2},{\boldsymbol{x}_3} \in \mathbb{R}^{d_3}$ are variables. Existing trilevel learning approaches focus on scenarios where TLL problems can be addressed with first order information available at each level. However, situations where first order information is unavailable (i.e., $\nabla f_1$, $\nabla f_2$, $\nabla f_3$ are non-available), such as when black-box models are employed, remain \textit{under-explored}. Additionally, in trilevel learning applications, data may be distributed across various nodes, necessitating strategies to address trilevel learning problems without centralizing data on servers in order to uphold data privacy \cite{jiao2024provably}.

\textbf{Complexity of Addressing TLL with Zeroth Order Constraints:} The complexity involved in solving problems characterized by hierarchical structures with three levels is \textit{significantly greater} than that of bilevel learning problems \cite{blair1992computational,avraamidou2018mixed}. It is worth mentioning that even \textit{finding a feasible solution} in TLL problem is \textbf{NP-hard} since it necessitates addressing the inner bilevel learning problem, which is NP-hard \cite{ben1990computational,sinha2017review}. Existing approaches are not applicable for addressing TLL with zeroth order constraints, as they either rely on the first order information to solve the TLL problems \cite{jiao2024provably,sato2021gradient} or focus on single-level and bilevel zeroth order learning problems \cite{fang2022communication,qiu2023zeroth}.

To this end, an effective \textbf{D}istributed \textbf{T}rilevel \textbf{Z}eroth \textbf{O}rder learning (DTZO) framework is proposed in this work. Specifically, we first introduce the cascaded zeroth order polynomial approximation for the trilevel learning problems, which consists of the inner layer and outer layer polynomial approximation. Next, how to generate the novel zeroth order cuts without using gradients or sub-gradients to gradually refine the cascaded polynomial approximation is discussed. Zeroth order cut is a type of cutting plane that does not rely on first order information during generation. Finally, the distributed zeroth order algorithm is developed to address trilevel zeroth order learning problems (i.e., TLL with level-wise zeroth order constraints) in a distributed manner. Theoretically, we demonstrate that the proposed zeroth order cuts can construct a polynomial relaxation for TLL problems, and this relaxation will be gradually tightened with zeroth order cuts added. Additionally, we also analyze the non-asymptotic convergence rate, i.e., iteration and communication complexities, for the proposed DTZO to achieve the $\epsilon$-stationary point. The contributions of this work are summarized as follows.

\textbf{1.} Different from the existing works on single-level and bilevel zeroth order learning, this work takes an initial step towards addressing trilevel zeroth order learning.  To the best of our knowledge, this is the first work to address the trilevel zeroth order learning problems.

\textbf{2.} An effective framework DTZO with novel zeroth order cuts is proposed for tackling trilevel zeroth order learning problems in a distributed manner. Different from the existing methods, the proposed DTZO is capable of constructing the cascaded zeroth order polynomial approximation without using gradients or sub-gradients. 

\textbf{3.} Extensive experiments on black-box large language models (LLMs) trilevel learning and robust hyperparameter optimization substantiate the superior performance of the proposed DTZO.

\section{Related Work}

\subsection{Distributed Zeroth Order Optimization}
Zeroth order optimization is widely-used for addressing machine learning problems where obtaining explicit gradient expressions is challenging or impractical  \cite{liu2018zeroth,chen2019zo,wang2018stochastic,chen2017zoo,heliou2021zeroth,cai2021zeroth,gao2020can,yue2023zeroth,li2022zeroth,ren2023escaping,nikolakakis2022black,tu2019autozoom,rando2024optimal}. In practical applications of zeroth order optimization, data may be distributed across different nodes. To address zeroth order optimization problems in a distributed manner, the distributed zeroth order optimization methods have recently garnered significant attention, e.g., \cite{lian2016comprehensive,tang2020distributed,fang2022communication,chen2024fine,akhavan2021distributed,sahu2018distributed,shu2023federated}. Furthermore, to tackle the bilevel zeroth order optimization problems in a distributed manner, the federated bilevel zeroth order optimization method FedRZO$_{\rm{bl}}$ \cite{qiu2023zeroth} has been proposed. However, how to address the higher-nested zeroth order optimization problems, e.g., trilevel, in a distributed manner remains under-explored. To the best of our knowledge, this is the \textbf{first work} that considers how to address the trilevel zeroth order optimization problems.

\subsection{Trilevel Learning}
Trilevel learning has found applications in various fields within machine learning. A robust neural architecture search (NAS) approach that integrates adversarial learning with NAS is introduced in \cite{guo2020meets}. The robust NAS can be viewed as a trilevel learning problem, as discussed in \cite{jiao2024provably}. A trilevel learning problem comprising two levels pretraining, fine-tuning and hyperparameter optimization, is explored in \cite{raghu2021meta}. In \cite{garg2022learning}, the trilevel learning problem involves data reweight, architecture search, and model training is investigated. In \cite{sato2021gradient}, the robust hyperparameter optimization is framed as a trilevel learning problem, and a hypergradient-based method is proposed to address such problems. In \cite{choe2022betty}, a general automatic differentiation technique is proposed, which can be applied to trilevel learning problems. Additionally, a cutting plane based distributed algorithm is proposed in \cite{jiao2024provably} for trilevel learning problems. Nevertheless, existing methods predominantly rely on first order information to solve trilevel learning problems. This is the \textbf{first framework} that can be used to solve trilevel learning problems \textit{without} relying on first order information.

\subsection{Cutting Plane Method}

Cutting plane methods are widely used in convex optimization \cite{bertsekas2015convex,franc2011cutting}, robust optimization \cite{yang2014distributed,burger2013polyhedral}, and so on. Recently, there has been notable interest in leveraging cutting plane methods to tackle distributed nested optimization problems. It is shown in \cite{jiao2022asynchronous} that the nested optimization problem can be transformed into a \textit{decomposable optimization problem} by utilizing cutting plane method, which \textit{significantly facilitates} the design of distributed algorithms for nested optimization. In \cite{jiao2022asynchronous}, the cutting plane method is employed to tackle bilevel optimization problems in a distributed manner. Similarly, \cite{chen2024robust} utilizes the cutting plane method to address distributed bilevel optimization problems within downlink multi-cell systems. Furthermore, \cite{jiao2024provably} applies the cutting plane method to solve distributed trilevel optimization problems. However, the existing cutting plane methods for nested optimization rely on the gradients or the sub-gradients to generate cutting planes, which is not available in zeroth order optimization. In this work, the proposed framework is capable of generating zeroth order cuts for nested optimization problems without using gradients or sub-gradients.


\section{Distributed Trilevel Zeroth Order Learning}

In the practical applications of trilevel zeroth order learning, data may be distributed across multiple nodes \cite{jiao2024provably}. Aggregating data on central servers may pose significant privacy risks \cite{subramanya2021centralized}. Therefore, it is crucial to develop an effective framework to address trilevel zeroth order learning problems in a distributed manner. The distributed trilevel zeroth order learning problem can be expressed as,
\begin{equation}
\label{eq:1}
    \begin{array}{l}
\min \sum\nolimits_{j = 1}^N {{f_{1,j}}({\boldsymbol{x}_1},{\boldsymbol{x}_2},{\boldsymbol{x}_3})} \\
{\rm{s}}{\rm{.t}}{\rm{.}}\;{\boldsymbol{x}_2} = \mathop {\arg \min }\limits_{{\boldsymbol{x}_2}'} \sum\nolimits_{j = 1}^N {{f_{2,j}}({\boldsymbol{x}_1},{\boldsymbol{x}_2}',{\boldsymbol{x}_3})} \\
\quad \; \; {\rm{s}}{\rm{.t}}{\rm{.}}\;{\boldsymbol{x}_3}=\mathop {\arg \min }\limits_{{\boldsymbol{x}_3}'} \sum\nolimits_{j = 1}^N {{f_{3,j}}({\boldsymbol{x}_1},{\boldsymbol{x}_2}',{\boldsymbol{x}_3}')} \\
{\mathop{\rm var}} .\qquad \qquad {\boldsymbol{x}_1},{\boldsymbol{x}_2},{\boldsymbol{x}_3},
\end{array}
\end{equation}
where ${f_{1,j}}, {f_{2,j}}, {f_{3,j}}$ respectively denote the first, second, and third level objectives in $j^{\rm{th}}$ worker, ${\boldsymbol{x}_1}\!\in\! \mathbb{R}^{d_1},{\boldsymbol{x}_2}\!\in\! \mathbb{R}^{d_2},{\boldsymbol{x}_3}\!\in\! \mathbb{R}^{d_3}$ are variables. The first order information of functions ${f_{1,j}}, {f_{2,j}}, {f_{3,j}}$, i.e., $\nabla{f_{1,j}}, \nabla{f_{2,j}}, \nabla{f_{3,j}}$, is not available in Eq. (\ref{eq:1}), corresponding to the level-wise zeroth order constraints. To facilitate the development of distributed algorithms in parameter-server architecture \cite{jiao2022asynchronous,assran2020advances}, the distributed TLL with zeroth order constraints in Eq. (\ref{eq:1}) is equivalently reformulated as a consensus trilevel zeroth order learning problem as follows.
\begin{equation}
\label{eq:2}
    \begin{array}{l}
\min \sum\nolimits_{j = 1}^N {{f_{1,j}}({\boldsymbol{x}_{1,j}},{\boldsymbol{x}_{2,j}},{\boldsymbol{x}_{3,j}})} \\
{\rm{s}}{\rm{.t}}{\rm{.}}\; {\boldsymbol{x}_{1,j}} = \boldsymbol{z}_1, \forall j =1,\cdots,N 

\\ \, \{{\boldsymbol{x}_{2,j}}\},\boldsymbol{z}_2 = \! \mathop {\arg \min }\limits_{\{{\boldsymbol{x}_{2,j}}'\},{\boldsymbol{z}_2}'} \! \sum\nolimits_{j = 1}^N {{f_{2,j}}(\boldsymbol{z}_1,{\boldsymbol{x}_{2,j}}',{\boldsymbol{x}_{3,j}})} \\

\qquad \;   \;\; \quad {\rm{s}}{\rm{.t}}{\rm{.}}\; {\boldsymbol{x}_{2,j}}' = {\boldsymbol{z}_2}', \forall j =1,\cdots,N  

\\  \qquad \quad\; \;\;\,  \{{\boldsymbol{x}_{3,j}}\}, \boldsymbol{z}_3 {\rm{ =  }}\mathop {\arg \min }\limits_{\{{\boldsymbol{x}_{3,j}}'\}, {\boldsymbol{z}_3}'} \sum\nolimits_{j = 1}^N {{f_{3,j}}(\boldsymbol{z}_1,{\boldsymbol{z}_2}',{\boldsymbol{x}_{3,j}}')} 
\\ \qquad  \qquad  \quad \;\; \;\; \qquad {\rm{s}}{\rm{.t}}{\rm{.}}\; {\boldsymbol{x}_{3,j}}' = {\boldsymbol{z}_3}', \forall j =1,\cdots,N  
\\
{\mathop{\rm var}} .\qquad \qquad \{{\boldsymbol{x}_{1,j}}\},\{{\boldsymbol{x}_{2,j}}\},\{{\boldsymbol{x}_{3,j}}\}, \boldsymbol{z}_1, \boldsymbol{z}_2, \boldsymbol{z}_3,
\end{array}
\end{equation}
where ${\boldsymbol{x}_{1,j}} \! \in \! \mathbb{R}^{d_1}, {\boldsymbol{x}_{2,j}} \!\in\! \mathbb{R}^{d_2}, {\boldsymbol{x}_{3,j}} \!\in\! \mathbb{R}^{d_3}$ denote the local variables in $j^{\rm{th}}$ worker, $\boldsymbol{z}_1 \! \in \! \mathbb{R}^{d_1}, \boldsymbol{z}_2\!\in\! \mathbb{R}^{d_2}, \boldsymbol{z}_3\!\in\! \mathbb{R}^{d_3}$ denote the consensus variables in the master, $N$ denotes the number of workers.

\textbf{Overview of the proposed framework.} In Sec. \ref{Cascaded}, the construction of cascaded zeroth order polynomial approximation for the trilevel zeroth order learning problem is proposed, which consists of the inner layer and outer layer polynomial approximation. Then, how to gradually update zeroth order cuts to refine the cascaded polynomial approximation is discussed in Sec. \ref{refine}. Finally, a distributed zeroth order algorithm is developed to effectively address the trilevel zeroth order learning problem in a distributed manner in Sec. \ref{distributed}. To improve the readability of this work, The notations used in this work and their corresponding definitions are summarized in Table \ref{tab:notation}.

\subsection{Cascaded Zeroth Order Polynomial Approximation}
\label{Cascaded}
In this section, how to construct the cascaded zeroth order polynomial approximation for trilevel zeroth order learning is introduced. The proposed cascaded zeroth order polynomial approximation consists of two key parts: 1) the inner layer polynomial approximation and 2) the outer layer polynomial approximation, which will be discussed in detail below.

\subsubsection{Inner Layer Polynomial Approximation}

In trilevel learning, the third-level optimization problem can be viewed as the constraint to the second-level optimization problem \cite{jiao2024provably,pan2024scalebio,kwon2023fully,jiang2023conditional}, it equals the constraint $\phi_{\rm{in}}( \{{\boldsymbol{x}_{3,j}}\}, \boldsymbol{z}_1, {\boldsymbol{z}_2}', \boldsymbol{z}_3)=0$, where $\phi_{\rm{in}}( \{{\boldsymbol{x}_{3,j}}\}, \boldsymbol{z}_1, {\boldsymbol{z}_2}', \boldsymbol{z}_3)=|| \left[ \begin{array}{l}
\{ {\boldsymbol{x}_{3,j}}\} \\
{\boldsymbol{z}_3}
\end{array} \right] - \mathop {\arg \min }\limits_{\{ {\boldsymbol{x}_{3,j}}'\} ,{\boldsymbol{z}_3}'} \sum\nolimits_j {{f_{3,j}}({\boldsymbol{z}_1},{\boldsymbol{z}_2}',{\boldsymbol{x}_{3,j}}')} \,{\rm{s.t.}}\,{\boldsymbol{x}_{3,j}}' \!=\! {\boldsymbol{z}_3}', \forall j||^2$. In many bilevel and trilevel machine learning applications, e.g., neural architecture search in \cite{liu2018darts}, robust hyperparameter optimization in \cite{jiao2024provably}, the lower-level optimization problem serves as a \textbf{soft constraint} \cite{kautz1996general} to the upper-level optimization problem, i.e., this constraint (constraint $\phi_{\rm{in}}( \{{\boldsymbol{x}_{3,j}}\}, \boldsymbol{z}_1, {\boldsymbol{z}_2}', \boldsymbol{z}_3)=0$ in our problem) can be violated to a certain extent while still yielding a feasible and meaningful solution, more discussions are provided in Appendix \ref{appedix:phi}. Inspired by \cite{jiao2022asynchronous,chen2024robust}, the cutting plane based method is utilized to construct a \textit{decomposable} polynomial relaxation for this constraint, which \textit{significantly facilitates} the development of distributed algorithms. Specifically, the inner layer zeroth order cuts are utilized to approximate the feasible region with respect to constraint $\phi_{\rm{in}}( \{{\boldsymbol{x}_{3,j}}\}, \boldsymbol{z}_1, {\boldsymbol{z}_2}', \boldsymbol{z}_3)=0$. Zeroth order cuts refer to the cutting planes that do not rely on first order information during generation. In this section, we focus on the construction of cascaded polynomial approximation, and how to generate the zeroth order cuts is discussed in detail in the next section \ref{refine}. Consequently, the feasible region formed by inner layer zeroth order cuts in $t^{\rm{th}}$ iteration can be expressed as,
\begin{equation}
\label{eq:5_15_4}
   \! P_{\rm{in}}^{t} \!=\! \left\{ \!\sum\nolimits_{j}\!{{{\boldsymbol{a}_{j,l}^{\rm{in}}}^{\top}\boldsymbol{x}_{3,j}^2} \!+\! {{\boldsymbol{b}_{j,l}^{\rm{in}}}^{\top}\boldsymbol{x}_{3,j}}}  \!+\!  \sum\nolimits_{i\in \! \{1,3\}} \!{{\boldsymbol{c}_{i,l}^{\rm{in}}}^{\top}\boldsymbol{z}_{i}^2}    \!+\!{{\boldsymbol{d}_{i,l}^{\rm{in}}}^{\top}\boldsymbol{z}_{i}} \!+\! {{\boldsymbol{c}_{2,l}^{\rm{in}}}^{\top}{\boldsymbol{z}_{2}^2}'} \!+\! {{\boldsymbol{d}_{2,l}^{\rm{in}}}^{\top}{\boldsymbol{z}_{2}}'} \!+\! e_{l}^{\rm{in}}\! \le\! \varepsilon_{\rm{in}},\! \forall l \!   \right\}\!,
\end{equation}
where ${\boldsymbol{x}_{i,j}^2} = [x_{i,j,1}^2, \cdots, x_{i,j,d_i}^2] \! \in \! \mathbb{R}^{d_i},{\boldsymbol{z}_{i}^2} = [z_{i,1}^2, \cdots, z_{i,d_i}^2] \! \in \! \mathbb{R}^{d_i},i\!=\!1,2,3$,  ${\boldsymbol{a}_{j,l}^{\rm{in}}} \! \in \! \mathbb{R}^{d_3}$, ${\boldsymbol{b}_{j,l}^{\rm{in}}} \! \in \! \mathbb{R}^{d_3}$, $\boldsymbol{c}_{i,l}^{\rm{in}} \! \in \! \mathbb{R}^{d_i}$, $\boldsymbol{d}_{i,l}^{\rm{in}} \! \in \! \mathbb{R}^{d_i}$, and $e_{l}^{\rm{in}}\! \in \! \mathbb{R}^{1}$ are the parameters of $l^{\rm{th}}$ inner layer zeroth order cut, $\varepsilon_{\rm{in}}\ge 0$ is a constant. By using the inner layer polynomial approximation according to Eq. (\ref{eq:5_15_4}), the resulting problem can be written as,
\begin{equation}
\label{eq:3}
    \begin{array}{l}
\min \sum\nolimits_{j=1}^{N} {{f_{1,j}}({\boldsymbol{x}_{1,j}},{\boldsymbol{x}_{2,j}},{\boldsymbol{x}_{3,j}})} \\
{\rm{s}}{\rm{.t}}{\rm{.}}\; {\boldsymbol{x}_{1,j}} = \boldsymbol{z}_1, \forall j =1,\cdots,N 

\\ \quad  \{{\boldsymbol{x}_{2,j}}\},\boldsymbol{z}_2 = \mathop {\arg \min }\limits_{\{{\boldsymbol{x}_{2,j}}'\},{\boldsymbol{z}_2}'} \sum\nolimits_{j=1}^{N} {{f_{2,j}}(\boldsymbol{z}_1,{\boldsymbol{x}_{2,j}}',{\boldsymbol{x}_{3,j}})} \\

\qquad \qquad \qquad   \; {\rm{s}}{\rm{.t}}{\rm{.}}\; {\boldsymbol{x}_{2,j}}' = {\boldsymbol{z}_2}', \forall j =1,\cdots,N   \vspace{0.5mm}

\\ \qquad \qquad \qquad \quad \; \;  ( \{{\boldsymbol{x}_{3,j}}\}, \boldsymbol{z}_1, {\boldsymbol{z}_2}', \boldsymbol{z}_3) \in P_{\rm{in}}^{t} \vspace{0.5mm}
\\
{\mathop{\rm var}} .\qquad \qquad \{{\boldsymbol{x}_{1,j}}\},\{{\boldsymbol{x}_{2,j}}\},\{{\boldsymbol{x}_{3,j}}\}, \boldsymbol{z}_1, \boldsymbol{z}_2, \boldsymbol{z}_3.
\end{array}
\end{equation}

\subsubsection{Outer Layer Polynomial Approximation}

Likewise, the lower-level optimization problem in Eq. (\ref{eq:3}) can be regarded as the constraint to the upper-level optimization problem. Defining $h_l^{\rm{in}}( \{{\boldsymbol{x}_{3,j}}\}, \boldsymbol{z}_1, {\boldsymbol{z}_2}', \boldsymbol{z}_3) = \sum\nolimits_{j}\!{{{\boldsymbol{a}_{j,l}^{\rm{in}}}^{\top}\boldsymbol{x}_{3,j}^2} \!+\! {{\boldsymbol{b}_{j,l}^{\rm{in}}}^{\top}\boldsymbol{x}_{3,j}}}  \!+\!  \sum\nolimits_{i\in  \{1,3\}} \!{{\boldsymbol{c}_{i,l}^{\rm{in}}}^{\top}\boldsymbol{z}_{i}^2}    \!+\!{{\boldsymbol{d}_{i,l}^{\rm{in}}}^{\top}\boldsymbol{z}_{i}} \!+\! {{\boldsymbol{c}_{2,l}^{\rm{in}}}^{\top}{\boldsymbol{z}_{2}^2}'} \!+\! {{\boldsymbol{d}_{2,l}^{\rm{in}}}^{\top}{\boldsymbol{z}_{2}}'} \!+\! e_{l}^{\rm{in}}$. This constraint equals $\phi_{\rm{out}}( \{{\boldsymbol{x}_{2,j}}\}, \{{\boldsymbol{x}_{3,j}}\}, \boldsymbol{z}_1, \boldsymbol{z}_2, \boldsymbol{z}_3)=0$, where
\begin{equation}
\begin{array}{l}
\phi_{\rm{out}}( \{{\boldsymbol{x}_{2,j}}\}, \{{\boldsymbol{x}_{3,j}}\}, \boldsymbol{z}_1, \boldsymbol{z}_2, \boldsymbol{z}_3)
\\=|| \left[ \begin{array}{l}
\{ {\boldsymbol{x}_{2,j}}\} \\
{\boldsymbol{z}_2}  
\end{array} \right] - \begin{array}{l}
\mathop {\arg \min }\limits_{\{{\boldsymbol{x}_{2,j}}'\},{\boldsymbol{z}_2}'} \sum\nolimits_{j = 1}^N {{f_{2,j}}(\boldsymbol{z}_1,{\boldsymbol{x}_{2,j}}',{\boldsymbol{x}_{3,j}})} \\
{\rm{s.t.}}\,{\boldsymbol{x}_{2,j}}' \!=\! {\boldsymbol{z}_2}',\forall j,h_l^{\rm{in}}( \{{\boldsymbol{x}_{3,j}}\}, \boldsymbol{z}_1, {\boldsymbol{z}_2}', \boldsymbol{z}_3) \! \le\! \varepsilon_{\rm{in}},\forall l
\end{array}  ||^2 .
\end{array}
\end{equation}
The constraint $\phi_{\rm{out}}( \{{\boldsymbol{x}_{2,j}}\}, \{{\boldsymbol{x}_{3,j}}\}, \boldsymbol{z}_1, \boldsymbol{z}_2, \boldsymbol{z}_3)\!=\!0$ also serves as a \textit{soft constraint} to the upper-level optimization problem, more discussions about the soft constraint are provided in Appendix \ref{appedix:phi}. Outer layer zeroth order cuts are utilized to construct the polynomial approximation for the feasible region with respect to the constraint $\phi_{\rm{out}}( \{{\boldsymbol{x}_{2,j}}\}, \{{\boldsymbol{x}_{3,j}}\}, \boldsymbol{z}_1, \boldsymbol{z}_2, \boldsymbol{z}_3) =0$, that is,
\begin{equation}
\label{eq:5_15_7}
        P_{\rm{out}}^{t}  
        \!=\! \left\{\{{\boldsymbol{x}_{2,j}}\},\! \{{\boldsymbol{x}_{3,j}}\},\! \boldsymbol{z}_1,\boldsymbol{z}_2,\boldsymbol{z}_3 |\;  h_l^{\rm{out}}( \{{\boldsymbol{x}_{2,j}}\},\! \{{\boldsymbol{x}_{3,j}}\},\! \boldsymbol{z}_1, \boldsymbol{z}_2, \boldsymbol{z}_3) \!\le\! \varepsilon_{\rm{out}}, \forall l  \right\},
\end{equation}
where $h_l^{\rm{out}}( \{{\boldsymbol{x}_{2,j}}\},\! \{{\boldsymbol{x}_{3,j}}\},\! \boldsymbol{z}_1, \boldsymbol{z}_2, \boldsymbol{z}_3)\!=\! \sum\nolimits_{i=2}^{3}  \sum\nolimits_{j=1}^{N}\!{{{\boldsymbol{a}_{i,j,l}^{\rm{out}}}^{\top}\boldsymbol{x}_{i,j}^2} + {{\boldsymbol{b}_{i,j,l}^{\rm{out}}}^{\top}\boldsymbol{x}_{i,j}}}  +  \sum\nolimits_{i=1}^{3} \!{{\boldsymbol{c}_{i,l}^{\rm{out}}}^{\top}\boldsymbol{z}_{i}^2} + {{\boldsymbol{d}_{i,l}^{\rm{out}}}^{\top}\boldsymbol{z}_{i}} + e_{l}^{\rm{out}}$, and $\varepsilon_{\rm{out}}\ge 0$ is a pre-set constant.  Based on Eq. (\ref{eq:5_15_7}), the resulting cascaded zeroth order polynomial approximation problem can be written as,
\begin{equation}
\label{eq:4}
    \begin{array}{l}
\min \sum\limits_{j = 1}^N {{f_{1,j}}({\boldsymbol{x}_{1,j}},{\boldsymbol{x}_{2,j}},{\boldsymbol{x}_{3,j}})} \\

{\rm{s}}{\rm{.t}}{\rm{.}}\; {\boldsymbol{x}_{1,j}} = \boldsymbol{z}_1, \forall j =1,\cdots,N 

\\ \sum\limits_{i=2}^{3} \! \sum\limits_{j=1}^{N}\!{{{\boldsymbol{a}_{i,j,l}^{\rm{out}}}^{\top}\boldsymbol{x}_{i,j}^2} \!+\! {{\boldsymbol{b}_{i,j,l}^{\rm{out}}}^{\top}\boldsymbol{x}_{i,j}}}  \!+\!  \sum\limits_{i=1}^{3} \!{{\boldsymbol{c}_{i,l}^{\rm{out}}}^{\top}\boldsymbol{z}_{i}^2} \!+\! {{\boldsymbol{d}_{i,l}^{\rm{out}}}^{\top}\boldsymbol{z}_{i}} \!+\! e_{l}^{\rm{out}} \!\le\! \varepsilon_{\rm{out}}, \forall l 

\\ {\mathop{\rm var}} .\qquad  \{{\boldsymbol{x}_{1,j}}\},\{{\boldsymbol{x}_{2,j}}\},\{{\boldsymbol{x}_{3,j}}\}, \boldsymbol{z}_1, \boldsymbol{z}_2, \boldsymbol{z}_3,

\end{array}
\end{equation}
where ${\boldsymbol{a}_{i,j,l}^{\rm{out}}} \! \in \! \mathbb{R}^{d_i}$, ${\boldsymbol{b}_{i,j,l}^{\rm{out}}} \! \in \! \mathbb{R}^{d_i}$, $\boldsymbol{c}_{i,l}^{\rm{out}} \! \in \! \mathbb{R}^{d_i}$, $\boldsymbol{d}_{i,l}^{\rm{out}} \! \in \! \mathbb{R}^{d_i}$, and $e_{l}^{\rm{out}}\! \in \! \mathbb{R}^{1}$ are the parameters of $l^{\rm{th}}$ outer layer zeroth order cut.

\subsection{Refining the Cascaded Polynomial Approximation}
\label{refine}

For every $\mathcal{T}$ iteration, the zeroth order cuts will be updated to refine the proposed cascaded polynomial approximation when $t <T_1$. Different from the existing cutting plane methods for nested optimization, the proposed zeroth order cuts can be generated without using gradients or sub-gradients, which is why we refer to them as zeroth order cuts. Specifically, in $t^{\rm{th}}$ iteration, the zeroth order cuts will be updated by three key steps: 1) generating inner layer zeroth order cut; 2) generating outer layer zeroth order cut; 3) removing inactive zeroth order cuts, which will be discussed as follows. In addition, we demonstrate the proposed zeroth order cuts can construct a relaxation for the original feasible regions in Proposition \ref{prop:1} and \ref{prop:2}.

\subsubsection{Generating Inner Layer Zeroth Order Cut}

At $t^{\rm{th}}$ iteration, based on point $( \{{\boldsymbol{x}_{3,j}^t}\}, \boldsymbol{z}_1^t, \boldsymbol{z}_2^t, \boldsymbol{z}_3^t)$, the new inner layer zeroth order cut will be generated to refine the inner layer polynomial approximation, i.e., Eq. (\ref{eq:5_15_4}), as follows.
\begin{equation}
\label{eq:4_25_17}
\begin{array}{l}
    \phi_{\rm{in}}( \{{\boldsymbol{x}_{3,j}^t}\},\! \boldsymbol{z}_1^t, {\boldsymbol{z}_2^t}', \boldsymbol{z}_3^t)  +  {G_\mu^{\rm{in}} }{( \{{\boldsymbol{x}_{3,j}^t}\},\! \boldsymbol{z}_1^t, {\boldsymbol{z}_2^t}', \boldsymbol{z}_3^t)^{\top}}\left( {\left[ \begin{array}{l}
\{{\boldsymbol{x}_{3,j}}\}\\
\boldsymbol{z}_1\\
{\boldsymbol{z}_2}'\\
\boldsymbol{z}_3
\end{array} \right] - \left[ \begin{array}{l}
\{{\boldsymbol{x}_{3,j}^t}\}\\
\boldsymbol{z}_1^t\\
{\boldsymbol{z}_2^t}'\\
\boldsymbol{z}_3^t
\end{array} \right]} \right) \vspace{1mm} \\

\le \frac{{L + 1}}{2}  \left( \sum_{j}\! || {\boldsymbol{x}_{3,j}} \!-\! {\boldsymbol{x}_{3,j}^t}||^2 \!+\!  ||\boldsymbol{z}_1\!-\!\boldsymbol{z}_1^t||^2 \!+\! ||{\boldsymbol{z}_2}'\!-\!{\boldsymbol{z}_2^t}'||^2 \!+\!  ||\boldsymbol{z}_3\!-\!\boldsymbol{z}_3^t||^2 \right) \!+\! \frac{\mu^2 }{8}L^2 d_{\rm{in}} \!+\! \varepsilon_{\rm{in}},
\end{array}
\end{equation}
where $d_{\rm{in}} = {({d_1}\!+\!  {d_2} \!+\! (N\!+\!1){d_3} \!+\! 3)^{3}}$ and
\begin{equation}
\label{eq:5_6_9}
\begin{array}{l}
     {G_\mu^{\rm{in}} }( \{{\boldsymbol{x}_{3,j}^t}\}, \boldsymbol{z}_1^t, {\boldsymbol{z}_2^t}', \boldsymbol{z}_3^t) = \frac{\phi_{\rm{in}}( \{{\boldsymbol{x}_{3,j}^t} + \mu \boldsymbol{\mu}_{x_{3,j}}\}, \boldsymbol{z}_1^t + \mu \boldsymbol{\mu}_{z_1}, {\boldsymbol{z}_2^t}' + \mu \boldsymbol{\mu}_{z_2}, \boldsymbol{z}_3^t + \mu \boldsymbol{\mu}_{z_3})   -   \phi_{\rm{in}}( \{{\boldsymbol{x}_{3,j}^t}\}, \boldsymbol{z}_1^t, {\boldsymbol{z}_2^t}', \boldsymbol{z}_3^t)}{\mu}  \boldsymbol{\mu}^{\rm{in}},
\end{array}
\end{equation}
where $\boldsymbol{\mu}^{\rm{in}}=[\{\boldsymbol{\mu}_{x_{3,j}}\}, \boldsymbol{\mu}_{z_1}, \boldsymbol{\mu}_{z_2}, \boldsymbol{\mu}_{z_3}]$ is a standard Gaussian random vector, $L>0$ is a constant, and $\mu>0$ is the smoothing parameter \cite{kornowski2024algorithm,ghadimi2013stochastic}. Then, the new generated zeroth order cut $cp_{\rm{in}}^{\rm{new}}$ will be added into $P_{\rm{in}}^{t}$, i.e., $P_{\rm{in}}^{t} = {\rm{Add}}(P_{\rm{in}}^{t-1}, cp_{\rm{in}}^{\rm{new}})$.

\begin{proposition}
\label{prop:1}
    The original feasible region of constraint $\phi_{\rm{in}}( \{{\boldsymbol{x}_{3,j}}\}, \boldsymbol{z}_1, {\boldsymbol{z}_2}', \boldsymbol{z}_3) = 0$ is a subset of the feasible region formed by inner layer zeroth order cuts, i.e., $P_{\rm{in}}^{t+1} = \left\{ h_l^{\rm{in}}( \{{\boldsymbol{x}_{3,j}}\}, \boldsymbol{z}_1, {\boldsymbol{z}_2}', \boldsymbol{z}_3) \le \varepsilon_{\rm{in}}, \forall l   \right\}$ when $\phi_{\rm{in}}$ has $L$-Lipschitz continuous gradient. The proof is provided in Appendix \ref{appendix:pro1_2}.
\end{proposition}

\subsubsection{Generating Outer Layer Zeroth Order Cut}

At $t^{\rm{th}}$ iteration, according to point $(\{{\boldsymbol{x}_{2,j}^t}\}, \{{\boldsymbol{x}_{3,j}^t}\}, \boldsymbol{z}_1^t, \boldsymbol{z}_2^t, \boldsymbol{z}_3^t)$, the new outer layer zeroth order cut will be generated to refine the outer layer polynomial approximation in Eq. (\ref{eq:5_15_7}) as follows.
\begin{equation}
\label{eq:4_25_25}
\begin{array}{l}
    \phi_{\rm{out}}(\{{\boldsymbol{x}_{2,j}^t}\},\! \{{\boldsymbol{x}_{3,j}^t}\},\! \boldsymbol{z}_1^t, \boldsymbol{z}_2^t, \boldsymbol{z}_3^t) \! +\!  {G_\mu^{\rm{out}} }{(\{{\boldsymbol{x}_{2,j}^t}\},\!\{{\boldsymbol{x}_{3,j}^t}\},\! \boldsymbol{z}_1^t, \boldsymbol{z}_2^t, \boldsymbol{z}_3^t)^{\top}}\left( {\left[ \begin{array}{l}
\!\{{\boldsymbol{x}_{2,j}}\}\\
\!\{{\boldsymbol{x}_{3,j}}\}\\
\boldsymbol{z}_1\\
\boldsymbol{z}_2\\
\boldsymbol{z}_3
\end{array} \right] \!-\! \left[ \begin{array}{l}
\!\{{\boldsymbol{x}_{2,j}^t}\}\\
\!\{{\boldsymbol{x}_{3,j}^t}\}\\
\boldsymbol{z}_1^t\\
\boldsymbol{z}_2^t\\
\boldsymbol{z}_3^t
\end{array} \right]} \right) \vspace{1mm} \\

\le \frac{{L + 1}}{2}  \left( \sum_{i=2}^3\sum_{j}\! || {\boldsymbol{x}_{i,j}} \!-\! {\boldsymbol{x}_{i,j}^t}||^2 \!+\! \sum_{i}\! ||\boldsymbol{z}_i\!-\!\boldsymbol{z}_i^t||^2  \right) \!+\! \frac{\mu^2 }{8}L^2{({d_1} \!+\! (N\!+\!1)({d_2} \!+\! {d_3}) \!+\! 3)^{3}} \!+\! \varepsilon_{\rm{out}}.
\end{array}
\end{equation}

In Eq. (\ref{eq:4_25_25}), we have that,
\begin{equation}
\label{eq:5_6_11}
\begin{array}{l}
    {G_\mu^{\rm{out}} }(\{{\boldsymbol{x}_{2,j}^t}\},\!\{{\boldsymbol{x}_{3,j}^t}\},\! \boldsymbol{z}_1^t, \boldsymbol{z}_2^t, \boldsymbol{z}_3^t) \\
    
    = \frac{\phi_{\rm{out}}(\{{\boldsymbol{x}_{2,j}^t} + \mu \boldsymbol{\mu}_{x_{2,j}}\}, \{{\boldsymbol{x}_{3,j}^t} + \mu \boldsymbol{\mu}_{x_{3,j}}\}, \boldsymbol{z}_1^t + \mu \boldsymbol{\mu}_{z_1}, \boldsymbol{z}_2^t + \mu \boldsymbol{\mu}_{z_2}, \boldsymbol{z}_3^t + \mu \boldsymbol{\mu}_{z_3})   -   \phi_{\rm{out}}(\{{\boldsymbol{x}_{2,j}^t}\}, \{{\boldsymbol{x}_{3,j}^t}\}, \boldsymbol{z}_1^t, \boldsymbol{z}_2^t, \boldsymbol{z}_3^t)}{\mu}  \boldsymbol{\mu}^{\rm{out}},
\end{array}
\end{equation}
where $\boldsymbol{\mu}^{\rm{out}}=[\{\boldsymbol{\mu}_{x_{2,j}}\},\{\boldsymbol{\mu}_{x_{3,j}}\}, \boldsymbol{\mu}_{z_1}, \boldsymbol{\mu}_{z_2}, \boldsymbol{\mu}_{z_3}]$ is a standard Gaussian random vector. Subsequently, the new generated outer layer zeroth order cut $cp_{\rm{out}}^{\rm{new}}$ will be added into $P_{\rm{out}}^{t}$, i.e., $P_{\rm{out}}^{t} = {\rm{Add}}(P_{\rm{out}}^{t-1}, cp_{\rm{out}}^{\rm{new}})$.

\begin{proposition}
\label{prop:2}
    The original feasible region of constraint $\phi_{\rm{out}}( \{{\boldsymbol{x}_{2,j}}\}, \{{\boldsymbol{x}_{3,j}}\}, \boldsymbol{z}_1, \boldsymbol{z}_2, \boldsymbol{z}_3)=0$ is a subset of the feasible region formed by outer layer zeroth order cuts, i.e., $ P_{\rm{out}}^{t+1} 
        \!=\! \left\{\!\{{\boldsymbol{x}_{2,j}}\},\! \{{\boldsymbol{x}_{3,j}}\},\! \boldsymbol{z}_1,\!\boldsymbol{z}_2,\!\boldsymbol{z}_3|  \sum\limits_{i=2}^{3} \! \sum\limits_{j=1}^{N}\!{{{\boldsymbol{a}_{i,j,l}^{\rm{out}}}^{\top}\boldsymbol{x}_{i,j}^2} \!+\! {{\boldsymbol{b}_{i,j,l}^{\rm{out}}}^{\top}\boldsymbol{x}_{i,j}}}  \!+\!  \sum\limits_{i=1}^{3} \!{{\boldsymbol{c}_{i,l}^{\rm{out}}}^{\top}\boldsymbol{z}_{i}^2} \!+\! {{\boldsymbol{d}_{i,l}^{\rm{out}}}^{\top}\boldsymbol{z}_{i}} \!+\! e_{l}^{\rm{out}} \!\le\! \varepsilon_{\rm{out}}, \forall l  \right\}$ when $\phi_{\rm{out}}$ has $L$-Lipschitz continuous gradient. Proofs are provided in Appendix \ref{appendix:pro1_2}.
\end{proposition}

\subsubsection{Removing Inactive Zeroth Order Cuts}

To improve the effectiveness and reduce the complexity \cite{yang2014distributed,jiao2022asynchronous}, the inactive zeroth order cuts will be removed during the iteration process. The corresponding inner layer $P_{\rm{in}}^{t}$ and outer layer $P_{\rm{out}}^{t}$ will be updated as follows.

\begin{equation}
\label{eq:5_6_12}
   P_{\rm{in}}^{t}  = \left\{ \begin{array}{l}
{\rm{Remove}}(P_{\rm{in}}^{t}, cp_{{\rm{in}},l}), {\rm{if}} \; h_l^{\rm{in}}( \{{\boldsymbol{x}_{3,j}^{t}}\}, \boldsymbol{z}_1^{t}, {\boldsymbol{z}_2^{t}}', \boldsymbol{z}_3^{t}) \!<\! \varepsilon_{\rm{in}}, \forall l \\
P_{\rm{in}}^{t}, {\rm{otherwise}}
\end{array} \right. ,
\end{equation}
\begin{equation}
\label{eq:5_6_13}
   P_{\rm{out}}^{t}  = \left\{ \begin{array}{l}
{\rm{Remove}}(P_{\rm{out}}^{t}, cp_{{\rm{out}},l}), {\rm{if}} \; h_l^{\rm{out}}( \{{\boldsymbol{x}_{2,j}^{t}}\},\! \{{\boldsymbol{x}_{3,j}^{t}}\},\! \boldsymbol{z}_1^{t}, \boldsymbol{z}_2^{t}, \boldsymbol{z}_3^{t}) \!<\! \varepsilon_{\rm{out}}, \forall l \\
P_{\rm{out}}^{t}, {\rm{otherwise}}
\end{array} \right. ,
\end{equation}
where ${\rm{Remove}}(P_{\rm{in}}^{t}, cp_{{\rm{in}},l})$ and ${\rm{Remove}}(P_{\rm{out}}^{t}, cp_{{\rm{out}},l})$ respectively represent that the $l^{\rm{th}}$ inner layer and outer layer zeroth order cuts will be removed from $P_{\rm{in}}^{t}$ and $P_{\rm{out}}^{t}$.

\subsection{Zeroth Order Distributed Algorithm}

\label{distributed}
In this section, a distributed zeroth order algorithm is proposed. First, defining function $o(\{{\boldsymbol{x}_{2,j}}\}, \{{\boldsymbol{x}_{3,j}}\}, \boldsymbol{z}_1, \boldsymbol{z}_2, \boldsymbol{z}_3) = \sum_l \! \lambda_l [\max\{h_l^{\rm{out}}( \{{\boldsymbol{x}_{2,j}}\}, \{{\boldsymbol{x}_{3,j}}\}, \boldsymbol{z}_1, \boldsymbol{z}_2, \boldsymbol{z}_3) \!-\! \varepsilon_{\rm{out}},0 \}]^2$, where $\lambda_l>0$ is a penalty parameter. The constrained optimization problem described in Eq. (\ref{eq:4}) is reformulated as an unconstrained optimization problem by using the exterior penalty method \cite{shen2023penalty,shi2021improved,boyd2004convex} as follows.
\begin{equation}
\label{eq:5}
\begin{array}{l}
\! F(\{{\boldsymbol{x}_{1,j}}\},\! \{{\boldsymbol{x}_{2,j}}\},\! \{{\boldsymbol{x}_{3,j}}\},\! \boldsymbol{z}_1,\! \boldsymbol{z}_2,\! \boldsymbol{z}_3) \!= \!\sum\nolimits_{j = 1}^N \! {{f_{1,j}}({\boldsymbol{x}_{1,j}},{\boldsymbol{x}_{2,j}},{\boldsymbol{x}_{3,j}}) + \phi_j ||\boldsymbol{x}_{1,j} \!-\! \boldsymbol{z}_1 ||^2} \vspace{1mm} \\

\qquad \qquad \qquad \qquad \qquad \qquad \qquad \quad \; + o(\{{\boldsymbol{x}_{2,j}}\}, \{{\boldsymbol{x}_{3,j}}\}, \boldsymbol{z}_1, \boldsymbol{z}_2, \boldsymbol{z}_3),
\end{array}
\end{equation}
where $\phi_j \!>\!0$ is a penalty parameter. It is worth noting that the proposed DTZO is an expandable framework, allowing the incorporation of approaches beyond exterior penalty method, e.g., gradient projection based approaches \cite{xu2020unified} and Frank-Wolfe based methods \cite{shen2019complexities}.
We chose exterior penalty method because the lower-level problem often serves as a soft constraint (as discussed in Sec. \ref{Cascaded} and Appendix \ref{appedix:phi}) and using exterior penalty method offers comparatively \textit{lower} complexity. In addition, we demonstrate that the optimal solution to problem in Eq. (\ref{eq:5}) is a feasible solution to the original constrained problem; 2) the gap between the problem in Eq. (\ref{eq:5}) and original constrained problem will continuously decrease as $\lambda_l, \phi_j$ increase. Detailed discussions are provided in Appendix \ref{appendix:penalty}. In $(t+1)^{\rm{th}}$ iteration, the proposed algorithm proceeds as follows.

\textbf{In Worker $j$}. After receiving the updated parameters $\boldsymbol{z}_i^{t}$ and $\nabla_{\boldsymbol{x}_{i,j}} o( \{{\boldsymbol{x}_{2,j}^{t}}\}, \{{\boldsymbol{x}_{3,j}^{t}}\}, \boldsymbol{z}_1^{t}, \boldsymbol{z}_2^{t}, \boldsymbol{z}_3^{t})$, worker $j$ updates the local variables as follows,
\begin{equation}
\label{eq:5_6_15}
  {\boldsymbol{x}_{1,j}^{t+1}} = {\boldsymbol{x}_{1,j}^{t}} - \eta_{\boldsymbol{x}_1}  G_{\boldsymbol{x}_{1,j}}(\{{\boldsymbol{x}_{1,j}^t}\}, \{{\boldsymbol{x}_{2,j}^t}\}, \{{\boldsymbol{x}_{3,j}^t}\}, \boldsymbol{z}_1^t, \boldsymbol{z}_2^t, \boldsymbol{z}_3^t),
\end{equation}
\begin{equation}
\label{eq:5_6_16}
  {\boldsymbol{x}_{2,j}^{t+1}} = {\boldsymbol{x}_{2,j}^{t}} - \eta_{\boldsymbol{x}_2} G_{\boldsymbol{x}_{2,j}}(\{{\boldsymbol{x}_{1,j}^t}\}, \{{\boldsymbol{x}_{2,j}^t}\}, \{{\boldsymbol{x}_{3,j}^t}\}, \boldsymbol{z}_1^t, \boldsymbol{z}_2^t, \boldsymbol{z}_3^t),
\end{equation}
\begin{equation}
\label{eq:5_6_17}
  {\boldsymbol{x}_{3,j}^{t+1}} = {\boldsymbol{x}_{3,j}^{t}} - \eta_{\boldsymbol{x}_3} G_{\boldsymbol{x}_{3,j}}(\{{\boldsymbol{x}_{1,j}^t}\}, \{{\boldsymbol{x}_{2,j}^t}\}, \{{\boldsymbol{x}_{3,j}^t}\}, \boldsymbol{z}_1^t, \boldsymbol{z}_2^t, \boldsymbol{z}_3^t),
\end{equation}
we have that, 
\begin{equation}
\label{eq:5_6_18}
\begin{array}{l}
  G_{\boldsymbol{x}_{1,j}}(\{{\boldsymbol{x}_{1,j}^t}\}, \{{\boldsymbol{x}_{2,j}^t}\}, \{{\boldsymbol{x}_{3,j}^t}\}, \boldsymbol{z}_1^t, \boldsymbol{z}_2^t, \boldsymbol{z}_3^t)
   \\   =  \frac{{{f_{1,j}}({\boldsymbol{x}_{1,j}^t} + \mu {\boldsymbol{u}_{k,1}},{\boldsymbol{x}_{2,j}^t},{\boldsymbol{x}_{3,j}^t}) -{f_{1,j}}({\boldsymbol{x}_{1,j}^t},{\boldsymbol{x}_{2,j}^t},{\boldsymbol{x}_{3,j}^t})}}{\mu } {\boldsymbol{u}_{k,1}}+  2\phi_j (\boldsymbol{x}_{1,j}^t - \boldsymbol{z}_1^t),
\end{array}
\end{equation}
\begin{equation}
\label{eq:5_6_19}
 \begin{array}{l}
  G_{\boldsymbol{x}_{2,j}}(\{{\boldsymbol{x}_{1,j}^t}\}, \{{\boldsymbol{x}_{2,j}^t}\}, \{{\boldsymbol{x}_{3,j}^t}\}, \boldsymbol{z}_1^t, \boldsymbol{z}_2^t, \boldsymbol{z}_3^t) \\
  \!=\!  \frac{{{f_{1,j}}({\boldsymbol{x}_{1,j}^t},{\boldsymbol{x}_{2,j}^t} + \mu {\boldsymbol{u}_{k,2}},{\boldsymbol{x}_{3,j}^t}) -{f_{1,j}}({\boldsymbol{x}_{1,j}^t},{\boldsymbol{x}_{2,j}^t},{\boldsymbol{x}_{3,j}^t})}}{\mu } {\boldsymbol{u}_{k,2}} \!+\! \nabla_{\boldsymbol{x}_{2,j}}  o( \{{\boldsymbol{x}_{2,j}^t}\},\! \{{\boldsymbol{x}_{3,j}^t}\},\! \boldsymbol{z}_1^t, \boldsymbol{z}_2^t, \boldsymbol{z}_3^t),
\end{array}
\end{equation}
\begin{equation}
\label{eq:5_6_20}
\begin{array}{l}
  G_{\boldsymbol{x}_{3,j}}(\{{\boldsymbol{x}_{1,j}^t}\}, \{{\boldsymbol{x}_{2,j}^t}\}, \{{\boldsymbol{x}_{3,j}^t}\}, \boldsymbol{z}_1^t, \boldsymbol{z}_2^t, \boldsymbol{z}_3^t) \\
 \! = \! \frac{{{f_{1,j}}({\boldsymbol{x}_{1,j}^t},{\boldsymbol{x}_{2,j}^t},{\boldsymbol{x}_{3,j}^t} + \mu {\boldsymbol{u}_{k,3}}) -{f_{1,j}}({\boldsymbol{x}_{1,j}^t},{\boldsymbol{x}_{2,j}^t},{\boldsymbol{x}_{3,j}^t})}}{\mu } {\boldsymbol{u}_{k,3}} \!+ \!\nabla_{\boldsymbol{x}_{3,j}} o( \{{\boldsymbol{x}_{2,j}^t}\},\! \{{\boldsymbol{x}_{3,j}^t}\},\! \boldsymbol{z}_1^t, \boldsymbol{z}_2^t, \boldsymbol{z}_3^t),
\end{array}
\end{equation}
where ${\boldsymbol{u}_{k,i}}\! \in \!\mathbb{R}^{d_i},\forall i$ are standard Gaussian random vectors, $\mu \!>\!0$ is smoothing parameter, $\eta_{\boldsymbol{x}_i},\forall i$ are step-sizes.  Then, the updated variables $ {\boldsymbol{x}_{1,j}^{t+1}},  {\boldsymbol{x}_{2,j}^{t+1}}, {\boldsymbol{x}_{3,j}^{t+1}}$ will be transmitted to the master.

\textbf{In Master}. After receiving updated variables from workers, the master performs the following steps,

1. Updating consensus variables,
    \begin{equation}
    \label{eq:5_6_21}
  {\boldsymbol{z}_1^{t+1}} = {\boldsymbol{z}_1^{t}}\! -\! \eta_{\boldsymbol{z}_1}  \left( \sum\nolimits_j \! 2\phi_j(\boldsymbol{z}_1^t\!-\!\boldsymbol{x}_{1,j}^t) \!+\! \nabla_{\boldsymbol{z}_1} o( \{{\boldsymbol{x}_{2,j}^t}\}, \{{\boldsymbol{x}_{3,j}^t}\}, \boldsymbol{z}_1^t, \boldsymbol{z}_2^t, \boldsymbol{z}_3^t) \right),
\end{equation}
\begin{equation}
\label{eq:5_6_22}
  {\boldsymbol{z}_2^{t+1}} = {\boldsymbol{z}_2^{t}} - \eta_{\boldsymbol{z}_2}  \nabla_{\boldsymbol{z}_2} o( \{{\boldsymbol{x}_{2,j}^t}\}, \{{\boldsymbol{x}_{3,j}^t}\}, \boldsymbol{z}_1^t, \boldsymbol{z}_2^t, \boldsymbol{z}_3^t),
\end{equation}
\begin{equation}
\label{eq:5_6_23}
  {\boldsymbol{z}_3^{t+1}} = {\boldsymbol{z}_3^{t}} - \eta_{\boldsymbol{z}_3} \nabla_{\boldsymbol{z}_3} o( \{{\boldsymbol{x}_{2,j}^t}\}, \{{\boldsymbol{x}_{3,j}^t}\}, \boldsymbol{z}_1^t, \boldsymbol{z}_2^t, \boldsymbol{z}_3^t),
\end{equation}
where $\eta_{\boldsymbol{z}_1}, \eta_{\boldsymbol{z}_2}$ and $\eta_{\boldsymbol{z}_3}$ are step-sizes.

2.  Computing gradient of $o( \{{\boldsymbol{x}_{2,j}^{t+1}}\}, \{{\boldsymbol{x}_{3,j}^{t+1}}\}, \boldsymbol{z}_1^{t+1}, \boldsymbol{z}_2^{t+1}, \boldsymbol{z}_3^{t+1})$. Broadcasting the updated parameters $\boldsymbol{z}_i^{t+1}, i=1,2,3$ and $\nabla_{\boldsymbol{x}_{i,j}} o( \{{\boldsymbol{x}_{2,j}^{t+1}}\}, \{{\boldsymbol{x}_{3,j}^{t+1}}\}, \boldsymbol{z}_1^{t+1}, \boldsymbol{z}_2^{t+1}, \boldsymbol{z}_3^{t+1}), i=2,3$ to workers.

\textbf{\underline{Discussion:}} TLL with \textit{level-wise} zeroth order constraints is considered in this work, where first order information at \textit{each level} is unavailable. Note that the proposed DTZO is versatile and can be adapted to a wide range of TLL, e.g., grey-box TLL (gradients at some levels in TLL are available \cite{huang2024enhancing}), with slight adjustments. For instance, if gradients at first-level in TLL are accessible, we can use gradient descent steps to replace Eq. (\ref{eq:5_6_15})-(\ref{eq:5_6_17}). Similarly, if the second or third-level gradients are available, first order based cuts, e.g., \cite{jiao2024provably}, can be employed to construct the cascaded polynomial approximation. Detailed discussions are offered in Appendix \ref{appendix:partial zeroth order}.

\begin{algorithm}[t]
   \caption{DTZO: Distributed Trilevel Zeroth Order Learning}
\begin{algorithmic}
   \STATE {\bfseries Initialization:}  master iteration $t = 0$, variables $\{{\boldsymbol{x}_{1,j}^0}\}, \{{\boldsymbol{x}_{2,j}^0}\}, \{{\boldsymbol{x}_{3,j}^0}\}, \boldsymbol{z}_1^0, \boldsymbol{z}_2^0, \boldsymbol{z}_3^0$.
   
   \REPEAT
   
   \FOR{\emph{local worker $j$}}
   \STATE updates the local variables ${\boldsymbol{x}_{1,j}^{t+1}},{\boldsymbol{x}_{2,j}^{t+1}},{\boldsymbol{x}_{3,j}^{t+1}}$ according to Eq. (\ref{eq:5_6_15})-(\ref{eq:5_6_20});
   \ENDFOR
   
   \STATE \textit{local workers} transmit the updated variables to the master;
   
   \FOR{\emph{master}}
   \STATE updates consensus variables ${\boldsymbol{z}_1^{t+1}}, {\boldsymbol{z}_2^{t+1}}, {\boldsymbol{z}_3^{t+1}}$ according to Eq. (\ref{eq:5_6_21})-(\ref{eq:5_6_23});

   \STATE computes $\nabla o( \{{\boldsymbol{x}_{2,j}^{t+1}}\}, \{{\boldsymbol{x}_{3,j}^{t+1}}\}, \boldsymbol{z}_1^{t+1}, \boldsymbol{z}_2^{t+1}, \boldsymbol{z}_3^{t+1})$;
   
   \ENDFOR

   \STATE \textit{master} broadcasts the updated parameters and gradients to workers;

   \IF{$(t+1)$ mod $\mathcal{T}$ $==0$ and $t<T_1$}

   \STATE  new inner layer zeroth order cuts are generated by Eq. (\ref{eq:4_25_17}) and (\ref{eq:5_6_9});

   \STATE  new outer layer zeroth order cuts are generated by Eq. (\ref{eq:4_25_25}) and (\ref{eq:5_6_11});

   \STATE inactive zeroth order cuts are deleted by (\ref{eq:5_6_12}) and (\ref{eq:5_6_13});
   
   \ENDIF

   \STATE $t =t+1$;
   \UNTIL{termination.}

\end{algorithmic}
\label{algorithm}
\end{algorithm}

\section{Theoretical Analysis}
\label{sec:theoretical analysis}

\begin{definition}
\label{def:1}
    \textbf{(Stationarity Gap)} Following \cite{xu2020unified,jiao2022asynchronous}, the stationarity gap at $t^{\rm{th}}$ iteration in this problem can be expressed as,
    \begin{equation}
    \label{eq:4_24_29}
        {{\cal G}^t} = \left[ \begin{array}{l}
\{{\nabla _{{\boldsymbol{x}_{1,j}}}}F(\{{\boldsymbol{x}_{1,j}^t}\}, \{{\boldsymbol{x}_{2,j}^t}\}, \{{\boldsymbol{x}_{3,j}^t}\}, \boldsymbol{z}_1^t, \boldsymbol{z}_2^t, \boldsymbol{z}_3^t)\}\\

\{{\nabla _{{\boldsymbol{x}_{2,j}}}}F(\{{\boldsymbol{x}_{1,j}^t}\}, \{{\boldsymbol{x}_{2,j}^t}\}, \{{\boldsymbol{x}_{3,j}^t}\}, \boldsymbol{z}_1^t, \boldsymbol{z}_2^t, \boldsymbol{z}_3^t)\}\\

\{{\nabla _{{\boldsymbol{x}_{3,j}}}}F(\{{\boldsymbol{x}_{1,j}^t}\}, \{{\boldsymbol{x}_{2,j}^t}\}, \{{\boldsymbol{x}_{3,j}^t}\}, \boldsymbol{z}_1^t, \boldsymbol{z}_2^t, \boldsymbol{z}_3^t)\} \\

{\nabla _{{\boldsymbol{z}_1}}}F(\{{\boldsymbol{x}_{1,j}^t}\}, \{{\boldsymbol{x}_{2,j}^t}\}, \{{\boldsymbol{x}_{3,j}^t}\}, \boldsymbol{z}_1^t, \boldsymbol{z}_2^t, \boldsymbol{z}_3^t)  \\

{\nabla _{{\boldsymbol{z}_2}}}F(\{{\boldsymbol{x}_{1,j}^t}\}, \{{\boldsymbol{x}_{2,j}^t}\}, \{{\boldsymbol{x}_{3,j}^t}\}, \boldsymbol{z}_1^t, \boldsymbol{z}_2^t, \boldsymbol{z}_3^t)  \\

{\nabla _{{\boldsymbol{z}_3}}}F(\{{\boldsymbol{x}_{1,j}^t}\}, \{{\boldsymbol{x}_{2,j}^t}\}, \{{\boldsymbol{x}_{3,j}^t}\}, \boldsymbol{z}_1^t, \boldsymbol{z}_2^t, \boldsymbol{z}_3^t)
\end{array} 
\right].
    \end{equation}
It is seen from Eq. (\ref{eq:4_24_29}) that,
\begin{equation}
\begin{array}{l}
    ||{{\cal G}^{t}}|{|^2}  
 = \sum\nolimits_{i = 1}^3 \sum\nolimits_{j = 1}^N  {||{\nabla _{{x_{i,j}}}}F(\{{\boldsymbol{x}_{1,j}^t}\}, \{{\boldsymbol{x}_{2,j}^t}\}, \{{\boldsymbol{x}_{3,j}^t}\}, \boldsymbol{z}_1^t, \boldsymbol{z}_2^t, \boldsymbol{z}_3^t)|{|^2}} 
\\
 \qquad \quad \;\; + \sum\nolimits_{i=1}^3 ||{\nabla _{{\boldsymbol{z}_i}}}{F }(\{{\boldsymbol{x}_{1,j}^t}\}, \{{\boldsymbol{x}_{2,j}^t}\}, \{{\boldsymbol{x}_{3,j}^t}\}, \boldsymbol{z}_1^t, \boldsymbol{z}_2^t, \boldsymbol{z}_3^t)||^2 .
\end{array}
\end{equation}

\end{definition}

\begin{definition}
\label{def:2}
    \textbf{($\epsilon$-Stationary Point)} $(\{{\boldsymbol{x}_{1,j}^t}\}, \{{\boldsymbol{x}_{2,j}^t}\}, \{{\boldsymbol{x}_{3,j}^t}\}, \boldsymbol{z}_1^t, \boldsymbol{z}_2^t, \boldsymbol{z}_3^t)$ is the stationary point when $||{{\cal G}^t}||^2=0$. $(\{{\boldsymbol{x}_{1,j}^t}\}, \{{\boldsymbol{x}_{2,j}^t}\}, \{{\boldsymbol{x}_{3,j}^t}\}, \boldsymbol{z}_1^t, \boldsymbol{z}_2^t, \boldsymbol{z}_3^t)$ is the $\epsilon$-stationary point when $||{{\cal G}^t}||^2 \le \epsilon$. Defining $T(\epsilon)$ as the first iteration when $||{{\cal G}^t}||^2 \le \epsilon$, i.e., $T(\epsilon) = \min \{ t|\;||{{\cal G}^t}|{|^2} \le \epsilon \} $.
\end{definition}

\begin{definition}
\label{def:3}
    \textbf{($\mu$-Smooth Approximation)}  Following \cite{ghadimi2013stochastic,fang2022communication,nesterov2017random,kornilov2024accelerated,rando2024optimal}, the $\mu$-smooth approximation of a function $F(\boldsymbol{w}): \mathbb{R}^d  \rightarrow  \mathbb{R}^1$ is given by,
\begin{equation}
F_\mu(\boldsymbol{w}) = \frac{1}{(2\pi)^{\frac{d}{2}}} \int F(\boldsymbol{w} + \mu \boldsymbol{u})e^{-\frac{1}{2}||\boldsymbol{u}||^2} d \boldsymbol{u} = \mathbb{E}_{\boldsymbol{u}}\left[F(\boldsymbol{w} + \mu \boldsymbol{u})\right],    
\end{equation}
where $\boldsymbol{u} \in \mathbb{R}^d$ is a standard Gaussian random vector and $\mu > 0$ is the smoothing parameter.
\end{definition}

\begin{assumption}
\label{assum:1}
    \textbf{(Boundedness)} Following many works in machine learning, e.g., \cite{deng2020distributionally,jiao2022asynchronous,qian2019robust,lei2018stochastic,zheng2017asynchronous}, the bounded domain is assumed, i.e., $||\boldsymbol{x}_{i,j}-\boldsymbol{x}_{i,j}^*||^2 \le \alpha_i, \forall \boldsymbol{x}_{i,j}, ||\boldsymbol{z}_{i}-\boldsymbol{z}_{i}^*||^2 \le \alpha_i, \forall \boldsymbol{z}_{i}$, where $\boldsymbol{x}_{i,j}^*, \boldsymbol{z}_{i}^*$ denote the optimal solution. Following \cite{cutkosky2019momentum,liu2021conflict,fang2022communication,shaban2019truncated}, we assume the optimal value ${F_\mu }^*>-\infty$. 
\end{assumption}

\begin{assumption}
\label{assum:2}
    \textbf{(L-smoothness)} Following many work in nested optimization and zeroth order learning, e.g., \cite{chen2023decentralized,lin2024non,ghadimi2013stochastic}, we assume the gradient of function $F$ is Lipschitz continuous with constant $L < \infty$, that is, for any point $\boldsymbol{w},\boldsymbol{w}'$, we have that,
    \begin{equation}
        ||\nabla F(\boldsymbol{w}) - \nabla F(\boldsymbol{w}')|| \le L||\boldsymbol{w}-\boldsymbol{w}'||.
    \end{equation}
    It is worth noting that both Assumptions \ref{assum:1} and \ref{assum:2} are mild and commonly used in machine learning. Detailed discussions of these assumptions are provided in Appendix \ref{appendix:assum}.
\end{assumption}

\begin{theorem}
\label{theorem:1}
    \textbf{(Iteration Complexity)} Under Assumption \ref{assum:1} and \ref{assum:2}, by setting step-sizes $ {\eta_{\boldsymbol{x}_i}} =
    {\eta_{\boldsymbol{z}_i}} =  \min \left\{ {\frac{1}{{8L({d_1} + 4)}},\frac{1}{{8L({d_2} + 4)}},\frac{1}{{8L({d_3} + 4)}}, \frac{3}{2(L+1)}, \frac{1}{{\sqrt {T(\epsilon) - {T_1}} }}} \right\}, i=1,2,3$ and letting smoothing parameter $0< \mu \le \frac{1}{\sqrt {T(\epsilon) - {T_1}}}$, we have that,
    \begin{equation}
    \begin{array}{l}
T(\epsilon)
 \sim \mathcal{O} \left( {\left( {\sum\nolimits_{i = 1}^3 {\overline{c_i} + \overline{d}\left( {\mathop {\max }\limits_{t \in [{T_1}]} {F_\mu }(\{{\boldsymbol{x}_{i,j}^t}\},\{ \boldsymbol{z}_i^t\}) - {F_\mu }^*} \right)} } \right)^2}\frac{1}{{{{\epsilon}^2}}} + {T_1}\right),
\end{array}
\end{equation}
where constants   $\overline{d}=4(1 + {\max \left\{ {8L({d_1} + 4),8L({d_2} + 4),8L({d_3} + 4), \frac{2(L+1)}{3}} \right\})}$, $\overline{c_i} = {\frac{{{L^2}{{({d_i} + 6)}^3}}}{{4\left( {{d_i} + 4} \right)}} + {L^2}{{({d_i} \!+\! 3)}^3}}  + 4 L (N+1)d_i \left({\max \left\{ {8L({d_1} + 4),8L({d_2} + 4),8L({d_3} + 4), \frac{2(L+1)}{3}} \right\} \!+\! 1}\right) $. $T_1 > 0$ is a constant that controls the cascaded polynomial approximation, as discussed in Sec. \ref{refine}. Detailed proofs of Theorem \ref{theorem:1} are provided in Appendix \ref{appendix:iteration_complexity}, with further discussions offered below.
\end{theorem}

\begin{theorem}
    \label{theorem:2}
    \textbf{(Communication Complexity)} The overall communication complexity of the proposed DTZO can be divided into the communication complexity at every iteration ($C_1$) and the communication complexity of updating zeroth order cuts ($C_2$). Specifically, the overall communication complexity can be expressed as $C_1 + C_2 = T(\epsilon)(2d_1 + 3d_2 +3d_3)N +   2 N \lfloor \frac{T_1}{\mathcal{T}} \rfloor  \mathcal{T} (d_2 + d_3)$. The detailed proofs are provided in Appendix \ref{appendix:communication_complexity}, with further discussions offered as follows.
\end{theorem}

\textbf{\underline{Discussion:}}  It is seen from Theorem \ref{theorem:1} and \ref{theorem:2} that the proposed framework DTZO can \textit{flexibly} control the trade-off between the performance of cascaded polynomial approximation and the iteration complexity (i.e., $T(\epsilon)$ in Theorem \ref{theorem:1}) and communication complexity (i.e., $C_1 + C_2$ in Theorem \ref{theorem:2}) by adjusting a single parameter $T_1$. Specifically, a larger $T_1$ corresponds to a better cascaded polynomial approximation, but it also entails higher iteration and communication complexity. Consequently, if the distributed system has limited computational and communication capabilities, a smaller value of $T_1$ can be selected. Conversely, if a higher quality of cascaded polynomial approximation is desired, a larger value of $T_1$ can be chosen, which demonstrates the flexibility in the proposed framework. In addition, as shown in Theorem \ref{theorem:1}, the iteration complexity of the proposed distributed trilevel zeroth order learning framework can be written as $\mathcal{O}(\sum_{i}d_i^6/\epsilon^2)$. It is worth mentioning that the dimension-dependent iteration complexity is \textit{common} in zeroth order optimization, as discussed in various works \cite{zhangrevisiting,zhang2024subspace,duchi2015optimal,sun2022black,qiu2023zeroth}. For instance, the iteration complexity of the state-of-the-art distributed zeroth order bilevel learning method \cite{qiu2023zeroth} is given by $\mathcal{O}(d^8/\epsilon^2)$, where $d$ denotes the dimension of variables.


\section{Experiments}
\label{sec:experiments}

In the experiment, two distributed trilevel zeroth order learning scenarios, i.e., black-box trilevel learning on large language models (LLMs) and robust hyperparameter optimization are used to evaluate the performance of the proposed DTZO. In the zeroth order setting, the existing distributed nested optimization algorithms based on first order information, e.g., \cite{jiao2024provably}, are not available in the experiment. The proposed DTZO is compared with the state-of-the-art distributed zeroth order learning method FedZOO \cite{fang2022communication} and distributed bilevel zeroth order learning method FedRZO$_{\rm{bl}}$ \cite{qiu2023zeroth}. In the experiment, all the models are implemented using PyTorch, and the experiments are conducted on a server equipped with two NVIDIA RTX 4090 GPUs. More experimental details are provided in Appendix \ref{appendix:experiment}.






\begin{figure*}[t]
\centering
\subfigure[Comparisons about ASR]
{\begin{minipage}{7cm}
\label{fig:ASR}
\includegraphics[scale=0.4]{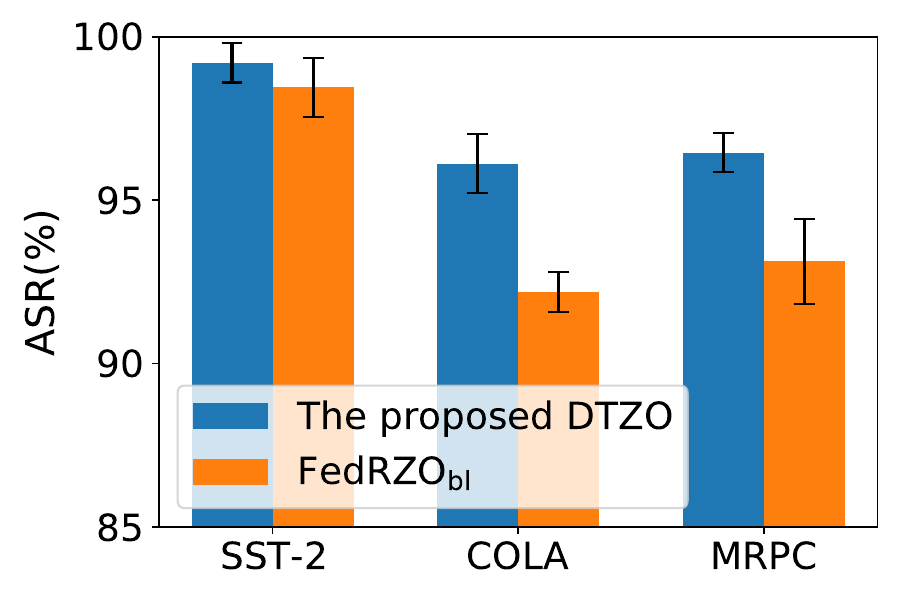}  
\end{minipage}}
\subfigure[Comparisons about ACC] 
{\begin{minipage}{7cm}
\label{fig:BA}
\includegraphics[scale=0.4]{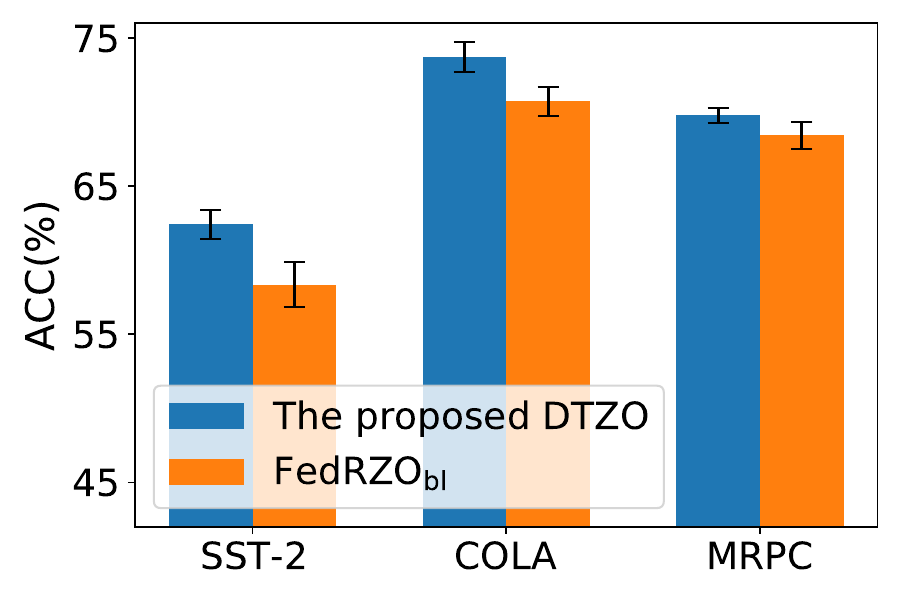} 
\end{minipage}}
\caption{Comparisons about ASR and ACC between the proposed DTZO and the state-of-the-art distributed bilevel zeroth order learning method FedRZO$_{\rm{bl}}$ \cite{qiu2023zeroth}.} 
\label{fig:LLM}
\end{figure*}

\subsection{Black-Box Trilevel Learning}
Prompt learning is a key technique for enabling LLMs to efficiently and effectively adapt to various downstream tasks \cite{ma2024fairness,wang2024grammar}. {In many practical scenarios involving LLMs, access to first-order information is restricted due to the proprietary nature of these models or API constraints. For instance, commercial LLM APIs only allow input-output interactions and do not provide visibility into gradients.}  Inspired by the black-box prompt learning \cite{diao2022black} and backdoor attack on prompt-based LLMs \cite{yao2024poisonprompt}, the backdoor attack on black-box LLMs is considered in the experiment, which can be expressed as a black-box trilevel learning problem,
\begin{equation}
    \begin{array}{l}
\mathop {\min }\limits_{\lambda}\sum\nolimits_{j = 1}^N\frac{1}{{|{D^{\rm{val}}_j}|}}\sum\limits_{(\boldsymbol{s}_i,{y_i}) \sim {D^{\rm{val}}_j}} {L({\cal G},[{\boldsymbol{k}_{\rm{tri}}},\boldsymbol{p} ,\boldsymbol{s}_i],{y_i})}  \\{\rm{s.t.}}\; {\boldsymbol{k}_{\rm{tri}}} = \mathop {\arg \min }\limits_{{\boldsymbol{k}_{\rm{tri}}}'}\sum\nolimits_{j = 1}^N  \frac{1}{{|{D^{\rm{tr}}_j}|}}\sum\limits_{(\boldsymbol{s}_i,{y_i}) \sim {D^{\rm{tr}}_j}} {L({\cal G},[{\boldsymbol{k}_{\rm{tri}}}',\boldsymbol{p} ,\boldsymbol{s}_i],{y_i})} + \lambda ||{\boldsymbol{k}_{\rm{tri}}}'||^2 \\\qquad \quad {\rm{s}}{\rm{.t}}{\rm{.}} \; \boldsymbol{p} = \mathop {\arg \min }\limits_{\boldsymbol{p}'}\sum\nolimits_{j = 1}^N \frac{1}{{|{D^{\rm{tr}}_j}|}}\sum\limits_{(\boldsymbol{s}_i,{y_i}) \sim {D^{\rm{tr}}_j}} {L({\cal G},[{\boldsymbol{k}_{\rm{tri}}}',\boldsymbol{p}',\boldsymbol{s}_i],{y_i})}
\\
{\mathop{\rm var}} .\qquad \qquad \qquad \lambda ,{\boldsymbol{k}_{\rm{tri}}}, \boldsymbol{p},
\end{array}
\end{equation}
where ${\cal G}$ denotes the black-box LLM. $\lambda$, ${\boldsymbol{k}_{\rm{tri}}}$, $ \boldsymbol{p}$ respectively denote the hyperparameter, backdoor trigger, and prompt. ${D^{\rm{tr}}_j}$ and ${D^{\rm{val}}_j}$ denote the training and validation dataset in $j^{\rm{th}}$ worker, and $N$ denotes the number of workers. $\boldsymbol{s}_i,{y_i}$ denote the $i^{\rm{th}}$ input sentence and label. In the experiment, Qwen 1.8B-Chat \cite{bai2023qwen} is utilized as the black-box LLM. The General Language Understanding Evaluation (GLUE) benchmark \cite{wang2018glue} is used to evaluate the proposed DTZO. Specifically, the experiments are carried out on: 1) SST-2 for sentiment analysis; 2) COLA for linguistic acceptability; and 3) MRPC for semantic equivalence of sentences. In this task, we aim to obtain the effective backdoor triggers while ensuring the model performance on clean inputs (i.e., inputs without triggers). Therefore, following \cite{yao2024poisonprompt}, the Attack Success Rate (ASR) when the triggers are activated and the Accuracy (ACC) on clean samples are utilized as the metrics in the experiments. The comparisons between the proposed DTZO and the state-of-the-art distributed bilevel zeroth order learning method FedRZO$_{\rm{bl}}$ are illustrated in Figure \ref{fig:LLM}. It is seen from Figure \ref{fig:ASR} and \ref{fig:BA} that the proposed DTZO can effectively tackle the distributed trilevel zeroth order learning problem and achieve superior performance than FedRZO$_{\rm{bl}}$ since the proposed DTZO is capable of addressing higher-nested zeroth order learning problems compared to FedRZO$_{\rm{bl}}$. 


\subsection{Robust Hyperparameter Optimization}

Inspired by \cite{sato2021gradient,jiao2024provably} in trilevel learning, the robust hyperparameter optimization is considered in the experiment, which can be formulated as follows.

\begin{equation}
    \begin{array}{l}
\min\limits_{\varphi} \sum\nolimits_{j = 1}^N {{f_j}(X_j^{{\mathop{\rm var}} },y_j^{{\mathop{\rm var}} },\boldsymbol{w})} \\{\rm{s}}{\rm{.t}}{\rm{.}}\; \boldsymbol{w} = \mathop {\arg \min }\limits_{\boldsymbol{w}'} \sum\nolimits_{j = 1}^N {{f_j}(X_j^{\rm{tr}} + {p_j},y_j^{\rm{tr}},\boldsymbol{w}')}  + \varphi ||\boldsymbol{w}'|{|^2}\\\qquad {\rm{s}}{\rm{.t}}{\rm{.}} \; \boldsymbol{p} = \mathop {\arg \max }\limits_{\boldsymbol{p}'} \sum\nolimits_{j = 1}^N {{f_j}(X_j^{\rm{tr}} + {p_j}',y_j^{\rm{tr}},\boldsymbol{w}')} \\
{\mathop{\rm var}} .\qquad \qquad \varphi ,\boldsymbol{w},\boldsymbol{p},
\end{array}
\end{equation}
where $N$ represents the number of workers in a distributed system, $\varphi$, $\boldsymbol{w}$, and $\boldsymbol{p}'=[{p_1}', \cdots, {p_N}']$ denote the regularization coefficient, model parameter, and adversarial noise, respectively. $X_j^{\rm{tr}}$ and $y_j^{\rm{tr}}$ represent the training data and labels, while $X_j^{\rm{var}}$ and $y_j^{\rm{var}}$ represent the validation data and labels, respectively. Following the setting for nondifferentiable functions as described in \cite{qiu2023zeroth}, ReLU neural networks are employed in the experiments. The digits recognition tasks in \cite{qian2019robust,wang2021discriminative} with four benchmark datasets, i.e., MNIST \cite{lecun1998gradient}, USPS, Fashion MNIST \cite{xiao2017fashion}, and QMNIST \cite{yadav2019cold}, are utilized to assess the performance of the proposed DTZO. The average across
accuracy on clean samples and robustness against adversarial samples is used as the metric, more details about the experimental setting are provided in Appendix \ref{appendix:experiment}.  We compare the proposed DTZO with the state-of-the-art methods FedZOO \cite{fang2022communication} and   FedRZO$_{\rm{bl}}$ \cite{qiu2023zeroth} in Table \ref{tab:experiments_nondiffe}. It is seen from Table \ref{tab:experiments_nondiffe} that the proposed DTZO can effectively tackle the trilevel zeroth order learning problem in a distributed manner. The superior performance of DTZO, as compared to state-of-the-art methods, can be attributed to its ability to address higher-nested zeroth order learning problems.

\begin{figure}[t]
    \centering
    \begin{minipage}{0.48\textwidth}
        \centering
\includegraphics[width=0.74\linewidth]{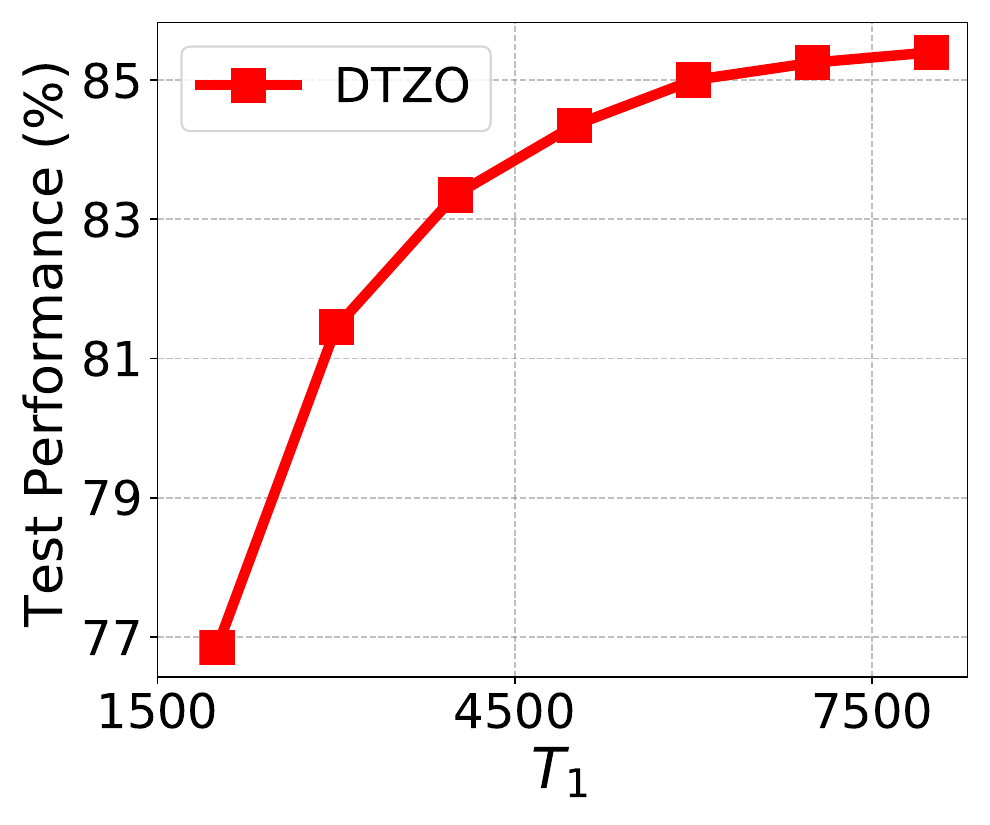}
    \caption{Adjusting $T_1$ can flexibly control the trade-off between performance and complexity, results on USPS dataset.}
    \label{fig:T1}
    \end{minipage}\hfill
    \begin{minipage}{0.48\textwidth}
        \centering
\includegraphics[width=0.80\linewidth]{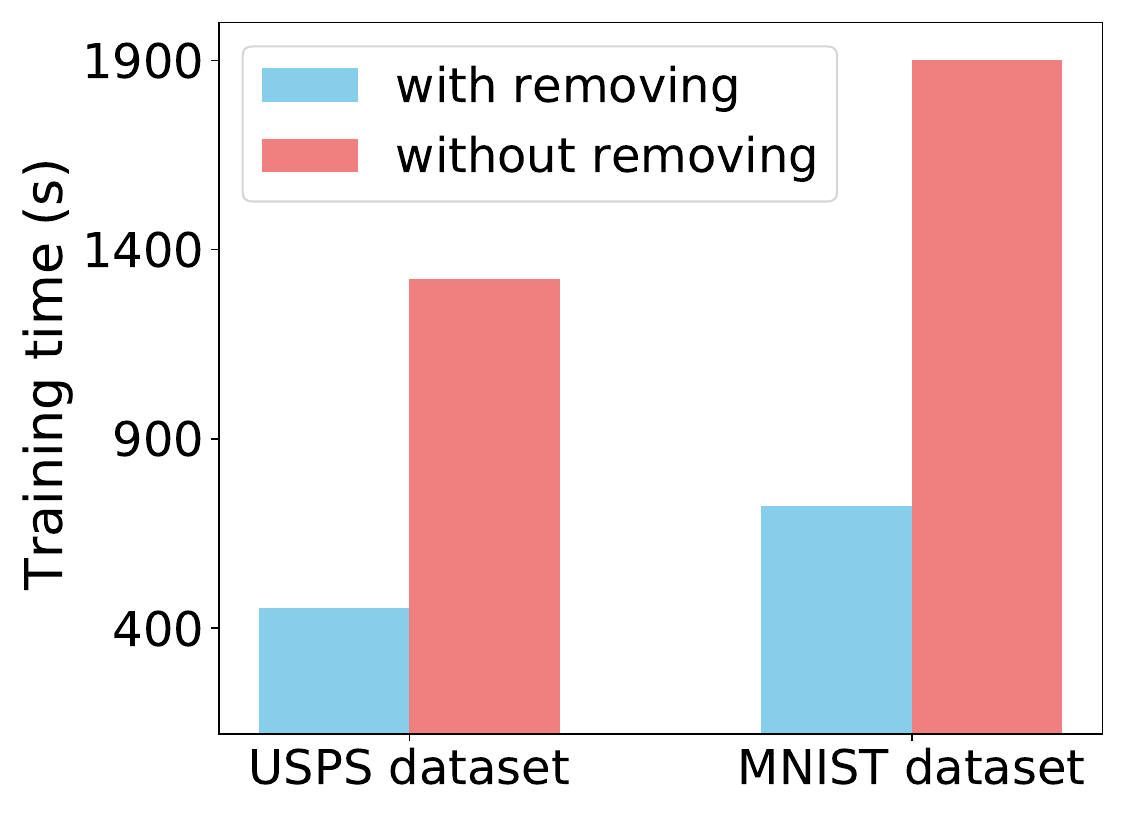}
    \caption{Training time (1000 communication rounds) of with and without removing inactive cuts.}
    \label{fig:training time}
    \end{minipage}
\end{figure}

\begin{figure}[t]
    \centering
    \begin{minipage}{0.48\textwidth}
        \centering
\includegraphics[width=0.75\linewidth]{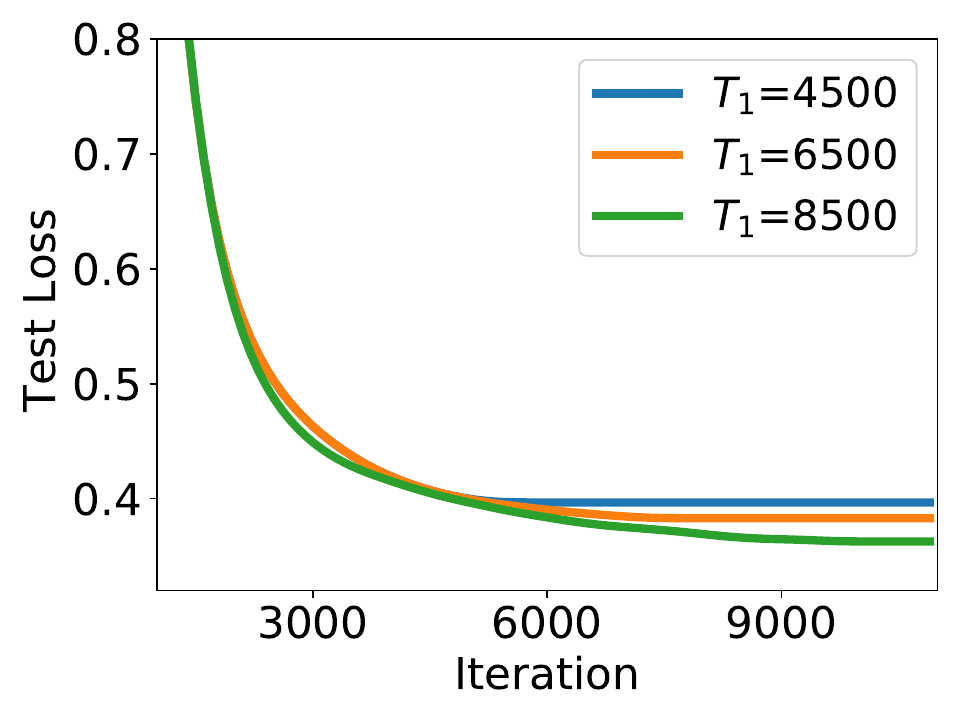}
    \caption{{Test loss of the proposed DTZO under various setting of $T_1$, results on USPS dataset.}}
    \label{fig:T1_test_loss}
    \end{minipage}\hfill
    \begin{minipage}{0.48\textwidth}
        \centering
\includegraphics[width=0.75\linewidth]{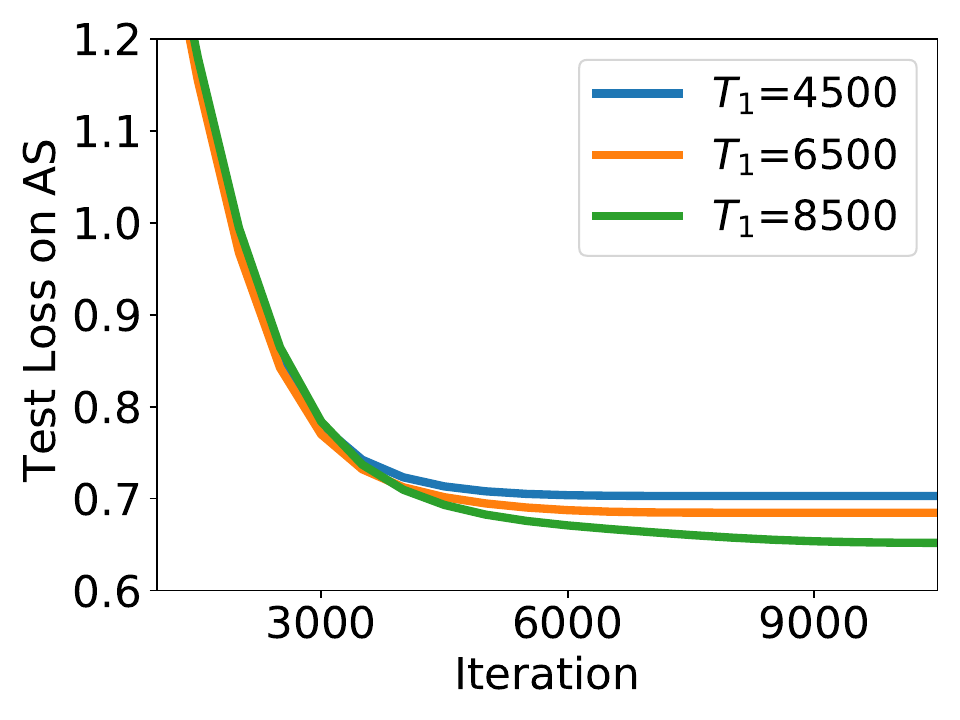}
    \caption{{Test loss on AS (adversarial samples) of DTZO under various setting of $T_1$.}}
\label{fig:T1_robust_test_loss}
    \end{minipage}
\end{figure}

Within the proposed framework, the trade-off between complexity and performance can be flexibly controlled by adjusting $T_1$, as discussed in Sec. \ref{sec:theoretical analysis}. Specifically, if the distributed system has limited computational and communication capabilities, a smaller $T_1$ can be selected. Conversely, if higher performance is required, a larger $T_1$ can be chosen. As shown in Figure \ref{fig:T1}, the performance of the proposed framework improves with increasing $T_1$, allowing for flexible adjustments based on system requirements. Removing inactive cuts can significantly improve the effectiveness of cutting plane method, as discussed in \cite{jiao2024provably,yang2014distributed}. In the experiment, we also investigate the effect of removing inactive cuts within the proposed DTZO. It is seen from Figure \ref{fig:training time} that pruning inactive cuts significantly reduces training time, indicating the importance of this procedure. In addition, the impact of different choices of $T_1$ on the convergence rate within the proposed framework is evaluated. As illustrated in Figures \ref{fig:T1_test_loss} and \ref{fig:T1_robust_test_loss}, a smaller $T_1$ leads to faster convergence but affects the method's performance, resulting in a higher test loss. Conversely, if a better performance is required, a larger $T_1$ can be selected, corresponding to a more refined polynomial relaxation. In the proposed framework, we can $\textit{flexibly}$ adjust $T_1$ based on distributed system requirements. The results in Figures \ref{fig:T1_test_loss} and \ref{fig:T1_robust_test_loss} are consistent with our theoretical analyses presented under Theorems \ref{theorem:1} and \ref{theorem:2}.

\renewcommand\arraystretch{1.1}
\renewcommand\tabcolsep{20pt}
\begin{table*}[t]
\centering
\renewcommand{\thetable}{\arabic{table}}
\caption{Comparisons between the proposed DTZO and the state-of-the-art methods. Experiments are repeated five times and higher scores represent better performance.}
{
\scalebox{1}{
\begin{tabular}{l|c|c|c}
\toprule
 Dataset   & FedZOO \cite{fang2022communication}     &  FedRZO$_{\rm{bl}}$ \cite{qiu2023zeroth} & \textbf{DTZO}  \\ \hline

MNIST    & 52.89 ± 0.49 $\%$  & 54.05 ± 0.81 $\%$   &  \textbf{79.27} ± \textbf{0.19} $\%$ \\

QMNIST    &  52.45 ± 0.88 $\%$  &  54.67 ± 0.65 $\%$   &  \textbf{78.04} ± \textbf{0.37} $\%$ \\

F-MNIST      &  48.74 ± 0.61 $\%$  & 50.23 ± 0.49 $\%$   &  \textbf{70.07} ± \textbf{0.45} $\%$ \\

USPS    &   72.77 ± 0.43 $\%$  & 73.79 ± 0.56 $\%$   &  \textbf{85.13} ± \textbf{0.14} $\%$ \\
\bottomrule  
\end{tabular}}
\label{tab:experiments_nondiffe}}
\end{table*}

\begin{figure}[t]
    \centering
    \begin{minipage}{0.48\textwidth}
        \centering
\includegraphics[width=0.75\linewidth]{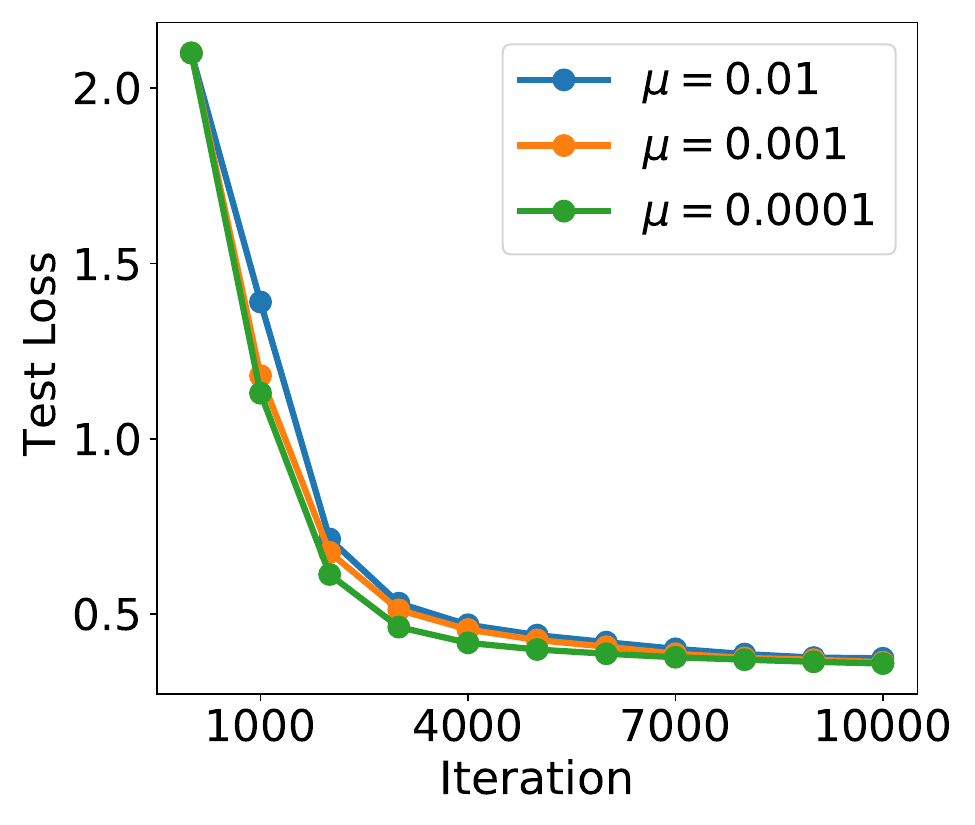}
    \caption{{Test loss of the proposed DTZO under various setting of smoothing parameter $\mu$, results on USPS dataset.}}
    \label{fig:u_test_loss}
    \end{minipage}\hfill
    \begin{minipage}{0.48\textwidth}
        \centering
\includegraphics[width=0.75\linewidth]{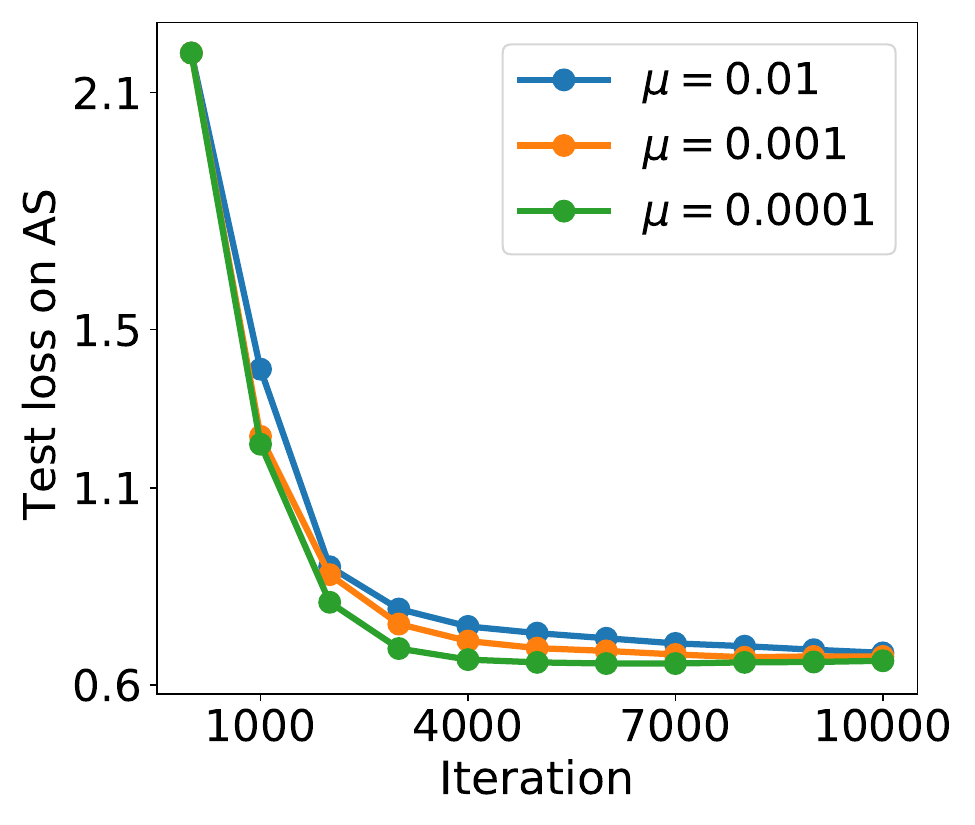}
    \caption{{Test loss on AS (adversarial samples) of DTZO under various setting of smoothing parameter $\mu$, results on USPS dataset.}}
\label{fig:u_robust_test_loss}
    \end{minipage}
\end{figure}

Following \cite{qiu2023zeroth}, the robustness in the proposed framework with respect to the choice of smoothing parameter $\mu$ is evaluated. The experiments are conducted on the robust hyperparameter optimization task under various setting of smoothing parameter, $\mu \in \{0.01, 0.001, 0.0001\}$. It is seen from Figure \ref{fig:u_test_loss} and \ref{fig:u_robust_test_loss} that the proposed DTZO is robust to the choice of smoothing parameter $\mu$. In addition, we also note that the proposed DTZO has faster convergence rate with a relatively smaller $\mu$, because the gradient estimate improves when $\mu$ becomes relatively smaller, as discussed in \cite{liu2020primer}.

\section{Conclusion}
In this work, a distributed trilevel zeroth order learning (DTZO) framework is proposed to address the trilevel learning problems in a distributed manner without using first order information. To our best knowledge, this is the first work that considers how to tackle the trilevel zeroth order learning problems. The proposed DTZO is capable of constructing the cascaded polynomial approximation for trilevel zeroth order learning problems without using gradients or sub-gradients by utilizing the novel zeroth order cuts. Additionally, we theoretically analyze the non-asymptotic convergence rate for the proposed DTZO to achieve the $\epsilon$-stationary point. Experiments on black-box LLMs trilevel learning and robust hyperparameter optimization demonstrate the superior performance of DTZO.
\bibliographystyle{unsrt}  
\bibliography{ref}  

\begin{thebibliography}{100}

\bibitem{sato2021gradient}
Ryo Sato, Mirai Tanaka, and Akiko Takeda.
\newblock A gradient method for multilevel optimization.
\newblock {\em Advances in Neural Information Processing Systems}, 34:7522--7533, 2021.

\bibitem{choe2022betty}
Sang~Keun Choe, Willie Neiswanger, Pengtao Xie, and Eric Xing.
\newblock Betty: An automatic differentiation library for multilevel optimization.
\newblock In {\em The Eleventh International Conference on Learning Representations}, 2023.

\bibitem{guo2020meets}
Minghao Guo, Yuzhe Yang, Rui Xu, Ziwei Liu, and Dahua Lin.
\newblock When nas meets robustness: In search of robust architectures against adversarial attacks.
\newblock In {\em Proceedings of the IEEE/CVF Conference on Computer Vision and Pattern Recognition}, pages 631--640, 2020.

\bibitem{jiao2024provably}
Yang Jiao, Kai Yang, Tiancheng Wu, Chengtao Jian, and Jianwei Huang.
\newblock Provably convergent federated trilevel learning.
\newblock In {\em Proceedings of the AAAI Conference on Artificial Intelligence}, volume~38, pages 12928--12937, 2024.

\bibitem{blair1992computational}
Charles Blair.
\newblock The computational complexity of multi-level linear programs.
\newblock {\em Annals of Operations Research}, 34, 1992.

\bibitem{avraamidou2018mixed}
Styliani Avraamidou.
\newblock Mixed-integer multi-level optimization through multi-parametric programming.
\newblock 2018.

\bibitem{ben1990computational}
Omar Ben-Ayed and Charles~E Blair.
\newblock Computational difficulties of bilevel linear programming.
\newblock {\em Operations Research}, 38(3):556--560, 1990.

\bibitem{sinha2017review}
Ankur Sinha, Pekka Malo, and Kalyanmoy Deb.
\newblock A review on bilevel optimization: From classical to evolutionary approaches and applications.
\newblock {\em IEEE Transactions on Evolutionary Computation}, 22(2):276--295, 2017.

\bibitem{fang2022communication}
Wenzhi Fang, Ziyi Yu, Yuning Jiang, Yuanming Shi, Colin~N Jones, and Yong Zhou.
\newblock Communication-efficient stochastic zeroth-order optimization for federated learning.
\newblock {\em IEEE Transactions on Signal Processing}, 70:5058--5073, 2022.

\bibitem{qiu2023zeroth}
Yuyang Qiu, Uday Shanbhag, and Farzad Yousefian.
\newblock Zeroth-order methods for nondifferentiable, nonconvex, and hierarchical federated optimization.
\newblock {\em Advances in Neural Information Processing Systems}, 36, 2023.

\bibitem{liu2018zeroth}
Sijia Liu, Bhavya Kailkhura, Pin-Yu Chen, Paishun Ting, Shiyu Chang, and Lisa Amini.
\newblock Zeroth-order stochastic variance reduction for nonconvex optimization.
\newblock {\em Advances in Neural Information Processing Systems}, 31, 2018.

\bibitem{chen2019zo}
Xiangyi Chen, Sijia Liu, Kaidi Xu, Xingguo Li, Xue Lin, Mingyi Hong, and David Cox.
\newblock Zo-adamm: Zeroth-order adaptive momentum method for black-box optimization.
\newblock {\em Advances in neural information processing systems}, 32, 2019.

\bibitem{wang2018stochastic}
Yining Wang, Simon Du, Sivaraman Balakrishnan, and Aarti Singh.
\newblock Stochastic zeroth-order optimization in high dimensions.
\newblock In {\em International conference on artificial intelligence and statistics}, pages 1356--1365. PMLR, 2018.

\bibitem{chen2017zoo}
Pin-Yu Chen, Huan Zhang, Yash Sharma, Jinfeng Yi, and Cho-Jui Hsieh.
\newblock Zoo: Zeroth order optimization based black-box attacks to deep neural networks without training substitute models.
\newblock In {\em Proceedings of the 10th ACM workshop on artificial intelligence and security}, pages 15--26, 2017.

\bibitem{heliou2021zeroth}
Am{\'e}lie H{\'e}liou, Matthieu Martin, Panayotis Mertikopoulos, and Thibaud Rahier.
\newblock Zeroth-order non-convex learning via hierarchical dual averaging.
\newblock In {\em International Conference on Machine Learning}, pages 4192--4202. PMLR, 2021.

\bibitem{cai2021zeroth}
HanQin Cai, Yuchen Lou, Daniel McKenzie, and Wotao Yin.
\newblock A zeroth-order block coordinate descent algorithm for huge-scale black-box optimization.
\newblock In {\em International Conference on Machine Learning}, pages 1193--1203. PMLR, 2021.

\bibitem{gao2020can}
Hongchang Gao and Heng Huang.
\newblock Can stochastic zeroth-order frank-wolfe method converge faster for non-convex problems?
\newblock In {\em International conference on machine learning}, pages 3377--3386. PMLR, 2020.

\bibitem{yue2023zeroth}
Pengyun Yue, Long Yang, Cong Fang, and Zhouchen Lin.
\newblock Zeroth-order optimization with weak dimension dependency.
\newblock In {\em The Thirty Sixth Annual Conference on Learning Theory}, pages 4429--4472. PMLR, 2023.

\bibitem{li2022zeroth}
Zichong Li, Pin-Yu Chen, Sijia Liu, Songtao Lu, and Yangyang Xu.
\newblock Zeroth-order optimization for composite problems with functional constraints.
\newblock In {\em Proceedings of the AAAI Conference on Artificial Intelligence}, volume~36, pages 7453--7461, 2022.

\bibitem{ren2023escaping}
Zhaolin Ren, Yujie Tang, and Na~Li.
\newblock Escaping saddle points in zeroth-order optimization: the power of two-point estimators.
\newblock In {\em International Conference on Machine Learning}, pages 28914--28975. PMLR, 2023.

\bibitem{nikolakakis2022black}
Konstantinos Nikolakakis, Farzin Haddadpour, Dionysis Kalogerias, and Amin Karbasi.
\newblock Black-box generalization: Stability of zeroth-order learning.
\newblock {\em Advances in Neural Information Processing Systems}, 35:31525--31541, 2022.

\bibitem{tu2019autozoom}
Chun-Chen Tu, Paishun Ting, Pin-Yu Chen, Sijia Liu, Huan Zhang, Jinfeng Yi, Cho-Jui Hsieh, and Shin-Ming Cheng.
\newblock Autozoom: Autoencoder-based zeroth order optimization method for attacking black-box neural networks.
\newblock In {\em Proceedings of the AAAI conference on artificial intelligence}, volume~33, pages 742--749, 2019.

\bibitem{rando2024optimal}
Marco Rando, Cesare Molinari, Lorenzo Rosasco, and Silvia Villa.
\newblock An optimal structured zeroth-order algorithm for non-smooth optimization.
\newblock {\em Advances in Neural Information Processing Systems}, 36, 2024.

\bibitem{lian2016comprehensive}
Xiangru Lian, Huan Zhang, Cho-Jui Hsieh, Yijun Huang, and Ji~Liu.
\newblock A comprehensive linear speedup analysis for asynchronous stochastic parallel optimization from zeroth-order to first-order.
\newblock {\em Advances in Neural Information Processing Systems}, 29, 2016.

\bibitem{tang2020distributed}
Yujie Tang, Junshan Zhang, and Na~Li.
\newblock Distributed zero-order algorithms for nonconvex multiagent optimization.
\newblock {\em IEEE Transactions on Control of Network Systems}, 8(1):269--281, 2020.

\bibitem{chen2024fine}
Jun Chen, Hong Chen, Bin Gu, and Hao Deng.
\newblock Fine-grained theoretical analysis of federated zeroth-order optimization.
\newblock {\em Advances in Neural Information Processing Systems}, 36, 2024.

\bibitem{akhavan2021distributed}
Arya Akhavan, Massimiliano Pontil, and Alexandre Tsybakov.
\newblock Distributed zero-order optimization under adversarial noise.
\newblock {\em Advances in Neural Information Processing Systems}, 34:10209--10220, 2021.

\bibitem{sahu2018distributed}
Anit~Kumar Sahu, Dusan Jakovetic, Dragana Bajovic, and Soummya Kar.
\newblock Distributed zeroth order optimization over random networks: A kiefer-wolfowitz stochastic approximation approach.
\newblock In {\em 2018 IEEE Conference on Decision and Control (CDC)}, pages 4951--4958. IEEE, 2018.

\bibitem{shu2023federated}
Yao Shu, Xiaoqiang Lin, Zhongxiang Dai, and Bryan Kian~Hsiang Low.
\newblock Federated zeroth-order optimization using trajectory-informed surrogate gradients.
\newblock {\em arXiv preprint arXiv:2308.04077}, 2023.

\bibitem{raghu2021meta}
Aniruddh Raghu, Jonathan Lorraine, Simon Kornblith, Matthew McDermott, and David~K Duvenaud.
\newblock Meta-learning to improve pre-training.
\newblock {\em Advances in Neural Information Processing Systems}, 34:23231--23244, 2021.

\bibitem{garg2022learning}
Bhanu Garg, Li~Zhang, Pradyumna Sridhara, Ramtin Hosseini, Eric Xing, and Pengtao Xie.
\newblock Learning from mistakes--a framework for neural architecture search.
\newblock In {\em Proceedings of the AAAI Conference on Artificial Intelligence}, volume~36, pages 10184--10192, 2022.

\bibitem{bertsekas2015convex}
Dimitri Bertsekas.
\newblock {\em Convex optimization algorithms}.
\newblock Athena Scientific, 2015.

\bibitem{franc2011cutting}
Vojtech Franc, S{\"o}ren Sonnenburg, and Tom{\'a}{\v{s}} Werner.
\newblock Cutting plane methods in machine learning.
\newblock {\em Optimization for Machine Learning}, pages 185--218, 2011.

\bibitem{yang2014distributed}
Kai Yang, Jianwei Huang, Yihong Wu, Xiaodong Wang, and Mung Chiang.
\newblock Distributed robust optimization ({DRO}), part {I}: {Framework} and example.
\newblock {\em Optimization and Engineering}, 15(1):35--67, 2014.

\bibitem{burger2013polyhedral}
Mathias B{\"u}rger, Giuseppe Notarstefano, and Frank Allg{\"o}wer.
\newblock A polyhedral approximation framework for convex and robust distributed optimization.
\newblock {\em IEEE Transactions on Automatic Control}, 59(2):384--395, 2013.

\bibitem{jiao2022asynchronous}
Yang Jiao, Kai Yang, Tiancheng Wu, Dongjin Song, and Chengtao Jian.
\newblock Asynchronous distributed bilevel optimization.
\newblock In {\em The Eleventh International Conference on Learning Representations}, 2023.

\bibitem{chen2024robust}
Xingdi Chen, Yu~Xiong, and Kai Yang.
\newblock Robust beamforming for downlink multi-cell systems: A bilevel optimization perspective.
\newblock In {\em Proceedings of the AAAI Conference on Artificial Intelligence}, 2024.

\bibitem{subramanya2021centralized}
Tejas Subramanya and Roberto Riggio.
\newblock Centralized and federated learning for predictive vnf autoscaling in multi-domain 5g networks and beyond.
\newblock {\em IEEE Transactions on Network and Service Management}, 18(1):63--78, 2021.

\bibitem{assran2020advances}
Mahmoud Assran, Arda Aytekin, Hamid~Reza Feyzmahdavian, Mikael Johansson, and Michael~G Rabbat.
\newblock Advances in asynchronous parallel and distributed optimization.
\newblock {\em Proceedings of the IEEE}, 108(11):2013--2031, 2020.

\bibitem{pan2024scalebio}
Rui Pan, Jipeng Zhang, Xingyuan Pan, Renjie Pi, Xiaoyu Wang, and Tong Zhang.
\newblock Scalebio: Scalable bilevel optimization for llm data reweighting.
\newblock {\em arXiv preprint arXiv:2406.19976}, 2024.

\bibitem{kwon2023fully}
Jeongyeol Kwon, Dohyun Kwon, Stephen Wright, and Robert~D Nowak.
\newblock A fully first-order method for stochastic bilevel optimization.
\newblock In {\em International Conference on Machine Learning}, pages 18083--18113. PMLR, 2023.

\bibitem{jiang2023conditional}
Ruichen Jiang, Nazanin Abolfazli, Aryan Mokhtari, and Erfan~Yazdandoost Hamedani.
\newblock A conditional gradient-based method for simple bilevel optimization with convex lower-level problem.
\newblock In {\em International Conference on Artificial Intelligence and Statistics}, pages 10305--10323. PMLR, 2023.

\bibitem{liu2018darts}
Hanxiao Liu, Karen Simonyan, and Yiming Yang.
\newblock Darts: Differentiable architecture search.
\newblock In {\em International Conference on Learning Representations}, 2018.

\bibitem{kautz1996general}
Henry~A Kautz, Bart Selman, and Yueyen Jiang.
\newblock A general stochastic approach to solving problems with hard and soft constraints.
\newblock {\em Satisfiability Problem: Theory and Applications}, 35:573--586, 1996.

\bibitem{kornowski2024algorithm}
Guy Kornowski and Ohad Shamir.
\newblock An algorithm with optimal dimension-dependence for zero-order nonsmooth nonconvex stochastic optimization.
\newblock {\em Journal of Machine Learning Research}, 25(122):1--14, 2024.

\bibitem{ghadimi2013stochastic}
Saeed Ghadimi and Guanghui Lan.
\newblock Stochastic first-and zeroth-order methods for nonconvex stochastic programming.
\newblock {\em SIAM journal on optimization}, 23(4):2341--2368, 2013.

\bibitem{shen2023penalty}
Han Shen and Tianyi Chen.
\newblock On penalty-based bilevel gradient descent method.
\newblock {\em arXiv preprint arXiv:2302.05185}, 2023.

\bibitem{shi2021improved}
Wanli Shi and Bin Gu.
\newblock Improved penalty method via doubly stochastic gradients for bilevel hyperparameter optimization.
\newblock In {\em Proceedings of the AAAI Conference on Artificial Intelligence}, volume~35, pages 9621--9629, 2021.

\bibitem{boyd2004convex}
Stephen~P Boyd and Lieven Vandenberghe.
\newblock {\em Convex optimization}.
\newblock Cambridge university press, 2004.

\bibitem{xu2020unified}
Zi~Xu, Huiling Zhang, Yang Xu, and Guanghui Lan.
\newblock A unified single-loop alternating gradient projection algorithm for nonconvex-concave and convex-nonconcave minimax problems.
\newblock {\em arXiv preprint arXiv:2006.02032}, 2020.

\bibitem{shen2019complexities}
Zebang Shen, Cong Fang, Peilin Zhao, Junzhou Huang, and Hui Qian.
\newblock Complexities in projection-free stochastic non-convex minimization.
\newblock In {\em The 22nd International Conference on Artificial Intelligence and Statistics}, pages 2868--2876. PMLR, 2019.

\bibitem{huang2024enhancing}
Hanbo Huang, Yihan Li, Bowen Jiang, Bo~Jiang, Lin Liu, Zhuotao Liu, Ruoyu Sun, and Shiyu Liang.
\newblock Enhancing the resilience of llms against grey-box extractions.
\newblock In {\em ICML 2024 Next Generation of AI Safety Workshop}, 2024.

\bibitem{nesterov2017random}
Yurii Nesterov and Vladimir Spokoiny.
\newblock Random gradient-free minimization of convex functions.
\newblock {\em Foundations of Computational Mathematics}, 17(2):527--566, 2017.

\bibitem{kornilov2024accelerated}
Nikita Kornilov, Ohad Shamir, Aleksandr Lobanov, Darina Dvinskikh, Alexander Gasnikov, Innokentiy Shibaev, Eduard Gorbunov, and Samuel Horv{\'a}th.
\newblock Accelerated zeroth-order method for non-smooth stochastic convex optimization problem with infinite variance.
\newblock {\em Advances in Neural Information Processing Systems}, 36, 2024.

\bibitem{deng2020distributionally}
Yuyang Deng, Mohammad~Mahdi Kamani, and Mehrdad Mahdavi.
\newblock Distributionally robust federated averaging.
\newblock {\em Advances in neural information processing systems}, 33:15111--15122, 2020.

\bibitem{qian2019robust}
Qi~Qian, Shenghuo Zhu, Jiasheng Tang, Rong Jin, Baigui Sun, and Hao Li.
\newblock Robust optimization over multiple domains.
\newblock In {\em Proceedings of the AAAI Conference on Artificial Intelligence}, volume~33, pages 4739--4746, 2019.

\bibitem{lei2018stochastic}
Yunwen Lei and Ke~Tang.
\newblock Stochastic composite mirror descent: Optimal bounds with high probabilities.
\newblock {\em Advances in Neural Information Processing Systems}, 31, 2018.

\bibitem{zheng2017asynchronous}
Shuxin Zheng, Qi~Meng, Taifeng Wang, Wei Chen, Nenghai Yu, Zhi-Ming Ma, and Tie-Yan Liu.
\newblock Asynchronous stochastic gradient descent with delay compensation.
\newblock In {\em International conference on machine learning}, pages 4120--4129. PMLR, 2017.

\bibitem{cutkosky2019momentum}
Ashok Cutkosky and Francesco Orabona.
\newblock Momentum-based variance reduction in non-convex sgd.
\newblock {\em Advances in neural information processing systems}, 32, 2019.

\bibitem{liu2021conflict}
Bo~Liu, Xingchao Liu, Xiaojie Jin, Peter Stone, and Qiang Liu.
\newblock Conflict-averse gradient descent for multi-task learning.
\newblock {\em Advances in Neural Information Processing Systems}, 34:18878--18890, 2021.

\bibitem{shaban2019truncated}
Amirreza Shaban, Ching-An Cheng, Nathan Hatch, and Byron Boots.
\newblock Truncated back-propagation for bilevel optimization.
\newblock In {\em The 22nd International Conference on Artificial Intelligence and Statistics}, pages 1723--1732. PMLR, 2019.

\bibitem{chen2023decentralized}
Xuxing Chen, Minhui Huang, Shiqian Ma, and Krishna Balasubramanian.
\newblock Decentralized stochastic bilevel optimization with improved per-iteration complexity.
\newblock In {\em International Conference on Machine Learning}, pages 4641--4671. PMLR, 2023.

\bibitem{lin2024non}
Sen Lin, Daouda Sow, Kaiyi Ji, Yingbin Liang, and Ness Shroff.
\newblock Non-convex bilevel optimization with time-varying objective functions.
\newblock {\em Advances in Neural Information Processing Systems}, 36, 2024.

\bibitem{zhangrevisiting}
Yihua Zhang, Pingzhi Li, Junyuan Hong, Jiaxiang Li, Yimeng Zhang, Wenqing Zheng, Pin-Yu Chen, Jason~D Lee, Wotao Yin, Mingyi Hong, et~al.
\newblock Revisiting zeroth-order optimization for memory-efficient llm fine-tuning: A benchmark.
\newblock In {\em Forty-first International Conference on Machine Learning}, 2024.

\bibitem{zhang2024subspace}
Haozhen Zhang, Hualin Zhang, Bin Gu, and Yi~Chang.
\newblock Subspace selection based prompt tuning with nonconvex nonsmooth black-box optimization.
\newblock In {\em Proceedings of the 30th ACM SIGKDD Conference on Knowledge Discovery and Data Mining}, pages 4179--4190, 2024.

\bibitem{duchi2015optimal}
John~C Duchi, Michael~I Jordan, Martin~J Wainwright, and Andre Wibisono.
\newblock Optimal rates for zero-order convex optimization: The power of two function evaluations.
\newblock {\em IEEE Transactions on Information Theory}, 61(5):2788--2806, 2015.

\bibitem{sun2022black}
Tianxiang Sun, Yunfan Shao, Hong Qian, Xuanjing Huang, and Xipeng Qiu.
\newblock Black-box tuning for language-model-as-a-service.
\newblock In {\em International Conference on Machine Learning}, pages 20841--20855. PMLR, 2022.

\bibitem{ma2024fairness}
Huan Ma, Changqing Zhang, Yatao Bian, Lemao Liu, Zhirui Zhang, Peilin Zhao, Shu Zhang, Huazhu Fu, Qinghua Hu, and Bingzhe Wu.
\newblock Fairness-guided few-shot prompting for large language models.
\newblock {\em Advances in Neural Information Processing Systems}, 36, 2024.

\bibitem{wang2024grammar}
Bailin Wang, Zi~Wang, Xuezhi Wang, Yuan Cao, Rif A~Saurous, and Yoon Kim.
\newblock Grammar prompting for domain-specific language generation with large language models.
\newblock {\em Advances in Neural Information Processing Systems}, 36, 2024.

\bibitem{diao2022black}
Shizhe Diao, Zhichao Huang, Ruijia Xu, Xuechun Li, LIN Yong, Xiao Zhou, and Tong Zhang.
\newblock Black-box prompt learning for pre-trained language models.
\newblock {\em Transactions on Machine Learning Research}, 2022.

\bibitem{yao2024poisonprompt}
Hongwei Yao, Jian Lou, and Zhan Qin.
\newblock Poisonprompt: Backdoor attack on prompt-based large language models.
\newblock In {\em ICASSP 2024-2024 IEEE International Conference on Acoustics, Speech and Signal Processing (ICASSP)}, pages 7745--7749. IEEE, 2024.

\bibitem{bai2023qwen}
Jinze Bai, Shuai Bai, Yunfei Chu, Zeyu Cui, Kai Dang, Xiaodong Deng, Yang Fan, Wenbin Ge, Yu~Han, Fei Huang, et~al.
\newblock Qwen technical report.
\newblock {\em arXiv preprint arXiv:2309.16609}, 2023.

\bibitem{wang2018glue}
Alex Wang, Amanpreet Singh, Julian Michael, Felix Hill, Omer Levy, and Samuel Bowman.
\newblock Glue: A multi-task benchmark and analysis platform for natural language understanding.
\newblock In {\em Proceedings of the 2018 EMNLP Workshop BlackboxNLP: Analyzing and Interpreting Neural Networks for NLP}, pages 353--355, 2018.

\bibitem{wang2021discriminative}
Jing Wang, Jiahong Chen, Jianzhe Lin, Leonid Sigal, and Clarence~W de~Silva.
\newblock Discriminative feature alignment: Improving transferability of unsupervised domain adaptation by gaussian-guided latent alignment.
\newblock {\em Pattern Recognition}, 116:107943, 2021.

\bibitem{lecun1998gradient}
Yann LeCun, L{\'e}on Bottou, Yoshua Bengio, and Patrick Haffner.
\newblock Gradient-based learning applied to document recognition.
\newblock {\em Proceedings of the IEEE}, 86(11):2278--2324, 1998.

\bibitem{xiao2017fashion}
Han Xiao, Kashif Rasul, and Roland Vollgraf.
\newblock Fashion-mnist: a novel image dataset for benchmarking machine learning algorithms.
\newblock {\em arXiv preprint arXiv:1708.07747}, 2017.

\bibitem{yadav2019cold}
Chhavi Yadav and L{\'e}on Bottou.
\newblock Cold case: The lost mnist digits.
\newblock {\em Advances in neural information processing systems}, 32, 2019.

\bibitem{liu2020primer}
Sijia Liu, Pin-Yu Chen, Bhavya Kailkhura, Gaoyuan Zhang, Alfred~O Hero~III, and Pramod~K Varshney.
\newblock A primer on zeroth-order optimization in signal processing and machine learning: Principals, recent advances, and applications.
\newblock {\em IEEE Signal Processing Magazine}, 37(5):43--54, 2020.

\bibitem{regin2011using}
Jean-Charles R{\'e}gin.
\newblock Using hard constraints for representing soft constraints.
\newblock In {\em Integration of AI and OR Techniques in Constraint Programming for Combinatorial Optimization Problems: 8th International Conference, CPAIOR 2011, Berlin, Germany, May 23-27, 2011. Proceedings 8}, pages 176--189. Springer, 2011.

\bibitem{wilson2022combining}
Ellis Wilson, Frank Mueller, and Scott Pakin.
\newblock Combining hard and soft constraints in quantum constraint-satisfaction systems.
\newblock In {\em SC22: International Conference for High Performance Computing, Networking, Storage and Analysis}, pages 1--14. IEEE, 2022.

\bibitem{ji2021bilevel}
Kaiyi Ji, Junjie Yang, and Yingbin Liang.
\newblock Bilevel optimization: Convergence analysis and enhanced design.
\newblock In {\em International conference on machine learning}, pages 4882--4892. PMLR, 2021.

\bibitem{finn2017model}
Chelsea Finn, Pieter Abbeel, and Sergey Levine.
\newblock Model-agnostic meta-learning for fast adaptation of deep networks.
\newblock In {\em International conference on machine learning}, pages 1126--1135. PMLR, 2017.

\bibitem{madry2018towards}
Aleksander Madry, Aleksandar Makelov, Ludwig Schmidt, Dimitris Tsipras, and Adrian Vladu.
\newblock Towards deep learning models resistant to adversarial attacks.
\newblock In {\em International Conference on Learning Representations}, 2018.

\bibitem{zhang2022revisiting}
Yihua Zhang, Guanhua Zhang, Prashant Khanduri, Mingyi Hong, Shiyu Chang, and Sijia Liu.
\newblock Revisiting and advancing fast adversarial training through the lens of bi-level optimization.
\newblock In {\em International Conference on Machine Learning}, pages 26693--26712. PMLR, 2022.

\bibitem{jiao2022distributed}
Yang Jiao, Kai Yang, and Dongjin Song.
\newblock Distributed distributionally robust optimization with non-convex objectives.
\newblock {\em Advances in neural information processing systems}, 35:7987--7999, 2022.

\bibitem{yang2021provably}
Junjie Yang, Kaiyi Ji, and Yingbin Liang.
\newblock Provably faster algorithms for bilevel optimization.
\newblock {\em Advances in Neural Information Processing Systems}, 34:13670--13682, 2021.

\bibitem{franceschi2018bilevel}
Luca Franceschi, Paolo Frasconi, Saverio Salzo, Riccardo Grazzi, and Massimiliano Pontil.
\newblock Bilevel programming for hyperparameter optimization and meta-learning.
\newblock In {\em International conference on machine learning}, pages 1568--1577. PMLR, 2018.

\bibitem{liu2021investigating}
Risheng Liu, Jiaxin Gao, Jin Zhang, Deyu Meng, and Zhouchen Lin.
\newblock Investigating bi-level optimization for learning and vision from a unified perspective: A survey and beyond.
\newblock {\em IEEE Transactions on Pattern Analysis and Machine Intelligence}, 44(12):10045--10067, 2021.

\bibitem{mackay2018self}
Matthew Mackay, Paul Vicol, Jonathan Lorraine, David Duvenaud, and Roger Grosse.
\newblock Self-tuning networks: Bilevel optimization of hyperparameters using structured best-response functions.
\newblock In {\em International Conference on Learning Representations}, 2018.

\bibitem{jiao2022timeautoad}
Yang Jiao, Kai Yang, Dongjing Song, and Dacheng Tao.
\newblock Timeautoad: Autonomous anomaly detection with self-supervised contrastive loss for multivariate time series.
\newblock {\em IEEE Transactions on Network Science and Engineering}, 9(3):1604--1619, 2022.

\bibitem{han2024fedal}
Pengchao Han, Xingyan Shi, and Jianwei Huang.
\newblock Fedal: Black-box federated knowledge distillation enabled by adversarial learning.
\newblock {\em IEEE Journal on Selected Areas in Communications}, 2024.

\bibitem{clanuwat2018deep}
Tarin Clanuwat, Mikel Bober-Irizar, Asanobu Kitamoto, Alex Lamb, Kazuaki Yamamoto, and David Ha.
\newblock Deep learning for classical japanese literature.
\newblock {\em arXiv preprint arXiv:1812.01718}, 2018.

\bibitem{sra2016adadelay}
Suvrit Sra, Adams~Wei Yu, Mu~Li, and Alex Smola.
\newblock Adadelay: Delay adaptive distributed stochastic optimization.
\newblock In {\em Artificial Intelligence and Statistics}, pages 957--965. PMLR, 2016.

\bibitem{li2021distributed}
Wenjie Li and Mohamad Assaad.
\newblock Distributed zeroth-order stochastic optimization in time-varying networks.
\newblock {\em arXiv preprint arXiv:2105.12597}, 2021.

\bibitem{cao2024projection}
Jincheng Cao, Ruichen Jiang, Nazanin Abolfazli, Erfan Yazdandoost~Hamedani, and Aryan Mokhtari.
\newblock Projection-free methods for stochastic simple bilevel optimization with convex lower-level problem.
\newblock {\em Advances in Neural Information Processing Systems}, 36, 2024.

\bibitem{liu2018signsgd}
Sijia Liu, Pin-Yu Chen, Xiangyi Chen, and Mingyi Hong.
\newblock signsgd via zeroth-order oracle.
\newblock In {\em International Conference on Learning Representations}, 2018.

\bibitem{liu2022bome}
Bo~Liu, Mao Ye, Stephen Wright, Peter Stone, and Qiang Liu.
\newblock Bome! bilevel optimization made easy: A simple first-order approach.
\newblock {\em Advances in Neural Information Processing Systems}, 35:17248--17262, 2022.

\bibitem{liang2024novel}
Huiping Liang, Bei Sun, Biao Huang, Yonggang Li, and Chunhua Yang.
\newblock A novel chattering-free discrete sliding mode controller with disturbance compensation for zinc roasting temperature distribution control.
\newblock {\em IEEE Transactions on Automation Science and Engineering}, 2024.

\bibitem{gao2024decentralized}
Hongchang Gao.
\newblock Decentralized multi-level compositional optimization algorithms with level-independent convergence rate.
\newblock In {\em International Conference on Artificial Intelligence and Statistics}, pages 4402--4410. PMLR, 2024.

\bibitem{gao2022convergence}
Hongchang Gao, Junyi Li, and Heng Huang.
\newblock On the convergence of local stochastic compositional gradient descent with momentum.
\newblock In {\em International Conference on Machine Learning}, pages 7017--7035. PMLR, 2022.

\bibitem{chen2023optimal}
Xuxing Chen, Tesi Xiao, and Krishnakumar Balasubramanian.
\newblock Optimal algorithms for stochastic bilevel optimization under relaxed smoothness conditions.
\newblock {\em arXiv preprint arXiv:2306.12067}, 2023.

\bibitem{li2024communication}
Junyi Li, Feihu Huang, and Heng Huang.
\newblock Communication-efficient federated bilevel optimization with global and local lower level problems.
\newblock {\em Advances in Neural Information Processing Systems}, 36, 2024.

\bibitem{wu2024federated}
Xidong Wu, Jianhui Sun, Zhengmian Hu, Junyi Li, Aidong Zhang, and Heng Huang.
\newblock Federated conditional stochastic optimization.
\newblock {\em Advances in Neural Information Processing Systems}, 36, 2024.

\bibitem{huang2024nonconvex}
Feihu Huang, Shangqian Gao, Jian Pei, and Heng Huang.
\newblock Nonconvex zeroth-order stochastic admm methods with lower function query complexity.
\newblock {\em IEEE Transactions on Pattern Analysis and Machine Intelligence}, 2024.

\bibitem{jing2024asynchronous}
Gangshan Jing, He~Bai, Jemin George, Aranya Chakrabortty, and Piyush~K Sharma.
\newblock Asynchronous distributed reinforcement learning for lqr control via zeroth-order block coordinate descent.
\newblock {\em IEEE Transactions on Automatic Control}, 2024.

\bibitem{chen2024optimal}
Xuxing Chen, Tesi Xiao, and Krishnakumar Balasubramanian.
\newblock Optimal algorithms for stochastic bilevel optimization under relaxed smoothness conditions.
\newblock {\em Journal of Machine Learning Research}, 25(151):1--51, 2024.

\bibitem{xiao2023alternating}
Quan Xiao, Han Shen, Wotao Yin, and Tianyi Chen.
\newblock Alternating projected sgd for equality-constrained bilevel optimization.
\newblock In {\em International Conference on Artificial Intelligence and Statistics}, pages 987--1023. PMLR, 2023.

\bibitem{hong2023two}
Mingyi Hong, Hoi-To Wai, Zhaoran Wang, and Zhuoran Yang.
\newblock A two-timescale stochastic algorithm framework for bilevel optimization: Complexity analysis and application to actor-critic.
\newblock {\em SIAM Journal on Optimization}, 33(1):147--180, 2023.

\bibitem{garber2015faster}
Dan Garber and Elad Hazan.
\newblock Faster rates for the frank-wolfe method over strongly-convex sets.
\newblock In {\em International Conference on Machine Learning}, pages 541--549. PMLR, 2015.

\bibitem{zhang2020one}
Mingrui Zhang, Zebang Shen, Aryan Mokhtari, Hamed Hassani, and Amin Karbasi.
\newblock One sample stochastic frank-wolfe.
\newblock In {\em International Conference on Artificial Intelligence and Statistics}, pages 4012--4023. PMLR, 2020.

\bibitem{xian2021communication}
Wenhan Xian, Feihu Huang, and Heng Huang.
\newblock Communication-efficient frank-wolfe algorithm for nonconvex decentralized distributed learning.
\newblock In {\em Proceedings of the AAAI Conference on Artificial Intelligence}, volume~35, pages 10405--10413, 2021.

\bibitem{wang2016parallel}
Yu-Xiang Wang, Veeranjaneyulu Sadhanala, Wei Dai, Willie Neiswanger, Suvrit Sra, and Eric Xing.
\newblock Parallel and distributed block-coordinate frank-wolfe algorithms.
\newblock In {\em International Conference on Machine Learning}, pages 1548--1557. PMLR, 2016.

\bibitem{balashov2020gradient}
MV~Balashov, BT~Polyak, and AA~Tremba.
\newblock Gradient projection and conditional gradient methods for constrained nonconvex minimization.
\newblock {\em Numerical Functional Analysis and Optimization}, 41(7):822--849, 2020.

\bibitem{shen2024principled}
Han Shen, Zhuoran Yang, and Tianyi Chen.
\newblock Principled penalty-based methods for bilevel reinforcement learning and rlhf.
\newblock {\em arXiv preprint arXiv:2402.06886}, 2024.

\bibitem{beykal2020domino}
Burcu Beykal, Styliani Avraamidou, Ioannis~PE Pistikopoulos, Melis Onel, and Efstratios~N Pistikopoulos.
\newblock Domino: Data-driven optimization of bi-level mixed-integer nonlinear problems.
\newblock {\em Journal of Global Optimization}, 78:1--36, 2020.

\bibitem{astudillo2021thinking}
Raul Astudillo and Peter~I Frazier.
\newblock Thinking inside the box: A tutorial on grey-box bayesian optimization.
\newblock In {\em 2021 Winter Simulation Conference (WSC)}, pages 1--15. IEEE, 2021.

\bibitem{bajaj2018trust}
Ishan Bajaj, Shachit~S Iyer, and MM~Faruque Hasan.
\newblock A trust region-based two phase algorithm for constrained black-box and grey-box optimization with infeasible initial point.
\newblock {\em Computers \& Chemical Engineering}, 116:306--321, 2018.

\bibitem{boyd2007localization}
Stephen Boyd and Lieven Vandenberghe.
\newblock Localization and cutting-plane methods.
\newblock {\em From Stanford EE 364b lecture notes}, 386, 2007.

\end{thebibliography}

\newpage
\appendix

\textbf{Appendix}

To improve the readability of the Appendix, we have organized its contents as follows: In Appendix \ref{appendix:iteration_complexity} and \ref{appendix:communication_complexity}, we delve into the comprehensive proofs of Theorem \ref{theorem:1} (Iteration Complexity) and Theorem \ref{theorem:2} (Communication Complexity). In Appendix \ref{appendix:pro1_2}, the detailed proofs of Propositions \ref{prop:1} and \ref{prop:2} are provided. Furthermore, we offer the theoretical analyses about the cascaded polynomial approximation in Appendix \ref{appendix:cascaded}. Additionally, detailed discussions about the soft constraint are given in Appendix \ref{appedix:phi}, and the discussions about $\phi_{\rm{in}}$ and $\phi_{\rm{out}}$ are also conducted in this part. In Appendix \ref{appendix:experiment}, details of the experimental setting and additional experimental results are provided. The discussions about Assumptions \ref{assum:1} and \ref{assum:2} are offered in Appendix \ref{appendix:assum}, we show that both Assumptions \ref{assum:1} and \ref{assum:2} are mild and widely-used in machine learning. In Appendix \ref{appendix:penalty}, the reasons why we choose the exterior penalty method in the proposed framework are discussed, and we demonstrate the close relationship between the original constrained optimization problem and the unconstrained optimization problem. In Appendix \ref{appendix:partial zeroth order}, we show that the proposed framework can be applied to a wide range of TLL problems, e.g., (grey-box) TLL with partial zeroth order constraints. {More discussions about the cutting plane method and the choice of gradient estimator are provided in Appendix \ref{appendix:discussion}.} Lastly, the future work is discussed in Appendix \ref{appendix:limitation}.

Furthermore, to enhance the readability of this work, the notations used in this work and their corresponding meanings are summarized in Table \ref{tab:notation}.

\vspace{10mm}

\textbf{Table of Contents}

$\quad$\ref{appendix:iteration_complexity}. Proof of Theorem \ref{theorem:1} (Iteration Complexity)

$\quad$\ref{appendix:communication_complexity}. Proof of Theorem \ref{theorem:2} (Communication Complexity)

$\quad$\ref{appendix:pro1_2}. Proofs of Proposition \ref{prop:1} and \ref{prop:2}

$\quad$\ref{appendix:cascaded}. Theoretical Analyses about the Cascaded Polynomial Approximation

$\quad$\ref{appedix:phi}. Discussion about Soft Constraint and $\phi_{\rm{in}}$, $\phi_{\rm{out}}$

$\quad$\ref{appendix:experiment}. Experiments

$\quad$\ref{appendix:assum}. Discussion about Assumption \ref{assum:1} and \ref{assum:2}

$\quad$\ref{appendix:penalty}. Exterior Penalty Method

$\quad$\ref{appendix:partial zeroth order}. TLL with Partial Zeroth Order Constraints

{$\quad$\ref{appendix:discussion}. Discussions}

$\quad$\ref{appendix:limitation}. Future Work

\vspace{5mm}

\renewcommand\arraystretch{1.25}
\renewcommand\tabcolsep{8pt}
\begin{table*}[h]
\centering
\renewcommand{\thetable}{\arabic{table}}
\caption{Notations used in this work and the corresponding meanings.}
{
\scalebox{0.90}{
\begin{tabular}{l|c}
\toprule
 Notation   & Meaning \\ \hline
 
  $f_{i}(\cdot), \forall i=1,2,3$  & $i^{\rm{th}}$ level objective.\\
  $\boldsymbol{x}_i, \forall i=1,2,3$  & $i^{\rm{th}}$ level variable. \\ 

  $f_{i,j}(\cdot), \forall i=1,2,3, j=1,\!\cdots\!,N$   & $i^{\rm{th}}$ level local objective in worker $j$. \\
  
  $\boldsymbol{x}_{i,j}, \forall i=1,2,3, j=1,\!\cdots\!,N$  &  $i^{\rm{th}}$ level local variable in worker $j$.\\

  $\boldsymbol{z}_{i}, \forall i=1,2,3$  &  $i^{\rm{th}}$ level global variable in master. \\

  $P_{\rm{in}}$, $P_{\rm{out}}$  &  feasible regions formed by inner and outer layer zeroth order cuts.\\

  $cp_{{\rm{in}},l}, cp_{{\rm{out}},l}$ & $l^{\rm{th}}$ inner layer and outer layer zeroth order cuts.\\
  
${\boldsymbol{a}_{j,l}^{\rm{in}}}$, ${\boldsymbol{b}_{j,l}^{\rm{in}}} $, $\boldsymbol{c}_{i,l}^{\rm{in}}$, $\boldsymbol{d}_{i,l}^{\rm{in}}$, $e_{l}^{\rm{in}}$  & $l^{\rm{th}}$ inner layer zeroth order cut's parameters. \\

${\boldsymbol{a}_{i,j,l}^{\rm{out}}} $, ${\boldsymbol{b}_{i,j,l}^{\rm{out}}} $, $\boldsymbol{c}_{i,l}^{\rm{out}} $, $\boldsymbol{d}_{i,l}^{\rm{out}} $, $e_{l}^{\rm{out}}$  & $l^{\rm{th}}$ outer layer zeroth order cut's parameters.\\

$F(\cdot)$  & penalty function.\\

$F_{\mu}(\cdot)$  & smooth approximation of $F(\cdot)$.\\

  $\mu$  & smoothing parameter.\\

  ${F_\mu }^*$  &  optimal objective value of $F_{\mu}(\cdot)$.\\

$\lambda_l, \phi_j$  & penalty parameters.\\

  $\phi_{\rm{in}}(\cdot), \phi_{\rm{out}}(\cdot)$  & functions used in third level and second level constraint.\\

$G_{\boldsymbol{x}_{i,j}}, \forall i=1,2,3, j=1,\!\cdots\!,N$ & gradient estimator for $i^{\rm{th}}$ level variable in worker $j$ \\

$\eta_{\boldsymbol{x}_i}, \eta_{\boldsymbol{z}_i}, \forall i=1,2,3 $ & step sizes for variables $\boldsymbol{x}_i$, $\boldsymbol{z}_i$. \\

$\boldsymbol{\mu}^{\rm{in}}, \boldsymbol{\mu}^{\rm{out}}, {\boldsymbol{u}_{k,1}},{\boldsymbol{u}_{k,2}},{\boldsymbol{u}_{k,3}}$ & standard Gaussian random vectors.\\

${{\cal G}^t}$  & stationarity gap. \\

$T(\epsilon)$  & iteration complexity to achieve $\epsilon$-stationary point. \\

$T_1$  &  parameter controls the trade-off between complexity and performance. \\

$\mathcal{T}$  &  zeroth order cuts will be updated every $\mathcal{T}$ iteration.  \\

  $N$  & the number of workers in distributed systems.\\
  $L$  & parameter in $L$-smoothness.\\
$d_i, \forall i=1,2,3$  & the dimension of $i^{\rm{th}}$ level variable.
  \\
\bottomrule  
\end{tabular}}
\label{tab:notation}}
\end{table*}











\newpage

\section{Proof of Theorem \ref{theorem:1}}
\label{appendix:iteration_complexity}
In this section, the detailed proofs of Theorem \ref{theorem:1}, i.e., iteration complexity of the proposed DTZO, are offered. The iteration complexity refers to the number of iterations for the proposed algorithm to obtain the $\epsilon$-stationary point \cite{jiao2022asynchronous}. According to \cite{ghadimi2013stochastic}, the gradient of the smooth approximation of $F$, i.e., ${F_\mu }$ (which is given in Definition \ref{def:3}), is also Lipschitz continuous with constant $L_{\mu}$ ($ 0<L_{\mu}\le L$), thus, we have that when $t\ge T_1$,
\begin{equation}
\label{eq:4_19_3}
    \begin{array}{l}
{F_\mu }(\{{\boldsymbol{x}_{i,j}^{t+1}}\},\{ \boldsymbol{z}_i^t\})\\
\! \le \! {F_\mu }(\{{\boldsymbol{x}_{i,j}^t}\},\{ \boldsymbol{z}_i^t\}) \!+\! {\left[ \begin{array}{l}
\!\{\boldsymbol{x}_{1,j}^{t + 1} \!-\! \boldsymbol{x}_{1,j}^t\}\!\\
\!\{\boldsymbol{x}_{2,j}^{t + 1} \!-\! \boldsymbol{x}_{2,j}^t\}\!\\
\!\{\boldsymbol{x}_{3,j}^{t + 1} \!-\! \boldsymbol{x}_{3,j}^t\}\!\\
\end{array} \right]^{\top}}\!\left[ \begin{array}{l}
\!\{ {\nabla _{{\boldsymbol{x}_{1,j}}}}{F_\mu }(\{{\boldsymbol{x}_{i,j}^t}\},\{ \boldsymbol{z}_i^t\}) \}\\
\!\{ {\nabla _{{\boldsymbol{x}_{2,j}}}}{F_\mu }(\{{\boldsymbol{x}_{i,j}^t}\},\{ \boldsymbol{z}_i^t\}) \}\!\\
\!\{ {\nabla _{{\boldsymbol{x}_{3,j}}}}{F_\mu }(\{{\boldsymbol{x}_{i,j}^t}\},\{ \boldsymbol{z}_i^t\}) \}
\end{array} \right] \!+\! \frac{L}{2}||\!\left[\begin{array}{l}
\!\{\boldsymbol{x}_{1,j}^{t + 1} \!-\! \boldsymbol{x}_{1,j}^t\}\!\\
\!\{\boldsymbol{x}_{2,j}^{t + 1} \!-\! \boldsymbol{x}_{2,j}^t\}\!\\
\!\{\boldsymbol{x}_{3,j}^{t + 1} \!-\! \boldsymbol{x}_{3,j}^t\}\!\\
\end{array} \right]\!|{|^2} \vspace{2mm}\\
\! = \! {F_\mu }(\{{\boldsymbol{x}_{i,j}^t}\},\{ \boldsymbol{z}_i^t\}) - {\left[ \begin{array}{l}
\{\eta_{\boldsymbol{x}_1}  G_{\boldsymbol{x}_{1,j}}(\{{\boldsymbol{x}_{i,j}^t}\},\{ \boldsymbol{z}_i^t\})\}\\
\{\eta_{\boldsymbol{x}_2}  G_{\boldsymbol{x}_{2,j}}(\{{\boldsymbol{x}_{i,j}^t}\},\{ \boldsymbol{z}_i^t\})\}\\
\{\eta_{\boldsymbol{x}_3}  G_{\boldsymbol{x}_{3,j}}(\{{\boldsymbol{x}_{i,j}^t}\},\{ \boldsymbol{z}_i^t\})\}
\end{array} \right]^T}\left[ \begin{array}{l}
\{ {\nabla _{{\boldsymbol{x}_{1,j}}}}{F_\mu }(\{{\boldsymbol{x}_{i,j}^t}\},\{ \boldsymbol{z}_i^t\}) \}\\
\{ {\nabla _{{\boldsymbol{x}_{2,j}}}}{F_\mu }(\{{\boldsymbol{x}_{i,j}^t}\},\{ \boldsymbol{z}_i^t\}) \}\\
\{ {\nabla _{{\boldsymbol{x}_{3,j}}}}{F_\mu }(\{{\boldsymbol{x}_{i,j}^t}\},\{ \boldsymbol{z}_i^t\}) \}
\end{array} \right] \vspace{2mm} \\
+ \frac{L}{2}\sum\limits_{i = 1}^3 \sum\limits_{j = 1}^N {{\eta _{\boldsymbol{x}_i}^2}||{G_{{x_{i,j}}}}(\{{\boldsymbol{x}_{i,j}^t}\},\{ \boldsymbol{z}_i^t\})|{|^2}} .
\end{array}
\end{equation}

According to Assumption \ref{assum:2} (i.e., function $F$ has $L$-Lipschitz continuous gradient) and combining it with Cauchy-Schwarz inequality, we have that,
\begin{equation}
\label{eq:4_23_32}
    \begin{array}{l}
{F}(\{{\boldsymbol{x}_{i,j}^{t+1}}\},\{ \boldsymbol{z}_i^{t+1}\})\\
 \le {F}(\{{\boldsymbol{x}_{i,j}^{t+1}}\},\{ \boldsymbol{z}_i^t\}) + {\left[ \begin{array}{l}
\boldsymbol{z}_1^{t + 1} - \boldsymbol{z}_1^t \\
\boldsymbol{z}_2^{t + 1} - \boldsymbol{z}_2^t \\
\boldsymbol{z}_3^{t + 1} - \boldsymbol{z}_3^t 
\end{array} \right]^T}\left[ \begin{array}{l}
 {\nabla _{{\boldsymbol{z}_1}}}{F }(\{{\boldsymbol{x}_{i,j}^{t+1}}\},\{ \boldsymbol{z}_i^t\})\\
{\nabla _{{\boldsymbol{z}_2}}}{F }(\{{\boldsymbol{x}_{i,j}^{t+1}}\},\{ \boldsymbol{z}_i^t\})\\
{\nabla _{{\boldsymbol{z}_3}}}{F }(\{{\boldsymbol{x}_{i,j}^{t+1}}\},\{ \boldsymbol{z}_i^t\})
\end{array} \right] + \frac{L}{2}||\left[ \begin{array}{l}
\boldsymbol{z}_1^{t + 1} - \boldsymbol{z}_1^t \\
\boldsymbol{z}_2^{t + 1} - \boldsymbol{z}_2^t \\
\boldsymbol{z}_3^{t + 1} - \boldsymbol{z}_3^t 
\end{array} \right]|{|^2} \vspace{2mm}\\

= {F}(\{{\boldsymbol{x}_{i,j}^{t+1}}\},\{ \boldsymbol{z}_i^t\}) + {\left[ \begin{array}{l}
\boldsymbol{z}_1^{t + 1} - \boldsymbol{z}_1^t \\
\boldsymbol{z}_2^{t + 1} - \boldsymbol{z}_2^t \\
\boldsymbol{z}_3^{t + 1} - \boldsymbol{z}_3^t 
\end{array} \right]^T}\left[ \begin{array}{l}
 {\nabla _{{\boldsymbol{z}_1}}}{F }(\{{\boldsymbol{x}_{i,j}^{t}}\},\{ \boldsymbol{z}_i^t\})\\
{\nabla _{{\boldsymbol{z}_2}}}{F }(\{{\boldsymbol{x}_{i,j}^{t}}\},\{ \boldsymbol{z}_i^t\})\\
{\nabla _{{\boldsymbol{z}_3}}}{F }(\{{\boldsymbol{x}_{i,j}^{t}}\},\{ \boldsymbol{z}_i^t\})
\end{array} \right]  

\vspace{2mm}\\ +  {\left[ \begin{array}{l}
\boldsymbol{z}_1^{t + 1} - \boldsymbol{z}_1^t \\
\boldsymbol{z}_2^{t + 1} - \boldsymbol{z}_2^t \\
\boldsymbol{z}_3^{t + 1} - \boldsymbol{z}_3^t 
\end{array} \right]^T}\left[ \begin{array}{l}
 {\nabla _{{\boldsymbol{z}_1}}}{F }(\{{\boldsymbol{x}_{i,j}^{t+1}}\},\{ \boldsymbol{z}_i^t\}) -  {\nabla _{{\boldsymbol{z}_1}}}{F }(\{{\boldsymbol{x}_{i,j}^{t}}\},\{ \boldsymbol{z}_i^t\}) \\
{\nabla _{{\boldsymbol{z}_2}}}{F }(\{{\boldsymbol{x}_{i,j}^{t+1}}\},\{ \boldsymbol{z}_i^t\}) -  {\nabla _{{\boldsymbol{z}_2}}}{F }(\{{\boldsymbol{x}_{i,j}^{t}}\},\{ \boldsymbol{z}_i^t\})\\
{\nabla _{{\boldsymbol{z}_3}}}{F }(\{{\boldsymbol{x}_{i,j}^{t+1}}\},\{ \boldsymbol{z}_i^t\}) -  {\nabla _{{\boldsymbol{z}_3}}}{F }(\{{\boldsymbol{x}_{i,j}^{t}}\},\{ \boldsymbol{z}_i^t\})
\end{array} \right] + \frac{L}{2}||\left[ \begin{array}{l}
\boldsymbol{z}_1^{t + 1} - \boldsymbol{z}_1^t \\
\boldsymbol{z}_2^{t + 1} - \boldsymbol{z}_2^t \\
\boldsymbol{z}_3^{t + 1} - \boldsymbol{z}_3^t 
\end{array} \right]|{|^2} \vspace{2mm}\\

\le {F}(\{{\boldsymbol{x}_{i,j}^{t+1}}\},\{ \boldsymbol{z}_i^t\})   
  \! -\! \sum\limits_{i=1}^3 (\eta_{{\boldsymbol{z}_i}}-\frac{L\eta_{{\boldsymbol{z}_i}}^2}{2} - \frac{\eta_{{\boldsymbol{z}_i}}^2}{2})||{\nabla _{{\boldsymbol{z}_i}}}{F }(\{{\boldsymbol{x}_{i,j}^{t}}\},\{ \boldsymbol{z}_i^t\})||^2 \!+\! \sum\limits_{i=1}^3 \sum\limits_{j=1}^N \frac{L}{2} ||\boldsymbol{x}_{i,j}^{t+1} \!-\! \boldsymbol{x}_{i,j}^t||^2.

\end{array}
\end{equation}

Combining Eq. (\ref{eq:4_23_32}) with the Eq. (3.5) in \cite{ghadimi2013stochastic}, we have that,
\begin{equation}
\label{eq:4_23_33}
    \begin{array}{l}
  {F_\mu }(\{{\boldsymbol{x}_{i,j}^{t+1}}\},\{ \boldsymbol{z}_i^{t+1}\}) - \frac{\mu^2 L (N+1)\sum_id_i}{2}  \vspace{2mm}  \\
 \le {F}(\{{\boldsymbol{x}_{i,j}^{t+1}}\},\{ \boldsymbol{z}_i^{t+1}\}) \vspace{2mm}\\
 \le {F}(\{{\boldsymbol{x}_{i,j}^{t+1}}\},\{ \boldsymbol{z}_i^t\}) - \sum\limits_{i=1}^3 (\eta_{{\boldsymbol{z}_i}}-\frac{(L+1)\eta_{{\boldsymbol{z}_i}}^2}{2})||{\nabla _{{\boldsymbol{z}_i}}}{F }(\{{\boldsymbol{x}_{i,j}^{t}}\},\{ \boldsymbol{z}_i^t\})||^2 + \sum\limits_{i=1}^3 \sum\limits_{j=1}^N \frac{L}{2} ||\boldsymbol{x}_{i,j}^{t + 1} - \boldsymbol{x}_{i,j}^t||^2 \vspace{2mm} \\
 
 \le {F_\mu}(\{{\boldsymbol{x}_{i,j}^{t+1}}\},\{ \boldsymbol{z}_i^t\}) - \sum\limits_{i=1}^3 (\eta_{{\boldsymbol{z}_i}}-\frac{(L+1)\eta_{{\boldsymbol{z}_i}}^2}{2})||{\nabla _{{\boldsymbol{z}_i}}}{F }(\{{\boldsymbol{x}_{i,j}^{t}}\},\{ \boldsymbol{z}_i^t\})||^2 + \sum\limits_{i=1}^3 \sum\limits_{j=1}^N \frac{L}{2} ||\boldsymbol{x}_{i,j}^{t + 1} - \boldsymbol{x}_{i,j}^t||^2
 
\vspace{2mm} \\
 
 + \frac{\mu^2 L (N+1)\sum_id_i}{2} .
 
\end{array}
\end{equation}

Combining Eq. (\ref{eq:4_19_3}) with Eq. (\ref{eq:4_23_33}), we can obtain that,
\begin{equation}
\label{eq:4_23_34}
    \begin{array}{l}
{F_\mu }(\{{\boldsymbol{x}_{i,j}^{t+1}}\},\{ \boldsymbol{z}_i^{t+1}\})\\
 \le {F_\mu }(\{{\boldsymbol{x}_{i,j}^t}\},\{ \boldsymbol{z}_i^t\}) - {\left[ \begin{array}{l}
\{\eta_{\boldsymbol{x}_1}  G_{\boldsymbol{x}_{1,j}}(\{{\boldsymbol{x}_{i,j}^t}\},\{ \boldsymbol{z}_i^t\})\}\\
\{\eta_{\boldsymbol{x}_1}  G_{\boldsymbol{x}_{2,j}}(\{{\boldsymbol{x}_{i,j}^t}\},\{ \boldsymbol{z}_i^t\})\}\\
\{\eta_{\boldsymbol{x}_1}  G_{\boldsymbol{x}_{3,j}}(\{{\boldsymbol{x}_{i,j}^t}\},\{ \boldsymbol{z}_i^t\})\}
\end{array} \right]^T}\left[ \begin{array}{l}
\{ {\nabla _{{\boldsymbol{x}_{1,j}}}}{F_\mu }(\{{\boldsymbol{x}_{i,j}^t}\},\{ \boldsymbol{z}_i^t\}) \}\\
\{ {\nabla _{{\boldsymbol{x}_{2,j}}}}{F_\mu }(\{{\boldsymbol{x}_{i,j}^t}\},\{ \boldsymbol{z}_i^t\}) \}\\
\{ {\nabla _{{\boldsymbol{x}_{3,j}}}}{F_\mu }(\{{\boldsymbol{x}_{i,j}^t}\},\{ \boldsymbol{z}_i^t\}) \}
\end{array} \right] \vspace{2mm} \\

+ \frac{L}{2}\sum\limits_{i = 1}^3 \sum\limits_{j = 1}^N {{\eta_{\boldsymbol{x}_i}^2}||{G_{{x_{i,j}}}}(\{{\boldsymbol{x}_{i,j}^t}\},\{ \boldsymbol{z}_i^t\})|{|^2}} - \sum\limits_{i=1}^3 (\eta_{{\boldsymbol{z}_i}}-\frac{(L+1)\eta_{{\boldsymbol{z}_i}}^2}{2})||{\nabla _{{\boldsymbol{z}_i}}}{F }(\{{\boldsymbol{x}_{i,j}^{t}}\},\{ \boldsymbol{z}_i^t\})||^2 \vspace{2mm} \\

+ \sum\limits_{i=1}^3 \sum\limits_{j=1}^N \frac{L}{2} ||\boldsymbol{x}_{i,j}^{t + 1} - \boldsymbol{x}_{i,j}^t||^2 + \mu^2 L (N+1)\sum_id_i

\\
= {F_\mu }(\{{\boldsymbol{x}_{i,j}^t}\},\{ \boldsymbol{z}_i^t\}) - {\left[ \begin{array}{l}
\{\eta_{\boldsymbol{x}_1}  G_{\boldsymbol{x}_{1,j}}(\{{\boldsymbol{x}_{i,j}^t}\},\{ \boldsymbol{z}_i^t\})\}\\
\{\eta_{\boldsymbol{x}_1}  G_{\boldsymbol{x}_{2,j}}(\{{\boldsymbol{x}_{i,j}^t}\},\{ \boldsymbol{z}_i^t\})\}\\
\{\eta_{\boldsymbol{x}_1}  G_{\boldsymbol{x}_{3,j}}(\{{\boldsymbol{x}_{i,j}^t}\},\{ \boldsymbol{z}_i^t\})\}
\end{array} \right]^T}\left[ \begin{array}{l}
\{ {\nabla _{{\boldsymbol{x}_{1,j}}}}{F_\mu }(\{{\boldsymbol{x}_{i,j}^t}\},\{ \boldsymbol{z}_i^t\}) \}\\
\{ {\nabla _{{\boldsymbol{x}_{2,j}}}}{F_\mu }(\{{\boldsymbol{x}_{i,j}^t}\},\{ \boldsymbol{z}_i^t\}) \}\\
\{ {\nabla _{{\boldsymbol{x}_{3,j}}}}{F_\mu }(\{{\boldsymbol{x}_{i,j}^t}\},\{ \boldsymbol{z}_i^t\}) \}
\end{array} \right] \vspace{2mm} \\
+ \sum\limits_{i = 1}^3 \sum\limits_{j = 1}^N {L{\eta_{\boldsymbol{x}_i}^2}||{G_{{x_{i,j}}}}(\{{\boldsymbol{x}_{i,j}^t}\},\{ \boldsymbol{z}_i^t\})|{|^2}} - \sum\limits_{i=1}^3 (\eta_{{\boldsymbol{z}_i}}-\frac{(L+1)\eta_{{\boldsymbol{z}_i}}^2}{2})||{\nabla _{{\boldsymbol{z}_i}}}{F }(\{{\boldsymbol{x}_{i,j}^{t}}\},\{ \boldsymbol{z}_i^t\})||^2 \vspace{2mm}\\ 

+ \mu^2 L (N+1)\sum_id_i .
\end{array}
\end{equation}

Taking expectation on the both sides of Eq. (\ref{eq:4_19_3}), we can obtain that,
\begin{equation}
\label{eq:4_19_4}
    \begin{array}{l}
\mathbb{E}[{F_\mu }(\{{\boldsymbol{x}_{i,j}^{t+1}}\},\{ \boldsymbol{z}_i^{t+1}\})] \vspace{2mm} \\
 \le \mathbb{E}[{F_\mu }(\{{\boldsymbol{x}_{i,j}^t}\},\{ \boldsymbol{z}_i^t\})] - \sum\limits_{i = 1}^3 \sum\limits_{j = 1}^N{\eta_{\boldsymbol{x}_i}||{\nabla _{{x_{i,j}}}}{F_\mu }(\{{\boldsymbol{x}_{i,j}^t}\},\{ \boldsymbol{z}_i^t\})|{|^2}} + \mu^2 L (N+1)\sum_id_i \\
 
 + \sum\limits_{i = 1}^3 \sum\limits_{j = 1}^N  {L{\eta_{\boldsymbol{x}_i}^2}\mathbb{E}[||{G_{{x_{i,j}}}}(\{{\boldsymbol{x}_{i,j}^t}\},\{ \boldsymbol{z}_i^t\})||^2]} - \sum\limits_{i=1}^3 (\eta_{{\boldsymbol{z}_i}}-\frac{(L+1)\eta_{{\boldsymbol{z}_i}}^2}{2})||{\nabla _{{\boldsymbol{z}_i}}}{F }(\{{\boldsymbol{x}_{i,j}^{t}}\},\{ \boldsymbol{z}_i^t\})||^2 .
\end{array}
\end{equation}

Combining the definition of ${G_{{\boldsymbol{x}_{1,j}}}}, {G_{{\boldsymbol{x}_{2,j}}}}, {G_{{\boldsymbol{x}_{3,j}}}}$ with the Eq. (3.12) in \cite{ghadimi2013stochastic}, we have that,
\begin{equation}
\label{eq:4_19_5}
    E[||{G_{{\boldsymbol{x}_{1,j}}}}(\{{\boldsymbol{x}_{i,j}^t}\},\{ \boldsymbol{z}_i^t\})|{|^2}] \le 2({d_1} + 4)||{\nabla _{{\boldsymbol{x}_{1,j}}}}F(\{{\boldsymbol{x}_{i,j}^t}\},\{ \boldsymbol{z}_i^t\})|{|^2} + \frac{{{\mu ^2}{L^2}}}{2}{({d_1} + 6)^3},
\end{equation}
\begin{equation}
\label{eq:4_19_6}
    E[||{G_{{\boldsymbol{x}_{2,j}}}}(\{{\boldsymbol{x}_{i,j}^t}\},\{ \boldsymbol{z}_i^t\})|{|^2}] \le 2({d_2} + 4)||{\nabla _{{\boldsymbol{x}_{2,j}}}}F(\{{\boldsymbol{x}_{i,j}^t}\},\{ \boldsymbol{z}_i^t\})|{|^2} + \frac{{{\mu ^2}{L^2}}}{2}{({d_2} + 6)^3},
\end{equation}
\begin{equation}
\label{eq:4_19_7}
    E[||{G_{{\boldsymbol{x}_{3,j}}}}(\{{\boldsymbol{x}_{i,j}^t}\},\{ \boldsymbol{z}_i^t\})|{|^2}] \le 2({d_3} + 4)||{\nabla _{{\boldsymbol{x}_{3,j}}}}F(\{{\boldsymbol{x}_{i,j}^t}\},\{ \boldsymbol{z}_i^t\})|{|^2} + \frac{{{\mu ^2}{L^2}}}{2}{({d_3} + 6)^3}.
\end{equation}

By combining Eq. (\ref{eq:4_19_4}) with Eq. (\ref{eq:4_19_5}), (\ref{eq:4_19_6}), and (\ref{eq:4_19_7}), we can get that,
\begin{equation}
    \begin{array}{l}
\mathbb{E}[{F_\mu }(\{{\boldsymbol{x}_{i,j}^{t+1}}\},\{ \boldsymbol{z}_i^{t+1}\})] \vspace{2mm} \\
 \le \mathbb{E}[{F_\mu }(\{{\boldsymbol{x}_{i,j}^t}\},\{ \boldsymbol{z}_i^t\})] - \sum\limits_{i = 1}^3 \sum\limits_{j = 1}^N{\eta_{\boldsymbol{x}_i}||{\nabla _{{x_{i,j}}}}{F_\mu }(\{{\boldsymbol{x}_{i,j}^t}\},\{ \boldsymbol{z}_i^t\})|{|^2}} + \mu^2 L (N+1)\sum_id_i \\
 
 + \sum\limits_{i = 1}^3 \sum\limits_{j = 1}^N {L{\eta_{\boldsymbol{x}_i}^2}\left( {2({d_i} + 4)||{\nabla _{{x_{i,j}}}}F(\{{\boldsymbol{x}_{i,j}^t}\},\{ \boldsymbol{z}_i^t\})|{|^2} + \frac{{{\mu ^2}{L^2}}}{2}{{({d_i} + 6)}^3}} \right) } \vspace{2mm} \\
 
 - \sum\limits_{i=1}^3 (\eta_{{\boldsymbol{z}_i}}-\frac{(L+1)\eta_{{\boldsymbol{z}_i}}^2}{2})||{\nabla _{{\boldsymbol{z}_i}}}{F }(\{{\boldsymbol{x}_{i,j}^{t}}\},\{ \boldsymbol{z}_i^t\})||^2,

\end{array}
\end{equation}
that is,
\begin{equation}
\label{eq:4_19_9}
    \begin{array}{l}
\sum\limits_{i = 1}^3 \sum\limits_{j = 1}^N{\eta_{\boldsymbol{x}_i}||{\nabla _{{x_{i,j}}}}{F_\mu }(\{{\boldsymbol{x}_{i,j}^t}\},\{ \boldsymbol{z}_i^t\})|{|^2}}   + \sum\limits_{i=1}^3 (\eta_{{\boldsymbol{z}_i}}-\frac{(L+1)\eta_{{\boldsymbol{z}_i}}^2}{2})||{\nabla _{{\boldsymbol{z}_i}}}{F }(\{{\boldsymbol{x}_{i,j}^{t}}\},\{ \boldsymbol{z}_i^t\})||^2 \vspace{2mm} \\
 \le \mathbb{E}[{F_\mu }(\{{\boldsymbol{x}_{i,j}^t}\},\{ \boldsymbol{z}_i^t\})] - \mathbb{E}[{F_\mu }(\{{\boldsymbol{x}_{i,j}^{t+1}}\},\{ \boldsymbol{z}_i^{t+1}\})]  + \mu^2 L (N+1)\sum_id_i \vspace{2mm} \\
 
 + \sum\limits_{i = 1}^3 \sum\limits_{j = 1}^N {L{\eta_{\boldsymbol{x}_i}^2}\left( {2({d_i} + 4)||{\nabla _{{x_{i,j}}}}F(\{{\boldsymbol{x}_{i,j}^t}\},\{ \boldsymbol{z}_i^t\})|{|^2} + \frac{{{\mu ^2}{L^2}}}{2}{{({d_i} + 6)}^3}} \right) } .

\end{array}
\end{equation}

Combining Eq. (\ref{eq:4_19_9}) with Eq. (3.8) in \cite{ghadimi2013stochastic}, we can obtain that,
\begin{equation}
    \begin{array}{l}
\sum\limits_{i = 1}^3 \sum\limits_{j = 1}^N {\eta_{\boldsymbol{x}_i}\left( {\frac{1}{2}||{\nabla _{{x_{i,j}}}}{F }(\{{\boldsymbol{x}_{i,j}^t}\},\{ \boldsymbol{z}_i^t\})|{|^2} - \frac{{{\mu ^2}{L^2}}}{4}{{({d_i} + 3)}^3}} \right)} \vspace{1mm} \\

+ \sum\limits_{i=1}^3 (\eta_{{\boldsymbol{z}_i}}-\frac{(L+1)\eta_{{\boldsymbol{z}_i}}^2}{2})||{\nabla _{{\boldsymbol{z}_i}}}{F }(\{{\boldsymbol{x}_{i,j}^{t}}\},\{ \boldsymbol{z}_i^t\})||^2 \vspace{1mm} \\
 \le \sum\limits_{i = 1}^3 \sum\limits_{j = 1}^N{\eta_{\boldsymbol{x}_i}||{\nabla _{{x_{i,j}}}}{F_\mu }(\{{\boldsymbol{x}_{i,j}^t}\},\{ \boldsymbol{z}_i^t\})|{|^2}}   + \sum\limits_{i=1}^3 (\eta_{{\boldsymbol{z}_i}}-\frac{(L+1)\eta_{{\boldsymbol{z}_i}}^2}{2})||{\nabla _{{\boldsymbol{z}_i}}}{F }(\{{\boldsymbol{x}_{i,j}^{t}}\},\{ \boldsymbol{z}_i^t\})||^2 \vspace{2mm} \\ 
 
 \le \mathbb{E}[{F_\mu }(\{{\boldsymbol{x}_{i,j}^t}\},\{ \boldsymbol{z}_i^t\})] - \mathbb{E}[{F_\mu }(\{{\boldsymbol{x}_{i,j}^{t+1}}\},\{ \boldsymbol{z}_i^{t+1}\})]  + \mu^2 L (N+1)\sum_id_i  \vspace{2mm}\\
 
 + \sum\limits_{i = 1}^3 \sum\limits_{j = 1}^N {L{\eta_{\boldsymbol{x}_i}^2}\left( {2({d_i} + 4)||{\nabla _{{x_{i,j}}}}F(\{{\boldsymbol{x}_{i,j}^t}\},\{ \boldsymbol{z}_i^t\})|{|^2} + \frac{{{\mu ^2}{L^2}}}{2}{{({d_i} + 6)}^3}} \right) } ,
\end{array}
\end{equation}
that is,
\begin{equation}
\label{eq:4_19_11}
    \begin{array}{l}
\sum\limits_{i = 1}^3 \sum\limits_{j = 1}^N  {\left( {\frac{{{\eta_{\boldsymbol{x}_i}}}}{2} - 2L({d_i} + 4){\eta_{\boldsymbol{x}_i}^2}} \right)||{\nabla _{{x_{i,j}}}}F(\{{\boldsymbol{x}_{i,j}^t}\},\{ \boldsymbol{z}_i^t\})|{|^2}} \vspace{1mm}\\
 + \sum\limits_{i=1}^3 (\eta_{{\boldsymbol{z}_i}}-\frac{(L+1)\eta_{{\boldsymbol{z}_i}}^2}{2})||{\nabla _{{\boldsymbol{z}_i}}}{F }(\{{\boldsymbol{x}_{i,j}^{t}}\},\{ \boldsymbol{z}_i^t\})||^2\vspace{1mm}\\
 
 \le {F_\mu }(\{{\boldsymbol{x}_{i,j}^t}\},\{ \boldsymbol{z}_i^t\}) - {F_\mu }(\{{\boldsymbol{x}_{i,j}^{t+1}}\},\{ \boldsymbol{z}_i^{t+1}\}) + \sum\limits_{i = 1}^3\sum\limits_{j = 1}^N {\frac{{{\eta_{\boldsymbol{x}_i}^2}{\mu ^2}{L^3}}}{2}{{({d_i} + 6)}^3}} \vspace{1mm} \\
 
 + \sum\limits_{i = 1}^3\sum\limits_{j = 1}^N {\frac{{{\mu ^2}{L^2}{\eta_{\boldsymbol{x}_i}}}}{4}{{({d_i} + 3)}^3}}  + \mu^2 L (N+1)\sum_id_i . 
\end{array}
\end{equation}

According to the setting of ${\eta_{\boldsymbol{x}_i}}, i=1,2,3$, i.e., $0 < {\eta_{\boldsymbol{x}_i}} \le \frac{1}{{8L({d_i} + 4)}}, i=1,2,3$, we have that,
\begin{equation}
\label{eq:4_19_12}
    \frac{{{\eta_{\boldsymbol{x}_i}}}}{2} - 2L({d_i} + 4){\eta_{\boldsymbol{x}_i}^2} > 0,i=1,2,3.
\end{equation}

Likewise, according to the setting of ${\eta_{\boldsymbol{z}_i}}, i=1,2,3$, i.e., $0<{\eta_{\boldsymbol{z}_i}} \le \frac{3}{2(L+1)}, i=1,2,3$, we have that,
\begin{equation}
\label{eq:4_23_44}
    \eta_{{\boldsymbol{z}_i}}-\frac{(L+1)\eta_{{\boldsymbol{z}_i}}^2}{2}>0,i=1,2,3.
\end{equation}

Combining Eq. (\ref{eq:4_19_11}) with Eq. (\ref{eq:4_19_12}) and (\ref{eq:4_23_44}), we can obtain that,

\begin{equation}
\label{eq:4_19_13}
    \begin{split}
&\sum\limits_{i = 1}^3 \sum\limits_{j = 1}^N  {||{\nabla _{{x_{i,j}}}}F(\{{\boldsymbol{x}_{i,j}^t}\},\{ \boldsymbol{z}_i^t\})|{|^2}} 
 + \sum\limits_{i=1}^3 ||{\nabla _{{\boldsymbol{z}_i}}}{F }(\{{\boldsymbol{x}_{i,j}^{t}}\},\{ \boldsymbol{z}_i^t\})||^2 \vspace{1mm} \\
 &\le \frac{\sum\limits_{i = 1}^3 \sum\limits_{j = 1}^N  {\left( {\frac{{{\eta_{\boldsymbol{x}_i}}}}{2} - 2L({d_i} + 4){\eta_{\boldsymbol{x}_i}^2}} \right)||{\nabla _{{x_{i,j}}}}F(\{{\boldsymbol{x}_{i,j}^t}\},\{ \boldsymbol{z}_i^t\})|{|^2}} 
 }{{\min \left\{ \frac{{{\eta_{\boldsymbol{x}_i}}}}{2} - 2L({d_i} + 4){\eta_{\boldsymbol{x}_i}^2}, \eta_{{\boldsymbol{z}_i}}-\frac{(L+1)\eta_{{\boldsymbol{z}_i}}^2}{2}, i=1,2,3 \right\}}} \vspace{2mm}\\
 & + \frac{\sum\limits_{i=1}^3 (\eta_{{\boldsymbol{z}_i}}-\frac{(L+1)\eta_{{\boldsymbol{z}_i}}^2}{2})||{\nabla _{{\boldsymbol{z}_i}}}{F }(\{{\boldsymbol{x}_{i,j}^{t}}\},\{ \boldsymbol{z}_i^t\})||^2}{{\min \left\{ \frac{{{\eta_{\boldsymbol{x}_i}}}}{2} - 2L({d_i} + 4){\eta_{\boldsymbol{x}_i}^2}, \eta_{{\boldsymbol{z}_i}}-\frac{(L+1)\eta_{{\boldsymbol{z}_i}}^2}{2}, i=1,2,3 \right\}}}
 \vspace{2mm}\\
& \le \frac{{F_\mu }(\{{\boldsymbol{x}_{i,j}^t}\},\{ \boldsymbol{z}_i^t\}) - {F_\mu }(\{{\boldsymbol{x}_{i,j}^{t+1}}\},\{ \boldsymbol{z}_i^{t+1}\}) + \sum\limits_{i = 1}^3 {\frac{{{\eta_{\boldsymbol{x}_i}^2}{\mu ^2}{L^3}}N}{2}{{({d_i} + 6)}^3}}  }{{\min \left\{ \frac{{{\eta_{\boldsymbol{x}_i}}}}{2} - 2L({d_i} + 4){\eta_{\boldsymbol{x}_i}^2}, \eta_{{\boldsymbol{z}_i}}-\frac{(L+1)\eta_{{\boldsymbol{z}_i}}^2}{2}, i=1,2,3 \right\}}}
\\
&+ \frac{ + \sum\limits_{i = 1}^3 {\frac{{{\mu ^2}{L^2}{\eta_{\boldsymbol{x}_i}}}N}{4}{{({d_i} + 3)}^3}}  + \mu^2 L (N+1)\sum_id_i}{{\min \left\{ \frac{{{\eta_{\boldsymbol{x}_i}}}}{2} - 2L({d_i} + 4){\eta_{\boldsymbol{x}_i}^2}, \eta_{{\boldsymbol{z}_i}}-\frac{(L+1)\eta_{{\boldsymbol{z}_i}}^2}{2}, i=1,2,3 \right\}}}.
    \end{split}
\end{equation}

Summing up the inequality in Eq. (\ref{eq:4_19_13}) from $t=T_1$ to $t=T(\epsilon)-1$, we have that,

\begin{equation}
    \begin{split}
  &     \frac{1}{{T(\epsilon) - {T_1}}}\sum\limits_{t = {T_1}}^{T(\epsilon) - 1} (\sum\limits_{i = 1}^3 \sum\limits_{j = 1}^N  {||{\nabla _{{x_{i,j}}}}F(\{{\boldsymbol{x}_{i,j}^t}\},\{ \boldsymbol{z}_i^t\})|{|^2}} 
 + \sum\limits_{i=1}^3 ||{\nabla _{{\boldsymbol{z}_i}}}{F }(\{{\boldsymbol{x}_{i,j}^{t}}\},\{ \boldsymbol{z}_i^t\})||^2) \vspace{2mm} \\
& \le \frac{{{F_\mu }(\{{\boldsymbol{x}_{i,j}^{T_1}}\},\{ \boldsymbol{z}_i^{T_1}\}) - {F_\mu }(\{{\boldsymbol{x}_{i,j}^{T(\epsilon)}}\},\{ \boldsymbol{z}_i^{T(\epsilon)}\})}}{{\min \left\{ \frac{{{\eta_{\boldsymbol{x}_i}}}}{2} - 2L({d_i} + 4){\eta_{\boldsymbol{x}_i}^2}, \eta_{{\boldsymbol{z}_i}}-\frac{(L+1)\eta_{{\boldsymbol{z}_i}}^2}{2}, i=1,2,3 \right\}\left( {T(\epsilon) - {T_1}} \right)}} 
 \vspace{2mm}\\
 &+ 
 \frac{\sum\limits_{i = 1}^3 {\frac{{{\eta_{\boldsymbol{x}_i}^2}{\mu ^2}{L^3}}N}{2}{{({d_i} + 6)}^3}}  
 + \sum\limits_{i = 1}^3 {\frac{{{\mu ^2}{L^2}{\eta_{\boldsymbol{x}_i}}}N}{4}{{({d_i} + 3)}^3}}  + \mu^2 L (N+1)\sum_id_i}{{\min \left\{ \frac{{{\eta_{\boldsymbol{x}_i}}}}{2} - 2L({d_i} + 4){\eta_{\boldsymbol{x}_i}^2}, \eta_{{\boldsymbol{z}_i}}-\frac{(L+1)\eta_{{\boldsymbol{z}_i}}^2}{2}, i=1,2,3 \right\}}} \vspace{2mm}\\
 & \le \frac{{\mathop {\max }\limits_{t \in [{T_1}]} {F_\mu }(\{{\boldsymbol{x}_{i,j}^t}\},\{ \boldsymbol{z}_i^t\}) - {F_\mu }^*}}{{\min \left\{ \frac{{{\eta_{\boldsymbol{x}_i}}}}{2} - 2L({d_i} + 4){\eta_{\boldsymbol{x}_i}^2}, \eta_{{\boldsymbol{z}_i}}-\frac{(L+1)\eta_{{\boldsymbol{z}_i}}^2}{2}, i=1,2,3 \right\}\left( {T(\epsilon) - {T_1}} \right)}}
\vspace{2mm}\\
&  + 
 \frac{ \sum\limits_{i = 1}^3 {\frac{{{\eta_{\boldsymbol{x}_i}^2}{\mu ^2}{L^3}}N}{2}{{({d_i} + 6)}^3}}  
 + \sum\limits_{i = 1}^3 {\frac{{{\mu ^2}{L^2}{\eta_{\boldsymbol{x}_i}}}N}{4}{{({d_i} + 3)}^3}}  + \mu^2 L (N+1)\sum_id_i}{{\min \left\{ \frac{{{\eta_{\boldsymbol{x}_i}}}}{2} - 2L({d_i} + 4){\eta_{\boldsymbol{x}_i}^2}, \eta_{{\boldsymbol{z}_i}}-\frac{(L+1)\eta_{{\boldsymbol{z}_i}}^2}{2}, i=1,2,3 \right\}}}.
    \end{split}
\end{equation}

According to the setting of ${\eta_{\boldsymbol{x}_i}},{\eta_{\boldsymbol{z}_i}}, i=1,2,3$, we can obtain that,
\begin{equation}
    \frac{{{\eta_{\boldsymbol{x}_i}}}}{2} - 2L({d_i} + 4){\eta_{\boldsymbol{x}_i}^2} = {\eta_{\boldsymbol{x}_i}}\left( {\frac{1}{2} - 2L({d_i} + 4){\eta_{\boldsymbol{x}_i}}} \right) \ge \frac{{{\eta_{\boldsymbol{x}_i}}}}{4}, i=1,2,3,
\end{equation}
\begin{equation}
    \eta_{{\boldsymbol{z}_i}}-\frac{(L+1)\eta_{{\boldsymbol{z}_i}}^2}{2} = \eta_{{\boldsymbol{z}_i}}(1 - \frac{(L+1)\eta_{{\boldsymbol{z}_i}}}{2}) \ge \frac{\eta_{{\boldsymbol{z}_i}}}{4}, i=1,2,3.
\end{equation}

Thus, we have that,
\begin{equation}
    \begin{split}
&\frac{1}{{T(\epsilon) - {T_1}}}\sum\limits_{t = {T_1}}^{T(\epsilon) - 1} (\sum\limits_{i = 1}^3 \sum\limits_{j = 1}^N  {||{\nabla _{{x_{i,j}}}}F(\{{\boldsymbol{x}_{i,j}^t}\},\{ \boldsymbol{z}_i^t\})|{|^2}} 
 + \sum\limits_{i=1}^3 ||{\nabla _{{\boldsymbol{z}_i}}}{F }(\{{\boldsymbol{x}_{i,j}^{t}}\},\{ \boldsymbol{z}_i^t\})||^2) \vspace{2mm} \\
 &\le \frac{{4\left( {\mathop {\max }\limits_{t \in [{T_1}]} {F_\mu }(\{{\boldsymbol{x}_{i,j}^t}\},\{ \boldsymbol{z}_i^t\}) - {F_\mu }^*} \right)}}{{\min \left\{ {{\eta_{\boldsymbol{x}_1}},{\eta_{\boldsymbol{x}_2}},{\eta_{\boldsymbol{x}_3}},{\eta_{\boldsymbol{z}_1}},{\eta_{\boldsymbol{z}_2}},{\eta_{\boldsymbol{z}_3}}} \right\}\left( {T(\epsilon) - {T_1}} \right)}}
 \vspace{2mm} \\
& + \frac{   \sum\limits_{i = 1}^3 2 {{{\eta_{\boldsymbol{x}_i}^2}{\mu ^2}{L^3}}N{{({d_i} + 6)}^3}}  
 + \sum\limits_{i = 1}^3 {{{\mu ^2}{L^2}{\eta_{\boldsymbol{x}_i}}}N{{({d_i} + 3)}^3}}  + 4\mu^2 L (N+1)\sum_id_i          }{{\min \left\{ {{\eta_{\boldsymbol{x}_1}},{\eta_{\boldsymbol{x}_2}},{\eta_{\boldsymbol{x}_3}},{\eta_{\boldsymbol{z}_1}},{\eta_{\boldsymbol{z}_2}},{\eta_{\boldsymbol{z}_3}}} \right\}}}.
\end{split}
\end{equation}

According to the setting that,
\begin{equation}
    {\eta_{\boldsymbol{x}_i}} = 
    {\eta_{\boldsymbol{z}_i}} = 
  \min \left\{ {\frac{1}{{8L({d_1} + 4)}},\frac{1}{{8L({d_2} + 4)}},\frac{1}{{8L({d_3} + 4)}}, \frac{3}{2(L+1)}, \frac{1}{{\sqrt {T(\epsilon) - {T_1}} }}} \right\}, i=1,2,3,
\end{equation}

we have that,
\begin{equation}
    \begin{split}
&\frac{1}{{T(\epsilon) - {T_1}}}\sum\limits_{t = {T_1}}^{T(\epsilon) - 1} (\sum\limits_{i = 1}^3 \sum\limits_{j = 1}^N  {||{\nabla _{{x_{i,j}}}}F(\{{\boldsymbol{x}_{i,j}^t}\},\{ \boldsymbol{z}_i^t\})|{|^2}} 
 + \sum\limits_{i=1}^3 ||{\nabla _{{\boldsymbol{z}_i}}}{F }(\{{\boldsymbol{x}_{i,j}^{t}}\},\{ \boldsymbol{z}_i^t\})||^2) \vspace{2mm} \\
& \le \frac{{4\left( {\mathop {\max }\limits_{t \in [{T_1}]} {F_\mu }(\{{\boldsymbol{x}_{i,j}^t}\},\{ \boldsymbol{z}_i^t\}) - {F_\mu }^*} \right)}}{{\min \left\{ {\frac{1}{{8L({d_1} + 4)}},\frac{1}{{8L({d_2} + 4)}},\frac{1}{{8L({d_3} + 4)}},\frac{3}{2(L+1)},\frac{1}{{\sqrt {T(\epsilon) - {T_1}} }}} \right\}\left( {T(\epsilon) - {T_1}} \right)}} 
 \vspace{2mm} \\
& + \sum\limits_{i = 1}^3 2{{{\eta_{\boldsymbol{x}_i}}{\mu ^2}{L^3}}N{{({d_i} + 6)}^3}}  
 + \sum\limits_{i = 1}^3 {{{\mu ^2}{L^2}{}}N{{({d_i} + 3)}^3}} \vspace{2mm} \\
& 
 + \sum\limits_{i = 1}^3 4\mu^2 L (N+1)d_i\frac{1}{\min \left\{ {\frac{1}{{8L({d_1} + 4)}},\frac{1}{{8L({d_2} + 4)}},\frac{1}{{8L({d_3} + 4)}},\frac{3}{2(L+1)},\frac{1}{{\sqrt {T(\epsilon) - {T_1}} }}} \right\}}  
\vspace{2mm} \\
&
 \le \frac{{4\left( {\mathop {\max }\limits_{t \in [{T_1}]} {F_\mu }(\{{\boldsymbol{x}_{i,j}^t}\},\{ \boldsymbol{z}_i^t\}) - {F_\mu }^*} \right)\left( {\max \left\{ {8L({d_1} + 4),8L({d_2} + 4),8L({d_3} + 4), \frac{2(L+1)}{3}} \right\} } \right)}}{{T(\epsilon) - {T_1}}} \vspace{2mm} \\
& + \frac{4\left( {\mathop {\max }\limits_{t \in [{T_1}]} {F_\mu }(\{{\boldsymbol{x}_{i,j}^t}\},\{ \boldsymbol{z}_i^t\}) - {F_\mu }^*} \right) \sqrt {T(\epsilon) - {T_1}}}{T(\epsilon) - {T_1}} \vspace{2mm} \\
& + \sum\limits_{i = 1}^3 2{{{\eta_{\boldsymbol{x}_i}}{\mu ^2}{L^3}}N{{({d_i} + 6)}^3}}  
 + \sum\limits_{i = 1}^3 {{{\mu ^2}{L^2}{}}N{{({d_i} + 3)}^3}} \vspace{2mm} \\
 &
 + \sum\limits_{i = 1}^3 4\mu^2 L (N+1)d_i\left({\max \left\{ {8L({d_1} + 4),8L({d_2} + 4),8L({d_3} + 4), \frac{2(L+1)}{3}} \right\} \!+\! \sqrt {T(\epsilon) - {T_1}} }\right).
\end{split}
\end{equation}

Since ${\eta_{\boldsymbol{x}_i}} \le \frac{1}{{8L({d_i} + 4)}},i = 1, 2 ,3$, we can obtain that,
\begin{equation}
    \begin{split}
&\frac{1}{{T(\epsilon) - {T_1}}}\sum\limits_{t = {T_1}}^{T(\epsilon) - 1} (\sum\limits_{i = 1}^3 \sum\limits_{j = 1}^N  {||{\nabla _{{x_{i,j}}}}F(\{{\boldsymbol{x}_{i,j}^t}\},\{ \boldsymbol{z}_i^t\})|{|^2}} 
 + \sum\limits_{i=1}^3 ||{\nabla _{{\boldsymbol{z}_i}}}{F }(\{{\boldsymbol{x}_{i,j}^{t}}\},\{ \boldsymbol{z}_i^t\})||^2) \vspace{2mm} \\
&
 \le \frac{{4\left( {\mathop {\max }\limits_{t \in [{T_1}]} {F_\mu }(\{{\boldsymbol{x}_{i,j}^t}\},\{ \boldsymbol{z}_i^t\}) - {F_\mu }^*} \right)\left( {\max \left\{ {8L({d_1} + 4),8L({d_2} + 4),8L({d_3} + 4), \frac{2(L+1)}{3}} \right\} } \right)}}{{T(\epsilon) - {T_1}}} \vspace{2mm} \\
 &
+ \frac{4\left( {\mathop {\max }\limits_{t \in [{T_1}]} {F_\mu }(\{{\boldsymbol{x}_{i,j}^t}\},\{ \boldsymbol{z}_i^t\}) - {F_\mu }^*} \right) \sqrt {T(\epsilon) - {T_1}}}{T(\epsilon) - {T_1}}\! +\! \frac{{{\mu ^2}{L^2}N}}{4}\sum\limits_{i = 1}^3 {\frac{{{{({d_i} + 6)}^3}}}{{{d_i} + 4}}} 
 \!+\! {\mu ^2}{L^2}\sum\limits_{i = 1}^3 {{{({d_i} + 3)}^3}} \vspace{2mm} \\
 &
 + \sum\limits_{i = 1}^3 4\mu^2 L (N+1)d_i\left({\max \left\{ {8L({d_1} + 4),8L({d_2} + 4),8L({d_3} + 4), \frac{2(L+1)}{3}} \right\} \!+\! \sqrt {T(\epsilon) - {T_1}} }\right).
\end{split}
\end{equation}

Because of $T(\epsilon) - T_1 \ge 1$, we have that $\frac{1}{T(\epsilon) - T_1} \le \frac{1}{\sqrt T(\epsilon) - T_1}$. Combining with the setting of ${\mu}$, i.e., ${\mu ^2} \le \frac{1}{{ {T(\epsilon)-T_1} }}$,  we can obtain that,
\begin{equation}
\label{eq:4_19_20}
    \begin{split}
&\frac{1}{{T(\epsilon) - {T_1}}}\sum\limits_{t = {T_1}}^{T(\epsilon) - 1} (\sum\limits_{i = 1}^3 \sum\limits_{j = 1}^N  {||{\nabla _{{x_{i,j}}}}F(\{{\boldsymbol{x}_{i,j}^t}\},\{ \boldsymbol{z}_i^t\})|{|^2}} 
 + \sum\limits_{i=1}^3 ||{\nabla _{{\boldsymbol{z}_i}}}{F }(\{{\boldsymbol{x}_{i,j}^{t}}\},\{ \boldsymbol{z}_i^t\})||^2) \vspace{2mm} \\
& 
 \le \frac{4{\max \left\{ {8L({d_1} + 4),8L({d_2} + 4),8L({d_3} + 4), \frac{2(L+1)}{3}} \right\}\left( {\mathop {\max }\limits_{t \in [{T_1}]} {F_\mu }(\{{\boldsymbol{x}_{i,j}^t}\},\{ \boldsymbol{z}_i^t\}) - {F_\mu }^*} \right)}}{{T(\epsilon) - {T_1}}} \vspace{2mm}\\
 &
 + \frac{{\mathop {\max }\limits_{t \in [{T_1}]} {F_\mu }(\{{\boldsymbol{x}_{i,j}^t}\},\{ \boldsymbol{z}_i^t\}) - {F_\mu }^*}}{{\sqrt {T(\epsilon) - {T_1}} }} 
 + \frac{{{L^2}}}{4}\sum\limits_{i = 1}^3 {\frac{{{{({d_i} + 6)}^3}}}{{{d_i} + 4}}} \frac{1}{{ {T(\epsilon)-T_1} }} + {L^2}\sum\limits_{i = 1}^3 {{{({d_i} + 3)}^3}} \frac{1}{{ {T(\epsilon)-T_1} }}  \vspace{2mm} \\
&
 + \sum\limits_{i = 1}^3 \left({\max \left\{ {8L({d_1} + 4),8L({d_2} + 4),8L({d_3} + 4), \frac{2(L+1)}{3}} \right\} + \sqrt {T(\epsilon) - {T_1}} }\right) \frac{4 L (N+1)d_i}{{ {T(\epsilon)-T_1} }}  \vspace{2mm} \\
&
 \le \frac{4(1 + {\max \left\{ {8L({d_1} + 4),8L({d_2} + 4),8L({d_3} + 4), \frac{2(L+1)}{3}} \right\})\left( {\mathop {\max }\limits_{t \in [{T_1}]} {F_\mu }(\{{\boldsymbol{x}_{i,j}^t}\},\{ \boldsymbol{z}_i^t\}) - {F_\mu }^*} \right)}}{{\sqrt {T(\epsilon) - {T_1}} }} \vspace{2mm}\\
 &
 + \frac{{{L^2}}}{4}\sum\limits_{i = 1}^3 {\frac{{{{({d_i} + 6)}^3}}}{{{d_i} + 4}}} \frac{1}{{\sqrt {T(\epsilon) - {T_1}} }}  
 + {L^2}\sum\limits_{i = 1}^3 {{{({d_i} + 3)}^3}} \frac{1}{{\sqrt {T(\epsilon) - {T_1}} }} \vspace{2mm} \\
&
 + \sum\limits_{i = 1}^3 \left({\max \left\{ {8L({d_1} + 4),8L({d_2} + 4),8L({d_3} + 4), \frac{2(L+1)}{3}} \right\} + 1}\right) 4 L (N+1)d_i \frac{1}{{\sqrt {T(\epsilon)-T_1} }}.
\end{split}
\end{equation}

Combining the definition of stationarity gap and $\epsilon$-stationary point in Definition \ref{def:1}, \ref{def:2} with Eq. (\ref{eq:4_19_20}), we have that,
\begin{equation}
    \begin{split}
&||{{\cal G}^{T(\epsilon)}}|{|^2}  \vspace{2mm}\\
& = \sum\limits_{i = 1}^3 \sum\limits_{j = 1}^N  {||{\nabla _{{x_{i,j}}}}F(\{{\boldsymbol{x}_{i,j}^{T(\epsilon)}}\},\{ \boldsymbol{z}_i^{T(\epsilon)}\})|{|^2}} 
 + \sum\limits_{i=1}^3 ||{\nabla _{{\boldsymbol{z}_i}}}{F }(\{{\boldsymbol{x}_{i,j}^{{T(\epsilon)}}}\},\{ \boldsymbol{z}_i^{T(\epsilon)}\})||^2
 \vspace{2mm}  \\
 &
 \le \frac{1}{{T(\epsilon) - {T_1}}}\sum\limits_{t = {T_1}}^{T(\epsilon) - 1} (\sum\limits_{i = 1}^3 \sum\limits_{j = 1}^N  {||{\nabla _{{x_{i,j}}}}F(\{{\boldsymbol{x}_{i,j}^t}\},\{ \boldsymbol{z}_i^t\})|{|^2}} 
 + \sum\limits_{i=1}^3 ||{\nabla _{{\boldsymbol{z}_i}}}{F }(\{{\boldsymbol{x}_{i,j}^{t}}\},\{ \boldsymbol{z}_i^t\})||^2) \vspace{2mm} \\
 &
 \le \frac{4(1 + {\max \left\{ {8L({d_1} + 4),8L({d_2} + 4),8L({d_3} + 4), \frac{2(L+1)}{3}} \right\})\left( {\mathop {\max }\limits_{t \in [{T_1}]} {F_\mu }(\{{\boldsymbol{x}_{i,j}^t}\},\{ \boldsymbol{z}_i^t\}) - {F_\mu }^*} \right)}}{{\sqrt {T(\epsilon) - {T_1}} }} \vspace{2mm}\\
& 
 + \frac{{{L^2}}}{4}\sum\limits_{i = 1}^3 {\frac{{{{({d_i} + 6)}^3}}}{{{d_i} + 4}}} \frac{1}{{\sqrt {T(\epsilon) - {T_1}} }}  
 + {L^2}\sum\limits_{i = 1}^3 {{{({d_i} + 3)}^3}} \frac{1}{{\sqrt {T(\epsilon) - {T_1}} }} \vspace{2mm} \\
&
 + \sum\limits_{i = 1}^3 \left({\max \left\{ {8L({d_1} + 4),8L({d_2} + 4),8L({d_3} + 4), \frac{2(L+1)}{3}} \right\} + 1}\right) 4 L (N+1)d_i \frac{1}{{\sqrt {T(\epsilon)-T_1} }}.
\end{split}
\end{equation}

Thus, we can conclude that, when 
\begin{equation}
    \begin{array}{l}
T(\epsilon)
 \ge {\left( {\sum\limits_{i = 1}^3 {\overline{c_i} + \overline{d}\left( {\mathop {\max }\limits_{t \in [{T_1}]} {F_\mu }(\{{\boldsymbol{x}_{i,j}^t}\},\{ \boldsymbol{z}_i^t\}) - {F_\mu }^*} \right)} } \right)^2}\frac{1}{{{{\epsilon}^2}}} + {T_1}
\end{array},
\end{equation}

we have that  $||{{\cal G}^{T(\epsilon)}}|{|^2} \le \epsilon$, where constants
\begin{equation}
    \overline{d}=4(1 + {\max \left\{ {8L({d_1} + 4),8L({d_2} + 4),8L({d_3} + 4), \frac{2(L+1)}{3}} \right\})},
\end{equation}
\begin{equation}
\begin{array}{l}
    \overline{c_i} = {\frac{{{L^2}{{({d_i} + 6)}^3}}}{{4\left( {{d_i} + 4} \right)}} + {L^2}{{({d_i} + 3)}^3}} \vspace{2mm}\\
    \qquad + 4 L (N+1)d_i \left({\max \left\{ {8L({d_1} + 4),8L({d_2} + 4),8L({d_3} + 4), \frac{2(L+1)}{3}} \right\} + 1}\right) .
\end{array}
\end{equation}

\section{Communication Complexity}
\label{appendix:communication_complexity}
The overall communication complexity of the proposed DTZO can be divided into 1) the communication complexity at every communication round and 2) the communication complexity of updating zeroth order cuts, which is discussed as follows.

1) The communication complexity at each iteration. 

At each iteration, e.g., $(t+1)^{\rm{th}}$ iteration, the workers transmit the updated variables ${\boldsymbol{x}_{1,j}^{t+1}},{\boldsymbol{x}_{2,j}^{t+1}},{\boldsymbol{x}_{3,j}^{t+1}}$ to the master, resulting in a communication complexity of $\sum_{j=1}^N \sum_{i=1}^3 d_i$. Upon receiving these updated local variables, the master proceeds to update the global variables. Then, the master broadcasts the updated variables $\boldsymbol{z}_1^{t+1}, \boldsymbol{z}_2^{t+1}, \boldsymbol{z}_3^{t+1}$ and gradients $\nabla_{\boldsymbol{x}_{i,j}} o( \{{\boldsymbol{x}_{2,j}^{t+1}}\}, \{{\boldsymbol{x}_{3,j}^{t+1}}\}, \boldsymbol{z}_1^{t+1}, \boldsymbol{z}_2^{t+1}, \boldsymbol{z}_3^{t+1}), i=2,3$ to worker $j$. Therefore, the cumulative communication complexity from $t=1$ to $t=T(\epsilon)$ is
\begin{equation}
\label{eq:5_13_92}
C_1 = T(\epsilon)(2d_1 + 3d_2 +3d_3)N.
\end{equation}

2) The communication complexity of updating zeroth order cuts. 

During every iteration $\mathcal{T}$ ($t < T_1$), the cutting planes are updated to refine the cascaded polynomial approximation, involving two main steps: 

2a) Updating the inner layer polynomial approximation: In this phase, local variables ${\boldsymbol{x}_{3,j}^{k+1}}$ are transmitted from worker $j$, while global variables $\boldsymbol{z}_3^{k+1}$ are sent from the master in the $(k+1)^{\rm{th}}$ iteration. The communication complexity associated with updating the inner layer polynomial approximation can be expressed as follows:
\begin{equation}
\label{eq:5_13_93}
\sum\nolimits_{j=1}^N 2 \lfloor \frac{T_1}{\mathcal{T}} \rfloor  \mathcal{T} K d_3.
\end{equation}


2b) Updating the outer layer polynomial approximation: During the $(k+1)^{\rm{th}}$ iteration when updating the outer layer approximation, the worker $j$ transmits the updated variables ${\boldsymbol{x}_{2,j}^{k+1}}$, to the master. Subsequently, the master broadcasts the updated global variables $\boldsymbol{z}_2^{k+1}$ to worker $j$. The communication complexity involved in this process can be expressed as,
\begin{equation}
\label{eq:5_13_94}
    \sum\nolimits_{j=1}^N  2 \lfloor \frac{T_1}{\mathcal{T}} \rfloor  \mathcal{T} Kd_2.
\end{equation}

Combining Eq. (\ref{eq:5_13_93}) with (\ref{eq:5_13_94}), and considering utilizing one communication round to approximate the $\phi_{\rm{in}}( \{{\boldsymbol{x}_{3,j}}\}, \boldsymbol{z}_1, \boldsymbol{z}_2, \boldsymbol{z}_3)$ and $\phi_{\rm{out}}( \{{\boldsymbol{x}_{2,j}}\}, \{{\boldsymbol{x}_{3,j}}\}, \boldsymbol{z}_1, \boldsymbol{z}_2, \boldsymbol{z}_3)$, i.e., $K=1$, we have that the communication complexity of updating cascaded polynomial approximation is,
\begin{equation}
\label{eq:5_13_95}
    C_2 = 2 N \lfloor \frac{T_1}{\mathcal{T}} \rfloor  \mathcal{T} (d_2 + d_3).
\end{equation}

Consequently, the overall communication of the proposed method is $C_1 + C_2$, which can be expressed as,
\begin{equation}
    3T(\epsilon)(d_1 + d_2 +d_3)N +   2 N \lfloor \frac{T_1}{\mathcal{T}} \rfloor  \mathcal{T} (d_2 + d_3).
\end{equation}

\section{Proof of Proposition \ref{prop:1} and \ref{prop:2}}
\label{appendix:pro1_2}

\subsection{Proof of Proposition \ref{prop:1}}

For any point $(\{{\boldsymbol{x}_{3,j}}\}, \boldsymbol{z}_1, {\boldsymbol{z}_2}', \boldsymbol{z}_3)$ in the original feasible region, i.e., $\phi_{\rm{in}}( \{{\boldsymbol{x}_{3,j}}\}, \boldsymbol{z}_1, {\boldsymbol{z}_2}', \boldsymbol{z}_3) = 0$, according to the properties of $L$-smoothness, we have that,
\begin{equation}
\label{eq:5_15_62}
\begin{array}{l}
\phi_{\rm{in}}(\{{\boldsymbol{x}_{3,j}}\}, \boldsymbol{z}_1, {\boldsymbol{z}_2}', \boldsymbol{z}_3)\\
 \ge \phi_{\rm{in}}( \{{\boldsymbol{x}_{3,j}^t}\}, \boldsymbol{z}_1^t, {\boldsymbol{z}_2^t}', \boldsymbol{z}_3^t) + {\frac{{\partial \phi_{\rm{in}}( \{{\boldsymbol{x}_{3,j}^t}\}, \boldsymbol{z}_1^t, {\boldsymbol{z}_2^t}', \boldsymbol{z}_3^t)}}{\partial (\{{\boldsymbol{x}_{3,j}}\}, \boldsymbol{z}_1, {\boldsymbol{z}_2}', \boldsymbol{z}_3)}^{\top}}\left( {\left[ \begin{array}{l}
\{{\boldsymbol{x}_{3,j}}\}\\
\boldsymbol{z}_1\\
{\boldsymbol{z}_2}'\\
\boldsymbol{z}_3
\end{array} \right] - \left[ \begin{array}{l}
\{{\boldsymbol{x}_{3,j}^t}\}\\
\boldsymbol{z}_1^t\\
{\boldsymbol{z}_2^t}'\\
\boldsymbol{z}_3^t
\end{array} \right]} \right) \\
- \frac{L}{2}||\left( {\left[ \begin{array}{l}
\{{\boldsymbol{x}_{3,j}}\}\\
\boldsymbol{z}_1\\
{\boldsymbol{z}_2}'\\
\boldsymbol{z}_3
\end{array} \right] - \left[ \begin{array}{l}
\{{\boldsymbol{x}_{3,j}^t}\}\\
\boldsymbol{z}_1^t\\
{\boldsymbol{z}_2^t}'\\
\boldsymbol{z}_3^t
\end{array} \right]} \right)|{|^2}\\
 = \phi_{\rm{in}}( \{{\boldsymbol{x}_{3,j}^t}\}, \boldsymbol{z}_1^t, {\boldsymbol{z}_2^t}', \boldsymbol{z}_3^t) + {G_\mu^{\rm{in}} }{( \{{\boldsymbol{x}_{3,j}^t}\},\! \boldsymbol{z}_1^t, {\boldsymbol{z}_2^t}', \boldsymbol{z}_3^t)^{\top}}\left( {\left[ \begin{array}{l}
\{{\boldsymbol{x}_{3,j}}\}\\
\boldsymbol{z}_1\\
{\boldsymbol{z}_2}'\\
\boldsymbol{z}_3
\end{array} \right] - \left[ \begin{array}{l}
\{{\boldsymbol{x}_{3,j}^t}\}\\
\boldsymbol{z}_1^t\\
{\boldsymbol{z}_2^t}'\\
\boldsymbol{z}_3^t
\end{array} \right]} \right) \\ + {\left( {\frac{{\partial \phi_{\rm{in}}( \{{\boldsymbol{x}_{3,j}^t}\}, \boldsymbol{z}_1^t, {\boldsymbol{z}_2^t}', \boldsymbol{z}_3^t)}}{\partial (\{{\boldsymbol{x}_{3,j}}\}, \boldsymbol{z}_1, {\boldsymbol{z}_2}', \boldsymbol{z}_3)} - {G_\mu^{\rm{in}} }{( \{{\boldsymbol{x}_{3,j}^t}\},\! \boldsymbol{z}_1^t, {\boldsymbol{z}_2^t}', \boldsymbol{z}_3^t)}} \right)^{\top}}\left( {\left[ \begin{array}{l}
\{{\boldsymbol{x}_{3,j}}\}\\
\boldsymbol{z}_1\\
{\boldsymbol{z}_2}'\\
\boldsymbol{z}_3
\end{array} \right] - \left[ \begin{array}{l}
\{{\boldsymbol{x}_{3,j}^t}\}\\
\boldsymbol{z}_1^t\\
{\boldsymbol{z}_2^t}'\\
\boldsymbol{z}_3^t
\end{array} \right]} \right) \\
- \frac{L}{2}||\left( {\left[ \begin{array}{l}
\{{\boldsymbol{x}_{3,j}}\}\\
\boldsymbol{z}_1\\
{\boldsymbol{z}_2}'\\
\boldsymbol{z}_3
\end{array} \right] - \left[ \begin{array}{l}
\{{\boldsymbol{x}_{3,j}^t}\}\\
\boldsymbol{z}_1^t\\
{\boldsymbol{z}_2^t}'\\
\boldsymbol{z}_3^t
\end{array} \right]} \right)|{|^2}.
\end{array}
\end{equation}

According to $\mathbb{E}[{G_\mu^{\rm{in}} }{( \{{\boldsymbol{x}_{3,j}^t}\},\! \boldsymbol{z}_1^t, {\boldsymbol{z}_2^t}', \boldsymbol{z}_3^t)}] = \phi_{\mu, \rm{in}}( \{{\boldsymbol{x}_{3,j}^t}\}, \boldsymbol{z}_1^t, {\boldsymbol{z}_2^t}', \boldsymbol{z}_3^t)$, taking expectation on both sides of Eq. (\ref{eq:5_15_62}), we have that,
\begin{equation}
 \begin{array}{l}
\mathbb{E}[\phi_{\rm{in}}( \{{\boldsymbol{x}_{3,j}}\}, \boldsymbol{z}_1, {\boldsymbol{z}_2}', \boldsymbol{z}_3)]\\
 \ge \mathbb{E}[\phi_{\rm{in}}( \{{\boldsymbol{x}_{3,j}^t}\}, \boldsymbol{z}_1^t, {\boldsymbol{z}_2^t}', \boldsymbol{z}_3^t)] + \phi_{\mu, \rm{in}}{( \{{\boldsymbol{x}_{3,j}^t}\}, \boldsymbol{z}_1^t, {\boldsymbol{z}_2^t}', \boldsymbol{z}_3^t)^{\top}}\left( {\left[ \begin{array}{l}
\{{\boldsymbol{x}_{3,j}}\}\\
\boldsymbol{z}_1\\
{\boldsymbol{z}_2}'\\
\boldsymbol{z}_3
\end{array} \right] - \left[ \begin{array}{l}
\{{\boldsymbol{x}_{3,j}^t}\}\\
\boldsymbol{z}_1^t\\
{\boldsymbol{z}_2^t}'\\
\boldsymbol{z}_3^t
\end{array} \right]} \right) \\ + {\left( {\frac{{\partial \phi_{\rm{in}}( \{{\boldsymbol{x}_{3,j}^t}\}, \boldsymbol{z}_1^t, {\boldsymbol{z}_2^t}', \boldsymbol{z}_3^t)}}{\partial (\{{\boldsymbol{x}_{3,j}}\}, \boldsymbol{z}_1, {\boldsymbol{z}_2}', \boldsymbol{z}_3)} - \phi_{\mu, \rm{in}}( \{{\boldsymbol{x}_{3,j}^t}\}, \boldsymbol{z}_1^t, {\boldsymbol{z}_2^t}', \boldsymbol{z}_3^t)} \right)^{\top}}\left( {\left[ \begin{array}{l}
\{{\boldsymbol{x}_{3,j}}\}\\
\boldsymbol{z}_1\\
{\boldsymbol{z}_2}'\\
\boldsymbol{z}_3
\end{array} \right] - \left[ \begin{array}{l}
\{{\boldsymbol{x}_{3,j}^t}\}\\
\boldsymbol{z}_1^t\\
{\boldsymbol{z}_2^t}'\\
\boldsymbol{z}_3^t
\end{array} \right]} \right) \\
- \frac{L}{2}||\left( {\left[ \begin{array}{l}
\{{\boldsymbol{x}_{3,j}}\}\\
\boldsymbol{z}_1\\
{\boldsymbol{z}_2}'\\
\boldsymbol{z}_3
\end{array} \right] - \left[ \begin{array}{l}
\{{\boldsymbol{x}_{3,j}^t}\}\\
\boldsymbol{z}_1^t\\
{\boldsymbol{z}_2^t}'\\
\boldsymbol{z}_3^t
\end{array} \right]} \right)|{|^2}.
\end{array}
\end{equation}

Combining with the Cauchy-Schwarz inequality, we have that,
\begin{equation}
\label{eq:4_25_21}
    \begin{array}{l}
\mathbb{E}[\phi_{\rm{in}}( \{{\boldsymbol{x}_{3,j}}\}, \boldsymbol{z}_1, {\boldsymbol{z}_2}', \boldsymbol{z}_3)]\\
 \ge \mathbb{E}[\phi_{\rm{in}}( \{{\boldsymbol{x}_{3,j}^t}\}, \boldsymbol{z}_1^t, {\boldsymbol{z}_2^t}', \boldsymbol{z}_3^t)] + \phi_{\mu, \rm{in}}{( \{{\boldsymbol{x}_{3,j}^t}\}, \boldsymbol{z}_1^t, {\boldsymbol{z}_2^t}', \boldsymbol{z}_3^t)^{\top}}\left( {\left[ \begin{array}{l}
\{{\boldsymbol{x}_{3,j}}\}\\
\boldsymbol{z}_1\\
{\boldsymbol{z}_2}'\\
\boldsymbol{z}_3
\end{array} \right] - \left[ \begin{array}{l}
\{{\boldsymbol{x}_{3,j}^t}\}\\
\boldsymbol{z}_1^t\\
{\boldsymbol{z}_2^t}'\\
\boldsymbol{z}_3^t
\end{array} \right]} \right) \\- \frac{1}{2}||\frac{{\partial \phi_{\rm{in}}( \{{\boldsymbol{x}_{3,j}^t}\}, \boldsymbol{z}_1^t, {\boldsymbol{z}_2^t}', \boldsymbol{z}_3^t)}}{\partial (\{{\boldsymbol{x}_{3,j}}\}, \boldsymbol{z}_1, {\boldsymbol{z}_2}', \boldsymbol{z}_3)} \!-\! \phi_{\mu, \rm{in}}( \{{\boldsymbol{x}_{3,j}^t}\}, \boldsymbol{z}_1^t, {\boldsymbol{z}_2^t}', \boldsymbol{z}_3^t)|{|^2} \!-\! \frac{L+1}{2}||\left( {\left[ \begin{array}{l}
\{{\boldsymbol{x}_{3,j}}\}\\
\boldsymbol{z}_1\\
{\boldsymbol{z}_2}'\\
\boldsymbol{z}_3
\end{array} \right] \!-\! \left[ \begin{array}{l}
\{{\boldsymbol{x}_{3,j}^t}\}\\
\boldsymbol{z}_1^t\\
{\boldsymbol{z}_2^t}'\\
\boldsymbol{z}_3^t
\end{array} \right]} \right)|{|^2}.
\end{array}
\end{equation}

And according to Eq. (3.6) in \cite{ghadimi2013stochastic}, we can obtain that,
\begin{equation}
\label{eq:4_25_22}
    ||\phi_{\mu, \rm{in}}( \{{\boldsymbol{x}_{3,j}^t}\}, \boldsymbol{z}_1^t, {\boldsymbol{z}_2^t}', \boldsymbol{z}_3^t) - \frac{{\partial \phi_{\rm{in}}( \{{\boldsymbol{x}_{3,j}^t}\}, \boldsymbol{z}_1^t, {\boldsymbol{z}_2^t}', \boldsymbol{z}_3^t)}}{\partial (\{{\boldsymbol{x}_{3,j}}\}, \boldsymbol{z}_1, \boldsymbol{z}_2, \boldsymbol{z}_3)}||^2 \le \frac{\mu^2 }{4}L^2{({d_1}\!+\!  {d_2} \!+\! (N\!+\!1){d_3}   \!+\! 3)^{3}}.
\end{equation}

By combining Eq. (\ref{eq:4_25_21}) with Eq. (\ref{eq:4_25_22}), we have that,
\begin{equation}
    \begin{array}{l}
\mathbb{E}[\phi_{\rm{in}}( \{{\boldsymbol{x}_{3,j}}\}, \boldsymbol{z}_1, {\boldsymbol{z}_2}', \boldsymbol{z}_3)]\\
 \ge \mathbb{E}[\phi_{\rm{in}}( \{{\boldsymbol{x}_{3,j}^t}\}, \boldsymbol{z}_1^t, {\boldsymbol{z}_2^t}', \boldsymbol{z}_3^t)] + \phi_{\mu, \rm{in}}{( \{{\boldsymbol{x}_{3,j}^t}\}, \boldsymbol{z}_1^t, {\boldsymbol{z}_2^t}', \boldsymbol{z}_3^t)^{\top}}\left( {\left[ \begin{array}{l}
\{{\boldsymbol{x}_{3,j}}\}\\
\boldsymbol{z}_1\\
{\boldsymbol{z}_2}'\\
\boldsymbol{z}_3
\end{array} \right] - \left[ \begin{array}{l}
\{{\boldsymbol{x}_{3,j}^t}\}\\
\boldsymbol{z}_1^t\\
{\boldsymbol{z}_2^t}'\\
\boldsymbol{z}_3^t
\end{array} \right]} \right) \\- \frac{\mu^2 }{8}L^2{({d_1}\!+\!  {d_2} \!+\! (N\!+\!1){d_3}   \!+\! 3)^{3}}  - \frac{L+1}{2}||\left( {\left[ \begin{array}{l}
\{{\boldsymbol{x}_{3,j}}\}\\
\boldsymbol{z}_1\\
{\boldsymbol{z}_2}'\\
\boldsymbol{z}_3
\end{array} \right] - \left[ \begin{array}{l}
\{{\boldsymbol{x}_{3,j}^t}\}\\
\boldsymbol{z}_1^t\\
{\boldsymbol{z}_2^t}'\\
\boldsymbol{z}_3^t
\end{array} \right]}\right)|{|^2}.
\end{array}
\end{equation}

For any point belongs to the original feasible region, i.e., $\phi_{\rm{in}}( \{{\boldsymbol{x}_{3,j}}\}, \boldsymbol{z}_1, {\boldsymbol{z}_2}', \boldsymbol{z}_3) = 0$, according to $\varepsilon_{\rm{in}} \ge 0$, we can obtain that it also satisfies that,
\begin{equation}
\label{eq:5_20_68}
\begin{array}{l}
    \mathbb{E}[\phi_{\rm{in}}( \{{\boldsymbol{x}_{3,j}^t}\}, \boldsymbol{z}_1^t, {\boldsymbol{z}_2^t}', \boldsymbol{z}_3^t)] + \mathbb{E}[{G_\mu^{\rm{in}}}{( \{{\boldsymbol{x}_{3,j}^t}\},\! \boldsymbol{z}_1^t, {\boldsymbol{z}_2^t}', \boldsymbol{z}_3^t)]^{\top}}\left( {\left[ \begin{array}{l}
\{{\boldsymbol{x}_{3,j}}\}\\
\boldsymbol{z}_1\\
{\boldsymbol{z}_2}'\\
\boldsymbol{z}_3
\end{array} \right] - \left[ \begin{array}{l}
\{{\boldsymbol{x}_{3,j}^t}\}\\
\boldsymbol{z}_1^t\\
{\boldsymbol{z}_2^t}'\\
\boldsymbol{z}_3^t
\end{array} \right]} \right)] \\

\le  \frac{L+1}{2}||\left( {\left[ \begin{array}{l}
\{{\boldsymbol{x}_{3,j}}\}\\
\boldsymbol{z}_1\\
{\boldsymbol{z}_2}'\\
\boldsymbol{z}_3
\end{array} \right] - \left[ \begin{array}{l}
\{{\boldsymbol{x}_{3,j}^t}\}\\
\boldsymbol{z}_1^t\\
{\boldsymbol{z}_2^t}'\\
\boldsymbol{z}_3^t
\end{array} \right]} \right)|{|^2}    
 +  \frac{\mu^2 }{8}L^2{({d_1}\!+\!  {d_2} \!+\! (N\!+\!1){d_3}   \!+\! 3)^{3}} + \varepsilon_{\rm{in}}.
\end{array}
\end{equation}

According to Eq. (\ref{eq:4_25_17}), we can conclude that for any point belongs to the original feasible region of constraint $\phi_{\rm{in}}( \{{\boldsymbol{x}_{3,j}}\}, \boldsymbol{z}_1, {\boldsymbol{z}_2}', \boldsymbol{z}_3) = 0$, it also belongs to
the $P_{\rm{in}}^{t+1}$, that is, the original feasible region is a subset of the feasible region formed by inner layer zeroth order cuts. Let $S_{\rm{in}}$ denote the original feasible region of constraint $\phi_{\rm{in}}( \{{\boldsymbol{x}_{3,j}}\}, \boldsymbol{z}_1, {\boldsymbol{z}_2}', \boldsymbol{z}_3) = 0$, we can obtain that the feasible region formed by inner layer zeroth order cuts will be gradually tightened with zeroth order cuts added according to Eq. (\ref{eq:5_20_68}), that is,
\begin{equation}
    S_{\rm{in}} \subseteq P_{\rm{in}}^{t+1} \subseteq P_{\rm{in}}^{t} \subseteq \cdots \subseteq P_{\rm{in}}^{0}.
\end{equation}

\subsection{Proof of Proposition \ref{prop:2}}
For any point $(\{{\boldsymbol{x}_{2,j}}\},\{{\boldsymbol{x}_{3,j}}\}, \boldsymbol{z}_1, \boldsymbol{z}_2, \boldsymbol{z}_3)$ in the original feasible region, i.e., $\phi_{\rm{out}}(\{{\boldsymbol{x}_{2,j}}\},\{{\boldsymbol{x}_{3,j}}\}, \boldsymbol{z}_1, \boldsymbol{z}_2, \boldsymbol{z}_3) = 0$, according to the properties of $L$-smoothness, we have that,
\begin{equation}
\label{eq:5_15_68}
\begin{array}{l}
\phi_{\rm{out}}(\{{\boldsymbol{x}_{2,j}}\},\{{\boldsymbol{x}_{3,j}}\}, \boldsymbol{z}_1, \boldsymbol{z}_2, \boldsymbol{z}_3)\\
 \ge \phi_{\rm{out}}(\{{\boldsymbol{x}_{2,j}^t}\},\{{\boldsymbol{x}_{3,j}^t}\}, \boldsymbol{z}_1^t, \boldsymbol{z}_2^t, \boldsymbol{z}_3^t) + {\frac{{\partial \phi_{\rm{out}}(\{{\boldsymbol{x}_{2,j}^t}\},\{{\boldsymbol{x}_{3,j}^t}\}, \boldsymbol{z}_1^t, \boldsymbol{z}_2^t, \boldsymbol{z}_3^t)}}{\partial (\{{\boldsymbol{x}_{2,j}}\},\{{\boldsymbol{x}_{3,j}}\}, \boldsymbol{z}_1, \boldsymbol{z}_2, \boldsymbol{z}_3)}^{\top}}\left( {\left[ \begin{array}{l}
\{{\boldsymbol{x}_{2,j}}\}\\ \{{\boldsymbol{x}_{3,j}}\}\\
\boldsymbol{z}_1\\
\boldsymbol{z}_2\\
\boldsymbol{z}_3
\end{array} \right] - \left[ \begin{array}{l}
\{{\boldsymbol{x}_{2,j}^t}\}\\ \{{\boldsymbol{x}_{3,j}^t}\}\\
\boldsymbol{z}_1^t\\
\boldsymbol{z}_2^t\\
\boldsymbol{z}_3^t
\end{array} \right]} \right) \\
- \frac{L}{2}||\left( {\left[ \begin{array}{l}
\{{\boldsymbol{x}_{2,j}}\}\\ \{{\boldsymbol{x}_{3,j}}\}\\
\boldsymbol{z}_1\\
\boldsymbol{z}_2\\
\boldsymbol{z}_3
\end{array} \right] - \left[ \begin{array}{l}
\{{\boldsymbol{x}_{2,j}^t}\}\\ \{{\boldsymbol{x}_{3,j}^t}\}\\
\boldsymbol{z}_1^t\\
\boldsymbol{z}_2^t\\
\boldsymbol{z}_3^t
\end{array} \right]} \right)|{|^2}\\
 = \phi_{\rm{out}}(\{{\boldsymbol{x}_{2,j}^t}\},\{{\boldsymbol{x}_{3,j}^t}\}, \boldsymbol{z}_1^t, \boldsymbol{z}_2^t, \boldsymbol{z}_3^t) + {G_\mu^{\rm{out}} }(t)^{\top}\left( {\left[ \begin{array}{l}
\!\{{\boldsymbol{x}_{2,j}}\}\\ \! \{{\boldsymbol{x}_{3,j}}\}\\
\boldsymbol{z}_1\\
\boldsymbol{z}_2\\
\boldsymbol{z}_3
\end{array} \right] - \left[ \begin{array}{l}
\! \{{\boldsymbol{x}_{2,j}^t}\}\\ \! \{{\boldsymbol{x}_{3,j}^t}\}\\
\boldsymbol{z}_1^t\\
\boldsymbol{z}_2^t\\
\boldsymbol{z}_3^t
\end{array} \right]} \right) \\ 
+ {\left( {\frac{{\partial \phi_{\rm{out}}(\{{\boldsymbol{x}_{2,j}^t}\},\{{\boldsymbol{x}_{3,j}^t}\}, \boldsymbol{z}_1^t, \boldsymbol{z}_2^t, \boldsymbol{z}_3^t)}}{\partial (\{{\boldsymbol{x}_{2,j}}\},\{{\boldsymbol{x}_{3,j}}\}, \boldsymbol{z}_1, \boldsymbol{z}_2, \boldsymbol{z}_3)} - {G_\mu^{\rm{out}} }(t)} \right)^{\top}}\left( {\left[ \begin{array}{l}
\!\{{\boldsymbol{x}_{2,j}}\}\\ \! \{{\boldsymbol{x}_{3,j}}\}\\
\boldsymbol{z}_1\\
\boldsymbol{z}_2\\
\boldsymbol{z}_3
\end{array} \right] - \left[ \begin{array}{l}
\! \{{\boldsymbol{x}_{2,j}^t}\}\\ \! \{{\boldsymbol{x}_{3,j}^t}\}\\
\boldsymbol{z}_1^t\\
\boldsymbol{z}_2^t\\
\boldsymbol{z}_3^t
\end{array} \right]} \right) \\
- \frac{L}{2}||\left( {\left[ \begin{array}{l}
\!\{{\boldsymbol{x}_{2,j}}\}\\ \! \{{\boldsymbol{x}_{3,j}}\}\\
\boldsymbol{z}_1\\
\boldsymbol{z}_2\\
\boldsymbol{z}_3
\end{array} \right] - \left[ \begin{array}{l}
\! \{{\boldsymbol{x}_{2,j}^t}\}\\ \! \{{\boldsymbol{x}_{3,j}^t}\}\\
\boldsymbol{z}_1^t\\
\boldsymbol{z}_2^t\\
\boldsymbol{z}_3^t
\end{array} \right]} \right)|{|^2},
\end{array}
\end{equation}
where ${G_\mu^{\rm{out}} }(t)$ is the simplified form of ${G_\mu^{\rm{out}} }(\{{\boldsymbol{x}_{2,j}^t}\},\{{\boldsymbol{x}_{3,j}^t}\}, \boldsymbol{z}_1^t, \boldsymbol{z}_2^t, \boldsymbol{z}_3^t)$. 
According to $\mathbb{E}[{G_\mu^{\rm{out}} }(t)] = \phi_{\mu, \rm{out}}(\{{\boldsymbol{x}_{2,j}^t}\}, \{{\boldsymbol{x}_{3,j}^t}\}, \boldsymbol{z}_1^t, \boldsymbol{z}_2^t, \boldsymbol{z}_3^t)$, taking expectation on both sides of Eq. (\ref{eq:5_15_68}), we have that,
\begin{equation}
 \begin{array}{l}
\mathbb{E}[\phi_{\rm{out}}(\{{\boldsymbol{x}_{2,j}}\},\{{\boldsymbol{x}_{3,j}}\}, \boldsymbol{z}_1, \boldsymbol{z}_2, \boldsymbol{z}_3)]\\
 \ge \mathbb{E}[\phi_{\rm{out}}(\{{\boldsymbol{x}_{2,j}^t}\},\{{\boldsymbol{x}_{3,j}^t}\}, \boldsymbol{z}_1^t, \boldsymbol{z}_2^t, \boldsymbol{z}_3^t)] + \phi_{\mu, \rm{out}}(t)^{\top}\left( {\left[ \begin{array}{l}
\!\{{\boldsymbol{x}_{2,j}}\}\\ \! \{{\boldsymbol{x}_{3,j}}\}\\
\boldsymbol{z}_1\\
\boldsymbol{z}_2\\
\boldsymbol{z}_3
\end{array} \right] - \left[ \begin{array}{l}
\! \{{\boldsymbol{x}_{2,j}^t}\}\\ \! \{{\boldsymbol{x}_{3,j}^t}\}\\
\boldsymbol{z}_1^t\\
\boldsymbol{z}_2^t\\
\boldsymbol{z}_3^t
\end{array} \right]} \right) \\ + {\left( {\frac{{\partial \phi_{\rm{out}}(\{{\boldsymbol{x}_{2,j}^t}\},\{{\boldsymbol{x}_{3,j}^t}\}, \boldsymbol{z}_1^t, \boldsymbol{z}_2^t, \boldsymbol{z}_3^t)}}{\partial (\{{\boldsymbol{x}_{2,j}}\},\{{\boldsymbol{x}_{3,j}}\}, \boldsymbol{z}_1, \boldsymbol{z}_2, \boldsymbol{z}_3)} - \phi_{\mu, \rm{out}}(t)} \right)^{\top}}\left( {\left[ \begin{array}{l}
\!\{{\boldsymbol{x}_{2,j}}\}\\ \! \{{\boldsymbol{x}_{3,j}}\}\\
\boldsymbol{z}_1\\
\boldsymbol{z}_2\\
\boldsymbol{z}_3
\end{array} \right] - \left[ \begin{array}{l}
\! \{{\boldsymbol{x}_{2,j}^t}\}\\ \! \{{\boldsymbol{x}_{3,j}^t}\}\\
\boldsymbol{z}_1^t\\
\boldsymbol{z}_2^t\\
\boldsymbol{z}_3^t
\end{array} \right]} \right) \\
- \frac{L}{2}||\left( {\left[ \begin{array}{l}
\!\{{\boldsymbol{x}_{2,j}}\}\\ \! \{{\boldsymbol{x}_{3,j}}\}\\
\boldsymbol{z}_1\\
\boldsymbol{z}_2\\
\boldsymbol{z}_3
\end{array} \right] - \left[ \begin{array}{l}
\! \{{\boldsymbol{x}_{2,j}^t}\}\\ \! \{{\boldsymbol{x}_{3,j}^t}\}\\
\boldsymbol{z}_1^t\\
\boldsymbol{z}_2^t\\
\boldsymbol{z}_3^t
\end{array} \right]} \right)|{|^2},
\end{array}
\end{equation}
where $\phi_{\mu, \rm{out}}(t)$ is the simplified form of $\phi_{\mu, \rm{out}}(\{{\boldsymbol{x}_{2,j}^t}\}, \{{\boldsymbol{x}_{3,j}^t}\}, \boldsymbol{z}_1^t, \boldsymbol{z}_2^t, \boldsymbol{z}_3^t)$. 
Combining with the Cauchy-Schwarz inequality, we have that,
\begin{equation}
\label{eq:4_25_68}
    \begin{array}{l}
\mathbb{E}[\phi_{\rm{out}}(\{{\boldsymbol{x}_{2,j}}\},\{{\boldsymbol{x}_{3,j}}\}, \boldsymbol{z}_1, \boldsymbol{z}_2, \boldsymbol{z}_3)]\\
 \ge \mathbb{E}[\phi_{\rm{out}}(\{{\boldsymbol{x}_{2,j}^t}\},\{{\boldsymbol{x}_{3,j}^t}\}, \boldsymbol{z}_1^t, \boldsymbol{z}_2^t, \boldsymbol{z}_3^t)] + \phi_{\mu, \rm{out}}(t)^{\top}\left( {\left[ \begin{array}{l}
\!\{{\boldsymbol{x}_{2,j}}\}\\ \! \{{\boldsymbol{x}_{3,j}}\}\\
\boldsymbol{z}_1\\
\boldsymbol{z}_2\\
\boldsymbol{z}_3
\end{array} \right] - \left[ \begin{array}{l}
\! \{{\boldsymbol{x}_{2,j}^t}\}\\ \! \{{\boldsymbol{x}_{3,j}^t}\}\\
\boldsymbol{z}_1^t\\
\boldsymbol{z}_2^t\\
\boldsymbol{z}_3^t
\end{array} \right]} \right) \\- \frac{1}{2}||\frac{{\partial \phi_{\rm{out}}(\{{\boldsymbol{x}_{2,j}^t}\},\{{\boldsymbol{x}_{3,j}^t}\}, \boldsymbol{z}_1^t, \boldsymbol{z}_2^t, \boldsymbol{z}_3^t)}}{\partial (\{{\boldsymbol{x}_{2,j}}\},\{{\boldsymbol{x}_{3,j}}\}, \boldsymbol{z}_1, \boldsymbol{z}_2, \boldsymbol{z}_3)} - \phi_{\mu, \rm{out}}(t)|{|^2} \\

- \frac{L+1}{2}||\left( {\left[ \begin{array}{l}
\!\{{\boldsymbol{x}_{2,j}}\}\\ \! \{{\boldsymbol{x}_{3,j}}\}\\
\boldsymbol{z}_1\\
\boldsymbol{z}_2\\
\boldsymbol{z}_3
\end{array} \right] - \left[ \begin{array}{l}
\! \{{\boldsymbol{x}_{2,j}^t}\}\\ \! \{{\boldsymbol{x}_{3,j}^t}\}\\
\boldsymbol{z}_1^t\\
\boldsymbol{z}_2^t\\
\boldsymbol{z}_3^t
\end{array} \right]} \right)|{|^2}.
\end{array}
\end{equation}

And according to Eq. (3.6) in \cite{ghadimi2013stochastic}, we can obtain that,
\begin{equation}
\label{eq:4_25_69}
    ||\phi_{\mu, \rm{out}}(t) - \frac{{\partial \phi_{\rm{out}}(\{{\boldsymbol{x}_{2,j}^t}\},\{{\boldsymbol{x}_{3,j}^t}\}, \boldsymbol{z}_1^t, \boldsymbol{z}_2^t, \boldsymbol{z}_3^t)}}{\partial (\{{\boldsymbol{x}_{2,j}}\},\{{\boldsymbol{x}_{3,j}}\}, \boldsymbol{z}_1, \boldsymbol{z}_2, \boldsymbol{z}_3)}||^2 \le \frac{\mu^2 }{4}L^2{({d_1} \!+\! (N\!+\!1)({d_2} \!+\!  {d_3})  \!+\! 3)^{3}}.
\end{equation}

By combining Eq. (\ref{eq:4_25_68}) with Eq. (\ref{eq:4_25_69}), we have that,
\begin{equation}
    \begin{array}{l}
\mathbb{E}[\phi_{\rm{out}}(\{{\boldsymbol{x}_{2,j}}\}, \{{\boldsymbol{x}_{3,j}}\}, \boldsymbol{z}_1, \boldsymbol{z}_2, \boldsymbol{z}_3)]\\
 \ge \mathbb{E}[\phi_{\rm{out}}(\{{\boldsymbol{x}_{2,j}^t}\},\{{\boldsymbol{x}_{3,j}^t}\}, \boldsymbol{z}_1^t, \boldsymbol{z}_2^t, \boldsymbol{z}_3^t)] + \phi_{\mu, \rm{out}}(t)^{\top}\left( {\left[ \begin{array}{l}
\!\{{\boldsymbol{x}_{2,j}}\}\\ \! \{{\boldsymbol{x}_{3,j}}\}\\
\boldsymbol{z}_1\\
\boldsymbol{z}_2\\
\boldsymbol{z}_3
\end{array} \right] - \left[ \begin{array}{l}
\! \{{\boldsymbol{x}_{2,j}^t}\}\\ \! \{{\boldsymbol{x}_{3,j}^t}\}\\
\boldsymbol{z}_1^t\\
\boldsymbol{z}_2^t\\
\boldsymbol{z}_3^t
\end{array} \right]} \right) \\- \frac{\mu^2 }{8}L^2{({d_1} \!+\! (N\!+\!1)({d_2} \!+\! {d_3})   \!+\! 3)^{3}}  - \frac{L+1}{2}||\left( {\left[ \begin{array}{l}
\!\{{\boldsymbol{x}_{2,j}}\}\\ \! \{{\boldsymbol{x}_{3,j}}\}\\
\boldsymbol{z}_1\\
\boldsymbol{z}_2\\
\boldsymbol{z}_3
\end{array} \right] - \left[ \begin{array}{l}
\! \{{\boldsymbol{x}_{2,j}^t}\}\\ \! \{{\boldsymbol{x}_{3,j}^t}\}\\
\boldsymbol{z}_1^t\\
\boldsymbol{z}_2^t\\
\boldsymbol{z}_3^t
\end{array} \right]} \right)|{|^2}.
\end{array}
\end{equation}

For any point belongs to the original feasible region, i.e., $\phi_{\rm{out}}(\{{\boldsymbol{x}_{2,j}}\}, \{{\boldsymbol{x}_{3,j}}\}, \boldsymbol{z}_1, \boldsymbol{z}_2, \boldsymbol{z}_3) = 0$, according to $\varepsilon_{\rm{in}} \ge 0$, we can obtain that it also satisfies that,
\begin{equation}
\label{eq:5_20_75}
\begin{array}{l}
    \phi_{\rm{out}}(\{{\boldsymbol{x}_{2,j}^t}\},\! \{{\boldsymbol{x}_{3,j}^t}\},\! \boldsymbol{z}_1^t, \boldsymbol{z}_2^t, \boldsymbol{z}_3^t) \! +\!  {G_\mu^{\rm{out}} }{(\{{\boldsymbol{x}_{2,j}^t}\},\!\{{\boldsymbol{x}_{3,j}^t}\},\! \boldsymbol{z}_1^t, \boldsymbol{z}_2^t, \boldsymbol{z}_3^t)^{\top}}\left( {\left[ \begin{array}{l}
\!\{{\boldsymbol{x}_{2,j}}\}\\
\!\{{\boldsymbol{x}_{3,j}}\}\\
\boldsymbol{z}_1\\
\boldsymbol{z}_2\\
\boldsymbol{z}_3
\end{array} \right] \!-\! \left[ \begin{array}{l}
\!\{{\boldsymbol{x}_{2,j}^t}\}\\
\!\{{\boldsymbol{x}_{3,j}^t}\}\\
\boldsymbol{z}_1^t\\
\boldsymbol{z}_2^t\\
\boldsymbol{z}_3^t
\end{array} \right]} \right) \vspace{1mm} \\

\le \frac{{L + 1}}{2}  \left( \sum_{i=2}^3\sum_{j}\! || {\boldsymbol{x}_{i,j}} \!-\! {\boldsymbol{x}_{i,j}^t}||^2 \!+\! \sum_{i}\! ||\boldsymbol{z}_i\!-\!\boldsymbol{z}_i^t||^2  \right) \!+\! \frac{\mu^2 }{8}L^2{({d_1} \!+\! (N\!+\!1)({d_2} \!+\! {d_3}) \!+\! 3)^{3}} \!+\! \varepsilon_{\rm{out}}.
\end{array}
\end{equation}

According to Eq. (\ref{eq:4_25_25}), we can conclude that for any point belongs to the original feasible region of constraint $\phi_{\rm{out}}( \{{\boldsymbol{x}_{2,j}}\}, \{{\boldsymbol{x}_{3,j}}\}, \boldsymbol{z}_1, \boldsymbol{z}_2, \boldsymbol{z}_3)=0$, it also belongs to
the $P_{\rm{out}}^{t+1}$, that is, the original feasible region is a subset of the feasible region formed by outer layer zeroth order cuts. In addition, let $S_{\rm{out}}$ denote the original feasible region of constraint $\phi_{\rm{out}}( \{{\boldsymbol{x}_{2,j}}\}, \{{\boldsymbol{x}_{3,j}}\}, \boldsymbol{z}_1, \boldsymbol{z}_2, \boldsymbol{z}_3)=0$, based on Eq. (\ref{eq:5_20_75}), we can obtain that the feasible region formed by outer layer zeroth order cuts will be gradually tightened with zeroth order cuts added, that is,
\begin{equation}
    S_{\rm{out}} \subseteq P_{\rm{out}}^{t+1} \subseteq P_{\rm{out}}^{t} \subseteq \cdots \subseteq P_{\rm{out}}^{0}.
\end{equation}

\section{Theoretical Analyses about the Cascaded Polynomial Approximation Problem}
\label{appendix:cascaded}
In this section, we theoretically analyze the connections between the original distributed trilevel zeroth order optimization problem in Eq. (\ref{eq:1}) and the cascaded polynomial approximation problem in Eq. (\ref{eq:4}). To facilitate this discussion, we start by examining the distributed bilevel zeroth order optimization problem, which can be expressed as follows,
\begin{equation}
\label{eq:5_6_67}
    \begin{array}{l}
\min \sum\limits_{j = 1}^N {{f_{1,j}}({\boldsymbol{x}_1},{\boldsymbol{x}_2})} \\
{\rm{s}}{\rm{.t}}{\rm{.}}\;{\boldsymbol{x}_2} = \mathop {\arg \min }\limits_{{\boldsymbol{x}_2}'} \sum\limits_{j = 1}^N {{f_{2,j}}({\boldsymbol{x}_1},{\boldsymbol{x}_2}')} \\
{\mathop{\rm var}} .\qquad \quad {\boldsymbol{x}_1},{\boldsymbol{x}_2}.
\end{array}
\end{equation}

The optimization problem in Eq. (\ref{eq:5_6_67}) can be equivalently reformulated as,
\begin{equation}
\label{eq:5_6_68}
    \begin{array}{l}
\min \sum\limits_{j = 1}^N {{f_{1,j}}({\boldsymbol{x}_{1,j}},{\boldsymbol{x}_{2,j}})} \\
{\rm{s}}{\rm{.t}}{\rm{.}}\; {\boldsymbol{x}_{1,j}} = \boldsymbol{z}_1, \forall j =1,\cdots,N 

\\ \quad \;\; \{{\boldsymbol{x}_{2,j}}\},\boldsymbol{z}_2 = \mathop {\arg \min }\limits_{\{{\boldsymbol{x}_{2,j}}'\},{\boldsymbol{z}_2}'} \sum\limits_{j = 1}^N {{f_{2,j}}(\boldsymbol{z}_1,{\boldsymbol{x}_{2,j}}')} \\

\qquad \;  \quad \;\; \qquad {\rm{s}}{\rm{.t}}{\rm{.}}\; {\boldsymbol{x}_{2,j}}' = {\boldsymbol{z}_2}', \forall j =1,\cdots,N  
\vspace{1mm} \\
{\mathop{\rm var}} .\qquad \qquad \{{\boldsymbol{x}_{1,j}}\},\{{\boldsymbol{x}_{2,j}}\}, \boldsymbol{z}_1, \boldsymbol{z}_2.
\end{array}
\end{equation}

By utilizing the proposed polynomial approximation with zeroth order cut, we can obtain the following zeroth order polynomial approximation problem,
\begin{equation}
\label{eq:5_6_69}
    \begin{array}{l}
\min \sum\limits_{j = 1}^N {{f_{1,j}}({\boldsymbol{x}_{1,j}},{\boldsymbol{x}_{2,j}})} \\

{\rm{s}}{\rm{.t}}{\rm{.}}\; {\boldsymbol{x}_{1,j}} = \boldsymbol{z}_1, \forall j =1,\cdots,N 

\\  \sum\limits_{j=1}^{N}\!{{{\boldsymbol{a}_{2,j,l}^{\rm{}}}^{\top}\boldsymbol{x}_{2,j}^2} \!+\! {{\boldsymbol{b}_{2,j,l}^{\rm{}}}^{\top}\boldsymbol{x}_{2,j}}}  \!+\!  \sum\limits_{i=1}^{2} \!{{\boldsymbol{c}_{i,l}^{\rm{}}}^{\top}\boldsymbol{z}_{i}^2} \!+\! {{\boldsymbol{d}_{i,l}^{\rm{}}}^{\top}\boldsymbol{z}_{i}} \!+\! e_{l}^{\rm{}} \!\le\! \varepsilon_{\rm{}}, \forall l 

\\ {\mathop{\rm var}} .\qquad  \{{\boldsymbol{x}_{1,j}}\},\{{\boldsymbol{x}_{2,j}}\}, \boldsymbol{z}_1, \boldsymbol{z}_2.

\end{array}
\end{equation}

According to Proposition \ref{prop:1} and \ref{prop:2}, we can obtain the feasible region of the problem in Eq. (\ref{eq:5_6_68}) is a subset of the feasible region of the problem in Eq. (\ref{eq:5_6_69}). Thus, we can conclude that the zeroth order polynomial approximation optimization problem in Eq. (\ref{eq:5_6_69}) is the relaxed problem of the distributed bilevel zeroth order optimization problem in Eq. (\ref{eq:5_6_67}).

For the distributed trilevel zeroth order optimization problem, we first define the following feasible regions.
\begin{equation}
\label{eq:5_6_70}
S_1 = 
     \left\{\{{\boldsymbol{x}_{i,j}}\},\!\{\boldsymbol{z}_i\}| \begin{array}{l}
h_l^{\rm{out}}( \{{\boldsymbol{x}_{2,j}}\},\! \{{\boldsymbol{x}_{3,j}}\},\! \boldsymbol{z}_1, \boldsymbol{z}_2, \boldsymbol{z}_3) \le \varepsilon_{\rm{out}}, \forall l, \\
\boldsymbol{z}_1 = \boldsymbol{x}_{1,j}, \forall j
\end{array}    \right\}  ,  
\end{equation}

\begin{equation}
\label{eq:5_6_71}
\begin{array}{l}
S_2 = \\

\left\{\{{\boldsymbol{x}_{i,j}}\},\!\{\boldsymbol{z}_i\}|  \begin{array}{l}
    || \left[ \begin{array}{l}
\{ {\boldsymbol{x}_{2,j}}\} \\
{\boldsymbol{z}_2} 
\end{array} \right] - \begin{array}{l}
\mathop {\arg \min }\limits_{\{{\boldsymbol{x}_{2,j}}'\},{\boldsymbol{z}_2}'} \sum\limits_{j = 1}^N {{f_{2,j}}(\boldsymbol{z}_1,{\boldsymbol{x}_{2,j}}',{\boldsymbol{x}_{3,j}})} \\
{\rm{s.t.}}\quad {\boldsymbol{x}_{2,j}}' = {\boldsymbol{z}_2}',\forall j, \\ \qquad h_l^{\rm{in}}( \{{\boldsymbol{x}_{3,j}}\}, \boldsymbol{z}_1, {\boldsymbol{z}_2}', \boldsymbol{z}_3) \! \le\! \varepsilon_{\rm{in}},\forall l
\end{array}  ||^2  \le \varepsilon_{\rm{out}}, \vspace{1mm} \\

\quad  \boldsymbol{z}_1 = \boldsymbol{x}_{1,j}, \forall j
\end{array}
  \right\}  ,
\end{array}
\end{equation}

\begin{equation}
\label{eq:5_6_72}
\begin{array}{l}
S_3 = \\

\left\{\{{\boldsymbol{x}_{i,j}}\},\!\{\boldsymbol{z}_i\}|  \begin{array}{l}
    || \left[ \begin{array}{l}
\{ {\boldsymbol{x}_{2,j}}\} \\
{\boldsymbol{z}_2}  
\end{array} \right] - \begin{array}{l}
\mathop {\arg \min }\limits_{\{{\boldsymbol{x}_{2,j}}'\},{\boldsymbol{z}_2}'} \sum\limits_{j = 1}^N {{f_{2,j}}(\boldsymbol{z}_1,{\boldsymbol{x}_{2,j}},{\boldsymbol{x}_{3,j}})} \\

{\rm{s}}{\rm{.t}}{\rm{.}}\; {\boldsymbol{x}_{2,j}}' = {\boldsymbol{z}_2}', \forall j =1,\cdots,N  

\\ \quad \; \{{\boldsymbol{x}_{3,j}}\}, \boldsymbol{z}_3 {\rm{ =  }}\mathop {\arg \min }\limits_{\{{\boldsymbol{x}_{3,j}}'\}, {\boldsymbol{z}_3}'} \sum\limits_{j = 1}^N {{f_{3,j}}(\boldsymbol{z}_1,{\boldsymbol{z}_2}',{\boldsymbol{x}_{3,j}}')} 
\\ \qquad \qquad \quad \; {\rm{s}}{\rm{.t}}{\rm{.}}\; {\boldsymbol{x}_{3,j}}' = {\boldsymbol{z}_3}', \forall j =1,\cdots,N  
\end{array}  ||^2  = 0, \vspace{1mm} \\

\quad  \boldsymbol{z}_1 = \boldsymbol{x}_{1,j}, \forall j
\end{array}
  \right\}  .
\end{array}
\end{equation}

It is seen from Eq. (\ref{eq:5_6_70}) and Eq. (\ref{eq:5_6_72}) that $S_1$ and $S_3$ respectively represent the feasible region of optimization problems in Eq. (\ref{eq:4}) and Eq. (\ref{eq:2}). For any feasible solution $\{{\hat{\boldsymbol{x}}_{i,j}}\},\!\{\hat{\boldsymbol{z}}_i\}$ of optimization problem in Eq. (\ref{eq:2}), it satisfies that,
\begin{equation}
\label{eq:5_6_73}
     || \left[ \begin{array}{l}
\{ {\hat{\boldsymbol{x}}_{2,j}}\} \\
{\hat{\boldsymbol{z}}_2} 
\end{array} \right] - \begin{array}{l}
\mathop {\arg \min }\limits_{\{{{\boldsymbol{x}}_{2,j}}'\},{\boldsymbol{z}_2}'} \sum\limits_{j = 1}^N {{f_{2,j}}(\hat{\boldsymbol{z}}_1,{{\boldsymbol{x}}_{2,j}}',{\hat{\boldsymbol{x}}_{3,j}})} \\

{\rm{s}}{\rm{.t}}{\rm{.}}\; {{\boldsymbol{x}}_{2,j}}' = {\boldsymbol{z}_2}', \forall j =1,\cdots,N  

\\ \quad \; \{{\hat{\boldsymbol{x}}_{3,j}}\}, \hat{\boldsymbol{z}}_3 {\rm{ =  }}\mathop {\arg \min }\limits_{\{{{\boldsymbol{x}}_{3,j}}'\}, {\boldsymbol{z}_3}'} \sum\limits_{j = 1}^N {{f_{3,j}}(\hat{\boldsymbol{z}}_1,{\boldsymbol{z}_2}',{{\boldsymbol{x}}_{3,j}}')} 
\\ \qquad \qquad \quad \; {\rm{s}}{\rm{.t}}{\rm{.}}\; {{\boldsymbol{x}}_{3,j}}' = {\boldsymbol{z}_3}', \forall j =1,\cdots,N  
\end{array}  ||^2 = 0 .
\end{equation}

Based on Proposition \ref{prop:1}, we have that the feasible region of constraint $\phi_{\rm{in}}( \{{\boldsymbol{x}_{3,j}}\}, \boldsymbol{z}_1, {\boldsymbol{z}_2}', \boldsymbol{z}_3) = 0$ is a subset of the feasible region formed by inner layer zeroth order cuts, i.e., $\left\{ \{{\boldsymbol{x}_{3,j}}\}, \boldsymbol{z}_1, {\boldsymbol{z}_2}', \boldsymbol{z}_3| \, h_l^{\rm{in}}( \{{\boldsymbol{x}_{3,j}}\}, \boldsymbol{z}_1, {\boldsymbol{z}_2}', \boldsymbol{z}_3) \! \le\! \varepsilon_{\rm{in}},\forall l   \right\}$. Moreover, the feasible region formed by inner layer zeroth order cuts will be continuously tightened with zeroth order cuts added. Thus, let $\beta \ge 0$ satisfy that,
\begin{equation}
\label{eq:5_6_74}
    \begin{array}{l}
||  \begin{array}{l}
\mathop {\arg \min }\limits_{\{{{\boldsymbol{x}}_{2,j}}'\},{\boldsymbol{z}_2}'} \sum\limits_{j = 1}^N {{f_{2,j}}(\hat{\boldsymbol{z}}_1,{{\boldsymbol{x}}_{2,j}}',{\hat{\boldsymbol{x}}_{3,j}})} \\
{\rm{s.t.}}\quad {{\boldsymbol{x}}_{2,j}}' \!=\! {{\boldsymbol{z}}_2}',\forall j, \\ \qquad h_l^{\rm{in}}( \{{\hat{\boldsymbol{x}}_{3,j}}\}, \hat{\boldsymbol{z}}_1, {\boldsymbol{z}_2}', \hat{\boldsymbol{z}}_3) \! \le\! \varepsilon_{\rm{in}},\forall l
\end{array}  -   \begin{array}{l}
\mathop {\arg \min }\limits_{\{{{\boldsymbol{x}}_{2,j}}'\},{\boldsymbol{z}_2}'} \sum\limits_{j = 1}^N {{f_{2,j}}(\hat{\boldsymbol{z}}_1,{{\boldsymbol{x}}_{2,j}}',{\hat{\boldsymbol{x}}_{3,j}})} \\

{\rm{s}}{\rm{.t}}{\rm{.}}\; {{\boldsymbol{x}}_{2,j}}' = {\boldsymbol{z}_2}', \forall j =1,\cdots,N  

\\ \quad \; \{{\hat{\boldsymbol{x}}_{3,j}}\}, \hat{\boldsymbol{z}}_3 {\rm{ =  }}\mathop {\arg \min }\limits_{\{{{\boldsymbol{x}}_{3,j}}'\}, {\boldsymbol{z}_3}'} \sum\limits_{j = 1}^N {{f_{3,j}}(\hat{\boldsymbol{z}}_1,{\boldsymbol{z}_2}',{{\boldsymbol{x}}_{3,j}}')} 
\\ \qquad \qquad \quad \; {\rm{s}}{\rm{.t}}{\rm{.}}\; {{\boldsymbol{x}}_{3,j}}' = {\boldsymbol{z}_3}', \forall j =1,\cdots,N  
\end{array} ||^2

\\

\le \beta.
    \end{array}
\end{equation}

By combining Proposition \ref{prop:1} with Eq. (\ref{eq:5_6_74}), we can obtain that $\beta$ will continuously decrease with inner layer zeroth order cuts added. By combining Eq. (\ref{eq:5_6_73}) with Cauchy-Schwarz inequality, we can obtain that,
\begin{equation}
\label{eq:5_6_75}
\begin{array}{l}
     || \left[ \begin{array}{l}
\{ {\hat{\boldsymbol{x}}_{2,j}}\} \\
{\hat{\boldsymbol{z}}_2} 
\end{array} \right] - \begin{array}{l}
\mathop {\arg \min }\limits_{\{{{\boldsymbol{x}}_{2,j}}'\},{\boldsymbol{z}_2}'} \sum\limits_{j = 1}^N {{f_{2,j}}(\hat{\boldsymbol{z}}_1,{{\boldsymbol{x}}_{2,j}}',{\hat{\boldsymbol{x}}_{3,j}})} \\
{\rm{s.t.}}\quad {{\boldsymbol{x}}_{2,j}}' \!=\! {{\boldsymbol{z}}_2}',\forall j, \\ \qquad h_l^{\rm{in}}( \{{\hat{\boldsymbol{x}}_{3,j}}\}, \hat{\boldsymbol{z}}_1, {\boldsymbol{z}_2}', \hat{\boldsymbol{z}}_3) \! \le\! \varepsilon_{\rm{in}},\forall l
\end{array}  ||^2  \\

=  || \left[ \begin{array}{l}
\{ {\hat{\boldsymbol{x}}_{2,j}}\} \\
{\hat{\boldsymbol{z}}_2}  
\end{array} \right] - \begin{array}{l}
\mathop {\arg \min }\limits_{\{{{\boldsymbol{x}}_{2,j}}'\},{\boldsymbol{z}_2}'} \sum\limits_{j = 1}^N {{f_{2,j}}(\hat{\boldsymbol{z}}_1,{{\boldsymbol{x}}_{2,j}}',{\hat{\boldsymbol{x}}_{3,j}})} \\

{\rm{s}}{\rm{.t}}{\rm{.}}\; {{\boldsymbol{x}}_{2,j}}' = {\boldsymbol{z}_2}', \forall j =1,\cdots,N  

\\ \quad \; \{{\hat{\boldsymbol{x}}_{3,j}}\}, \hat{\boldsymbol{z}}_3 {\rm{ =  }}\mathop {\arg \min }\limits_{\{{{\boldsymbol{x}}_{3,j}}'\}, {\boldsymbol{z}_3}'} \sum\limits_{j = 1}^N {{f_{3,j}}(\hat{\boldsymbol{z}}_1,{\boldsymbol{z}_2}',{{\boldsymbol{x}}_{3,j}}')} 
\\ \qquad \qquad \quad \; {\rm{s}}{\rm{.t}}{\rm{.}}\; {{\boldsymbol{x}}_{3,j}}' = {\boldsymbol{z}_3}', \forall j =1,\cdots,N  
\end{array} \\

+ \begin{array}{l}
\mathop {\arg \min }\limits_{\{{{\boldsymbol{x}}_{2,j}}'\},{\boldsymbol{z}_2}'} \sum\limits_{j = 1}^N {{f_{2,j}}(\hat{\boldsymbol{z}}_1,{{\boldsymbol{x}}_{2,j}}',{\hat{\boldsymbol{x}}_{3,j}})} \\

{\rm{s}}{\rm{.t}}{\rm{.}}\; {{\boldsymbol{x}}_{2,j}}' = {\boldsymbol{z}_2}', \forall j =1,\cdots,N  

\\ \quad \; \{{\hat{\boldsymbol{x}}_{3,j}}\}, \hat{\boldsymbol{z}}_3 {\rm{ =  }}\mathop {\arg \min }\limits_{\{{{\boldsymbol{x}}_{3,j}}'\}, {\boldsymbol{z}_3}'} \sum\limits_{j = 1}^N {{f_{3,j}}(\hat{\boldsymbol{z}}_1,{\boldsymbol{z}_2}',{{\boldsymbol{x}}_{3,j}}')} 
\\ \qquad \qquad \quad \; {\rm{s}}{\rm{.t}}{\rm{.}}\; {{\boldsymbol{x}}_{3,j}}' = {\boldsymbol{z}_3}', \forall j =1,\cdots,N  
\end{array} - \begin{array}{l}
\mathop {\arg \min }\limits_{\{{{\boldsymbol{x}}_{2,j}}'\},{\boldsymbol{z}_2}'} \sum\limits_{j = 1}^N {{f_{2,j}}(\hat{\boldsymbol{z}}_1,{{\boldsymbol{x}}_{2,j}}',{\hat{\boldsymbol{x}}_{3,j}})} \\
{\rm{s.t.}}\quad {{\boldsymbol{x}}_{2,j}}' \!=\! {{\boldsymbol{z}}_2}',\forall j, \\ \qquad h_l^{\rm{in}}( \{{\hat{\boldsymbol{x}}_{3,j}}\}, \hat{\boldsymbol{z}}_1, {\boldsymbol{z}_2}', \hat{\boldsymbol{z}}_3) \! \le\! \varepsilon_{\rm{in}},\forall l
\end{array}  ||^2

\\

\le 2 ||  \begin{array}{l}
\mathop {\arg \min }\limits_{\{{{\boldsymbol{x}}_{2,j}}'\},{\boldsymbol{z}_2}'} \sum\limits_{j = 1}^N {{f_{2,j}}(\hat{\boldsymbol{z}}_1,{{\boldsymbol{x}}_{2,j}}',{\hat{\boldsymbol{x}}_{3,j}})} \\
{\rm{s.t.}}\quad {{\boldsymbol{x}}_{2,j}}' \!=\! {{\boldsymbol{z}}_2}',\forall j, \\ \qquad h_l^{\rm{in}}( \{{\hat{\boldsymbol{x}}_{3,j}}\}, \hat{\boldsymbol{z}}_1, {\boldsymbol{z}_2}', \hat{\boldsymbol{z}}_3) \! \le\! \varepsilon_{\rm{in}},\forall l
\end{array}  -   \begin{array}{l}
\mathop {\arg \min }\limits_{\{{{\boldsymbol{x}}_{2,j}}'\},{\boldsymbol{z}_2}'} \sum\limits_{j = 1}^N {{f_{2,j}}(\hat{\boldsymbol{z}}_1,{{\boldsymbol{x}}_{2,j}}',{\hat{\boldsymbol{x}}_{3,j}})} \\
{\rm{s}}{\rm{.t}}{\rm{.}}\; {{\boldsymbol{x}}_{2,j}}' = {\boldsymbol{z}_2}', \forall j =1,\cdots,N  
\\ \quad \; \{{\hat{\boldsymbol{x}}_{3,j}}\}, \hat{\boldsymbol{z}}_3 {\rm{ =  }}\mathop {\arg \min }\limits_{\{{{\boldsymbol{x}}_{3,j}}'\}, {\boldsymbol{z}_3}'} \sum\limits_{j = 1}^N {{f_{3,j}}(\hat{\boldsymbol{z}}_1,{\boldsymbol{z}_2}',{{\boldsymbol{x}}_{3,j}}')} 
\\ \qquad \qquad \quad \; {\rm{s}}{\rm{.t}}{\rm{.}}\; {{\boldsymbol{x}}_{3,j}}' = {\boldsymbol{z}_3}', \forall j =1,\cdots,N  
\end{array} ||^2

\\
\le 2 \beta . 
\end{array}
\end{equation}

By combining the definition of $S_2$ in Eq. (\ref{eq:5_6_72}) with Eq. (\ref{eq:5_6_75}), we can get that $S_3$ is a subset of $S_2$, i.e., $S_3 \in S_2$ when we set $\varepsilon_{\rm{in}} \ge 0$ and $\varepsilon_{\rm{out}} \ge 2 \beta$. Based on Proposition \ref{prop:2}, we have that $S_2$ is a subset of $S_1$, i.e., $S_2 \in S_1$. Consequently, we can get  $S_3 \in S_1$, indicating that the cascaded polynomial approximation problem is the relaxed problem of the original distributed trilevel zeroth order optimization problem. Moreover, this relaxation will be gradually tightened with the addition of zeroth order cuts based on Proposition \ref{prop:1} and \ref{prop:2}.

\section{Discussion about Soft Constraint and $\phi_{\rm{in}}$, $\phi_{\rm{out}}$}
\label{appedix:phi}

\textbf{Soft constraint.} A \textit{soft constraint} refers to a constraint that can be partially violated without rendering the optimization problem meaningless \cite{kautz1996general,regin2011using,wilson2022combining}. It is shown in many bilevel and trilevel learning works that the lower-optimization problem often serves as a soft constraint to the upper-level optimization problem. Examples are provided as follows.

\begin{enumerate}
    \item [*] In bilevel neural architecture search \cite{liu2018darts}, rather than computing the optimal solution for the lower-level optimization problem, the result obtained after a single gradient descent step can be used as an approximation of the optimal solution.

    \item [*] In bilevel meta-learning \cite{ji2021bilevel,finn2017model}, instead of solving the lower-level optimization problem to optimality, the results obtained after multiple gradient descent steps can serve as an approximation.

    \item [*] In bilevel adversarial learning \cite{madry2018towards,zhang2022revisiting}, which is a min-max optimization problem, instead of solving the maximization problem to obtain the optimal solution, the results after several projected gradient descent steps are used as the approximation.

    \item [*] In trilevel learning, AFTO \cite{jiao2024provably} used the results after $K$ communication rounds to replace the optimal solution to the lower-level optimization problem in federated trilevel optimization problems.
\end{enumerate}

It is seen from $\phi_{\rm{in}}( \{{\boldsymbol{x}_{3,j}}\}, \boldsymbol{z}_1, {\boldsymbol{z}_2}', {\boldsymbol{z}_3})=|| \left[ \begin{array}{l}
\{ {\boldsymbol{x}_{3,j}}\} \\
{\boldsymbol{z}_3}
\end{array} \right] - \mathop {\arg \min }\limits_{\{ {\boldsymbol{x}_{3,j}}'\} ,{\boldsymbol{z}_3}'} \sum\nolimits_j {{f_{3,j}}({\boldsymbol{z}_1},{\boldsymbol{z}_2}',{\boldsymbol{x}_{3,j}}')} \,{\rm{s.t.}}\,{\boldsymbol{x}_{3,j}}' \!=\! {\boldsymbol{z}_3}', \forall j||^2$ that a distributed optimization problem needs to be solved if an exact $\phi_{\rm{in}}( \{{\boldsymbol{x}_{3,j}}\}, \boldsymbol{z}_1, \boldsymbol{z}_2, \boldsymbol{z}_3)$ is required. The lower-level optimization problem (i.e., $ \left[ \begin{array}{l}
\{ {\boldsymbol{x}_{3,j}}\} \\
{\boldsymbol{z}_3}
\end{array} \right] = \mathop {\arg \min }\limits_{\{ {\boldsymbol{x}_{3,j}}'\} ,{\boldsymbol{z}_3}'} \sum\nolimits_j {{f_{3,j}}({\boldsymbol{z}_1},{\boldsymbol{z}_2}',{\boldsymbol{x}_{3,j}}')} \,{\rm{s.t.}}\,{\boldsymbol{x}_{3,j}}' \!=\! {\boldsymbol{z}_3}', \forall j$) can be regarded as a soft constraint to the upper-level optimization problem. Inspired by many works in bilevel optimization and trilevel optimization, e.g. \cite{ji2021bilevel,jiao2022distributed,yang2021provably,franceschi2018bilevel,liu2021investigating,mackay2018self,choe2022betty}, that utilize $K$ steps gradient descent steps to approximate the optimal solution to the lower-level optimization problem, function $\phi_{\rm{in}}( \{{\boldsymbol{x}_{3,j}}\}, \boldsymbol{z}_1, {\boldsymbol{z}_2}', \boldsymbol{z}_3)$ in this work can also be approximated based on the solution after $K$ communication rounds following \cite{jiao2024provably}. Specifically, we have the following steps in $(k+1)^{\rm{th}}$ iteration,


Local worker $j$ update the local variables as,
\begin{equation}
    {\boldsymbol{x}_{3,j}^{k+1}} = {\boldsymbol{x}_{3,j}^{k}} - \eta_{x} G_{{\rm{in}},j}({\boldsymbol{z}_1},{\boldsymbol{z}_2},{\boldsymbol{x}_{3,j}^{k}},{\boldsymbol{z}_3^{k}}),
\end{equation}
where $\eta_{x}$ denotes the step-size, and
\begin{equation}
\begin{array}{l}
    G_{{\rm{in}},j}({\boldsymbol{z}_1},{\boldsymbol{z}_2},{\boldsymbol{x}_{3,j}^{k}},{\boldsymbol{z}_3^{k}}) = \frac{{{f_{3,j}}({\boldsymbol{x}_{1,j}},{\boldsymbol{x}_{2,j}},{\boldsymbol{x}_{3,j}^k}+ \mu {\boldsymbol{u}_{k,3}}) -{f_{1,j}}({\boldsymbol{x}_{1,j}},{\boldsymbol{x}_{2,j}},{\boldsymbol{x}_{3,j}^k})}}{\mu } {\boldsymbol{u}_{k,3}} + 2\gamma_j({\boldsymbol{x}_{3,j}^k}- {\boldsymbol{z}_3^k}). 
\end{array}
\end{equation}
where ${\boldsymbol{u}_{k,3}}$ is a standard Gaussian random vector, $\gamma_j>0$ is a constant. Then, workers transmit the updated local variables, i.e., ${\boldsymbol{x}_{3,j}^{k+1}}$, to the master.

After receiving the updated variables, the master updates the consensus variables as follows.
\begin{equation}
    {\boldsymbol{z}_3^{k+1}} = {\boldsymbol{z}_3^{k}} - \eta_{z} \sum\nolimits_{j=1}^N \gamma_j({\boldsymbol{z}_3^{k}} - {\boldsymbol{x}_{3,j}^{k+1}}),
\end{equation}
where $\eta_{z}$ represents the step-size. Subsequently, the master broadcasts the updated variables ${\boldsymbol{z}_3^{k+1}}$ to workers. Thus, the approximated $\phi_{\rm{in}}( \{{\boldsymbol{x}_{3,j}}\}, \boldsymbol{z}_1, \boldsymbol{z}_2, \boldsymbol{z}_3)$ can be expressed as,
\begin{equation}
   \phi_{\rm{in}}( \{{\boldsymbol{x}_{3,j}}\}, \boldsymbol{z}_1, \boldsymbol{z}_2, \boldsymbol{z}_3) = \left[ \begin{array}{l}
       \{ {\boldsymbol{x}_{3,j}} - {\boldsymbol{x}_{3,j}^0} + \eta_{x} \sum\nolimits_{k=0}^{K-1} G_{{\rm{in}},j}({\boldsymbol{z}_1},{\boldsymbol{z}_2},{\boldsymbol{x}_{3,j}^{k}},{\boldsymbol{z}_3^{k}}) \} \\
{\boldsymbol{z}_3} - {\boldsymbol{z}_3^0} + \eta_{z} \sum_{k=0}^{K-1} \sum\nolimits_{j=1}^N \gamma_j ({\boldsymbol{z}_3^{k}} - {\boldsymbol{x}_{3,j}^{k+1}})
   \end{array} \right].
\end{equation}

Likewise, constraint $\phi_{\rm{out}}( \{{\boldsymbol{x}_{2,j}}\}, \{{\boldsymbol{x}_{3,j}}\}, \boldsymbol{z}_1, \boldsymbol{z}_2, \boldsymbol{z}_3) = 0$ also serves as a soft constraint to the upper-level optimization problem. According to the definition of $\phi_{\rm{out}}( \{{\boldsymbol{x}_{2,j}}\}, \{{\boldsymbol{x}_{3,j}}\}, \boldsymbol{z}_1, \boldsymbol{z}_2, \boldsymbol{z}_3)$, that is,
\begin{equation}
\begin{array}{l}
\phi_{\rm{out}}( \{{\boldsymbol{x}_{2,j}}\}, \{{\boldsymbol{x}_{3,j}}\}, \boldsymbol{z}_1, \boldsymbol{z}_2, \boldsymbol{z}_3)
\\=|| \left[ \begin{array}{l}
\{ {\boldsymbol{x}_{2,j}}\} \\
{\boldsymbol{z}_2} 
\end{array} \right] - \begin{array}{l}
\mathop {\arg \min }\limits_{\{{\boldsymbol{x}_{2,j}}\},\boldsymbol{z}_2} \sum\limits_{j = 1}^N {{f_{2,j}}(\boldsymbol{z}_1,{\boldsymbol{x}_{2,j}},{\boldsymbol{x}_{3,j}})} \\
{\rm{s.t.}}\,{\boldsymbol{x}_{2,j}} \!=\! {\boldsymbol{z}_2},\forall j,h_l^{\rm{in}}( \{{\boldsymbol{x}_{3,j}}\}, \boldsymbol{z}_1, \boldsymbol{z}_2, \boldsymbol{z}_3) \! \le\! \varepsilon_{\rm{in}},\forall l
\end{array}  ||^2,
\end{array}
\end{equation}
the results after $K$ communication rounds can also be utilized to compute the estimate of $\phi_{\rm{out}}( \{{\boldsymbol{x}_{2,j}}\}, \{{\boldsymbol{x}_{3,j}}\}, \boldsymbol{z}_1, \boldsymbol{z}_2, \boldsymbol{z}_3)$ following previous works \cite{liu2018darts,jiao2024provably}.  In $(k+1)^{\rm{th}}$ iteration, we have that,


Local worker $j$ updates the local variables as follows,
\begin{equation}
        {\boldsymbol{x}_{2,j}^{k+1}} = {\boldsymbol{x}_{2,j}^{k}} - \eta_{x} G_{{\boldsymbol{x}_{2,j}}}(\boldsymbol{z}_1,{\boldsymbol{x}_{2,j}^{k}},{\boldsymbol{x}_{3,j}}, {\boldsymbol{z}_2^{k}}, {\boldsymbol{z}_3}), 
\end{equation}
where we have,
\begin{equation}
\begin{array}{l}
    G_{{\boldsymbol{x}_{2,j}}}(\boldsymbol{z}_1,{\boldsymbol{x}_{2,j}^{k}},{\boldsymbol{x}_{3,j}}, {\boldsymbol{z}_2^{k}}, {\boldsymbol{z}_3}) \\
    = \frac{{{f_{2,j}}({\boldsymbol{z}_1},{\boldsymbol{x}_{2,j}^k}+ \mu {\boldsymbol{u}_{k,2}},{\boldsymbol{x}_{3,j}}) -{f_{2,j}}({\boldsymbol{z}_1},{\boldsymbol{x}_{2,j}^k},{\boldsymbol{x}_{3,j}})}}{\mu } {\boldsymbol{u}_{k,2}} + 2 \varphi_j({\boldsymbol{x}_{2,j}^k} - {\boldsymbol{z}_2^k}),
\end{array}
\end{equation}
where ${\boldsymbol{u}_{k,2}}$  is the standard Gaussian random vector, $\varphi_j>0$ is a constant. Then, worker $j$ transmits the updated ${\boldsymbol{x}_{2,j}^{k+1}}$ to the master.

After receiving the updated parameters from workers, the master updates the consensus variables as,
\begin{equation}
    {\boldsymbol{z}_2^{k+1}} = {\boldsymbol{z}_2^{k}} - \eta_{z}\left( 2 \varphi_j({\boldsymbol{z}_2^k} - {\boldsymbol{x}_{2,j}^{k+1}}) + \nabla_{\boldsymbol{z}_2} p_l\sum\nolimits_l [\max\{ h_l^{\rm{in}}( \{{\boldsymbol{x}_{3,j}}\}, \boldsymbol{z}_1, \boldsymbol{z}_2^k, \boldsymbol{z}_3) - \varepsilon_{\rm{in}},0\}]^2 \right).
\end{equation}

Next, the master broadcasts the updated variables ${\boldsymbol{z}_2^{k+1}}$ to workers. Consequently, the approximated $\phi_{\rm{out}}( \{{\boldsymbol{x}_{2,j}}\}, \{{\boldsymbol{x}_{3,j}}\}, \boldsymbol{z}_1, \boldsymbol{z}_2, \boldsymbol{z}_3)$ can be written as,
\begin{equation}
\begin{array}{l}
    \phi_{\rm{out}}( \{{\boldsymbol{x}_{2,j}}\}, \{{\boldsymbol{x}_{3,j}}\}, \boldsymbol{z}_1, \boldsymbol{z}_2, \boldsymbol{z}_3)
    
    \\ \!=\! \left[ \begin{array}{l}
\{ {\boldsymbol{x}_{2,j}} - {\boldsymbol{x}_{2,j}^0} + \sum_{k=0}^{K-1}\eta_{x}G_{{\boldsymbol{x}_{2,j}}}(\boldsymbol{z}_1,{\boldsymbol{x}_{2,j}^{k}},{\boldsymbol{x}_{3,j}}, {\boldsymbol{z}_2^{k}}, {\boldsymbol{z}_3})\} \\

{\boldsymbol{z}_2} \!-\! {\boldsymbol{z}_2^0} \!+\! \sum_{k=0}^{K-1} \!\eta_{z}\!\left( 2 \varphi_j({\boldsymbol{z}_2^k} \!- \!{\boldsymbol{x}_{2,j}^{k+1}}) \!+ \!\nabla_{\boldsymbol{z}_2} p_l\sum\nolimits_l [\max\{ h_l^{\rm{in}}(\{{\boldsymbol{x}_{3,j}}\}, \boldsymbol{z}_1, \boldsymbol{z}_2^k, \boldsymbol{z}_3) \!-\! \varepsilon_{\rm{in}},0\}]^2 \right) 
\end{array} \right].
\end{array}
\end{equation}

\section{Experimental Setting}
\label{appendix:experiment}

In this section, we provide the details of the experimental setting. In the experiment, all the models are implemented using PyTorch, and the experiments are conducted on a server equipped with two NVIDIA RTX 4090 GPUs. 

In the experiment, we compare the proposed method with the state-of-the-art distributed zeroth order learning method FedZOO \cite{fang2022communication} and state-of-the-art distributed bilevel zeroth order learning method FedRZO$_{\rm{bl}}$ \cite{qiu2023zeroth}, which are introduced as follows. FedZOO \cite{fang2022communication} is a derivative-free federated zeroth-order optimization method, which can be applied to solve the single-level optimization problems in a distributed manner. In FedZOO, clients perform several local updates based on gradient estimators in each communication round. After receiving local updates, the servers will perform the aggregation and update the global parameters. FedRZO$_{\rm{bl}}$ \cite{qiu2023zeroth} is designed for zeroth order bilevel optimization problems. In each communication round, FedRZO$_{\rm{bl}}$ involves the following steps: clients first compute the estimated optimal solution to the lower-level optimization problem and the inexact implicit zeroth-order gradient. They then update the local parameters and transmit them to the server. Upon receiving the updates, the server aggregates them to obtain the global parameters.

\subsection{Black-box Trilevel Learning}
In this section, the details of the experimental setting in black-box trilevel learning are provided. Prompt learning is a key technique for enabling LLMs to efficiently and effectively adapt to various downstream tasks \cite{ma2024fairness,wang2024grammar}. Inspired by the black-box prompt learning \cite{diao2022black} and the backdoor attack on prompt-based LLMs \cite{yao2024poisonprompt}, the backdoor attack on black-box LLMs is considered with hyperparameter optimization in the experiment. In the experiment, Qwen 1.8B-Chat \cite{bai2023qwen} is utilized as the black-box LLM. The General Language Understanding Evaluation (GLUE) benchmark \cite{wang2018glue} is used to evaluate the proposed DTZO. Specifically, the experiments are carried out on: 1) SST-2 for sentiment analysis; 2) COLA for linguistic acceptability; and 3) MRPC for semantic equivalence of sentences. In the black-box trilevel learning problem, we compare the proposed DTZO with the state-of-the-art distributed bilevel zeroth order learning method FedRZO$_{\rm{bl}}$ \cite{qiu2023zeroth}, which is used to address the following distributed bilevel zeroth order learning problem,
\begin{equation}
    \begin{array}{l}
\mathop {\min }\sum\nolimits_{j = 1}^N\frac{1}{{|{D^{\rm{tr}}_j}|}}\sum\limits_{(\boldsymbol{s}_i,{y_i}) \sim {D^{\rm{tr}}_j}} {L({\cal G},[{\boldsymbol{k}_{\rm{tri}}} ,\boldsymbol{p},\boldsymbol{s}_i],{y_i})}  \\
{\rm{s.t.}}\; \boldsymbol{p} = \mathop {\arg \min }\limits_{\boldsymbol{p}'}\sum\nolimits_{j = 1}^N \frac{1}{{|{D^{\rm{tr}}_j}|}}\sum\limits_{(\boldsymbol{s}_i,{y_i}) \sim {D^{\rm{tr}}_j}} {L({\cal G},[{\boldsymbol{k}_{\rm{tri}}},\boldsymbol{p}' ,\boldsymbol{s}_i],{y_i})}  \\

{\mathop{\rm var}} .\qquad \qquad \qquad {\boldsymbol{k}_{\rm{tri}}},\boldsymbol{p},
\end{array}
\end{equation}
where ${\cal G}$ denotes the black-box LLM.  ${\boldsymbol{k}_{\rm{tri}}}$ and  $\boldsymbol{p}$ respectively denote the backdoor trigger and prompt. ${D^{\rm{tr}}_j}$ represents the training dataset in $j^{\rm{th}}$ worker, $|{D^{\rm{tr}}_j}|$ represents the number of data in training dataset, and $N$ denotes the number of workers. $\boldsymbol{s}_i,{y_i}$ denote the $i^{\rm{th}}$ input sentence and label.

\subsection{Robust Hyperparameter Optimization}
Robust hyperparameter optimization is a widely used trilevel learning application \cite{jiao2024provably, sato2021gradient}, aiming to optimize hyperparameters \cite{ji2021bilevel,franceschi2018bilevel,jiao2022timeautoad,yang2021provably} and train a machine learning model that is robust against adversarial attacks \cite{han2024fedal}. In this work, we consider the robust hyperparameter optimization, which can be viewed as a trilevel zeroth order learning problem. In this task, compared to single-level optimization, bilevel optimization considers the hyperparameter optimization, which can enhance the generalization ability of the machine learning model. Compared to bilevel
 optimization, trilevel optimization incorporates min-max robust training, which can improve the adversarial robustness of ML model. In the experiments, the digits recognition tasks in \cite{qian2019robust,wang2021discriminative} with four benchmark datasets, i.e., MNIST \cite{lecun1998gradient}, USPS, Fashion MNIST \cite{xiao2017fashion}, KMNIST \cite{clanuwat2018deep}, and QMNIST \cite{yadav2019cold}, are utilized to assess the performance of the proposed DTZO. To evaluate the robustness of each method, the PGD-7 attack \cite{madry2018towards} with $\varepsilon=0.05$ is utilized.
For the state-of-the-art distributed zeroth order learning method FedZOO \cite{fang2022communication}, it is used to address the following distributed zeroth order learning problem in this task,
\begin{equation}
    \begin{array}{l}
 { \min } \sum\nolimits_{j = 1}^N {{f_j}(X_j^{\rm{tr}},y_j^{\rm{tr}},\boldsymbol{w})} \\
{\mathop{\rm var}} .\qquad \qquad \boldsymbol{w},
\end{array}
\end{equation}
where $N$ represents the number of workers in a distributed system, $\boldsymbol{w}$ denotes the model parameter. $X_j^{\rm{tr}}$ and $y_j^{\rm{tr}}$ represent the training data and labels, respectively. 
For the state-of-the-art distributed bilevel zeroth order learning method FedRZO$_{\rm{bl}}$ \cite{qiu2023zeroth}, the following distributed bilevel zeroth order learning problem is considered in this task,
\begin{equation}
    \begin{array}{l}
\min \sum\nolimits_{j = 1}^N {{f_j}(X_j^{{\mathop{\rm var}} },y_j^{{\mathop{\rm var}} },\boldsymbol{w})} \vspace{1mm}\\
{\rm{s}}{\rm{.t}}{\rm{.}}\; \boldsymbol{w} = \mathop {\arg \min }\limits_{\boldsymbol{w}'} \sum\nolimits_{j = 1}^N {{f_j}(X_j^{\rm{tr}},y_j^{\rm{tr}},\boldsymbol{w}')}  + \varphi ||\boldsymbol{w}'|{|^2}\\
{\mathop{\rm var}} .\qquad \qquad \varphi ,\boldsymbol{w},
\end{array}
\end{equation}
where $\varphi$ and $\boldsymbol{w}$ denote the regularization coefficient and model parameter, respectively. $X_j^{\rm{tr}}$ and $y_j^{\rm{tr}}$ represent the training data and labels, while $X_j^{\rm{var}}$ and $y_j^{\rm{var}}$ represent the validation data and labels, respectively.

\section{Discussion about Assumption \ref{assum:1} and \ref{assum:2}}
\label{appendix:assum}

The assumption that the domains of optimization variables are bounded is mild and widely used in the theoretical analyses in machine learning, e.g.,  Assumption 3 in \cite{deng2020distributionally}, Assumption 2.3 in \cite{sra2016adadelay}, Assumption A2 in \cite{li2021distributed}, Assumption 2.1 in \cite{cao2024projection} and so on.  

Let $(\{{\boldsymbol{x}_{1,j}^*}\}, \{{\boldsymbol{x}_{2,j}^*}\}, \{{\boldsymbol{x}_{3,j}^*}\}, \boldsymbol{z}_1^*, \boldsymbol{z}_2^*, \boldsymbol{z}_3^*)$ represent the optimal solution of minimizing $F_{\mu}(\{{\boldsymbol{x}_{1,j}}\},\! \{{\boldsymbol{x}_{2,j}}\},\! \{{\boldsymbol{x}_{3,j}}\},\! \boldsymbol{z}_1,\! \boldsymbol{z}_2,\! \boldsymbol{z}_3)$, $(\{{\boldsymbol{x}_{1,j}^+}\}, \{{\boldsymbol{x}_{2,j}^+}\}, \{{\boldsymbol{x}_{3,j}^+}\})$ denote the optimal solution of minimizing $\sum\limits_{j = 1}^N \! {{f_{1,j}}({\boldsymbol{x}_{1,j}},{\boldsymbol{x}_{2,j}},{\boldsymbol{x}_{3,j}})}$, and $\boldsymbol{x}_{1,j}^{-}, \boldsymbol{x}_{2,j}^{-}, \boldsymbol{x}_{3,j}^{-}$ denote the optimal solution of minimizing ${f_{1,j}}({\boldsymbol{x}_{1,j}},{\boldsymbol{x}_{2,j}},{\boldsymbol{x}_{3,j}})$. Thus, we have that,
\begin{equation}
    \sum\limits_{j = 1}^N \! {{f_{1,j}}({\boldsymbol{x}_{1,j}^-},{\boldsymbol{x}_{2,j}^-},{\boldsymbol{x}_{3,j}^-})} \le \sum\limits_{j = 1}^N \! {{f_{1,j}}({\boldsymbol{x}_{1,j}^+},{\boldsymbol{x}_{2,j}^+},{\boldsymbol{x}_{3,j}^+})} \le \sum\limits_{j = 1}^N \! {{f_{1,j}}({\boldsymbol{x}_{1,j}^*},{\boldsymbol{x}_{2,j}^*},{\boldsymbol{x}_{3,j}^*})}.
\end{equation}
Combining the definition of $F(\{{\boldsymbol{x}_{1,j}}\},\! \{{\boldsymbol{x}_{2,j}}\},\! \{{\boldsymbol{x}_{3,j}}\},\! \boldsymbol{z}_1,\! \boldsymbol{z}_2,\! \boldsymbol{z}_3)$ in Eq. (\ref{eq:5}) with the fact that $\phi_j ||\boldsymbol{x}_{1,j}^* \!-\! \boldsymbol{z}_1^* ||^2 \ge 0$, $\lambda_l [\max\{h_l^{\rm{out}}( \{{\boldsymbol{x}_{2,j}^*}\}, \{{\boldsymbol{x}_{3,j}^*}\}, \boldsymbol{z}_1^*, \boldsymbol{z}_2^*, \boldsymbol{z}_3^*) - \varepsilon_{\rm{out}} \}]^2 \ge 0$, we can obtain that,
\begin{equation}
\label{eq:5_10_87}
\begin{array}{l}
\sum\limits_{j = 1}^N \! {{f_{1,j}}({\boldsymbol{x}_{1,j}^-},{\boldsymbol{x}_{2,j}^-},{\boldsymbol{x}_{3,j}^-})} - \frac{\mu^2}{2}L(N+1)\sum_i d_i 

\\
    \le \sum\limits_{j = 1}^N \! {{f_{1,j}}({\boldsymbol{x}_{1,j}^+},{\boldsymbol{x}_{2,j}^+},{\boldsymbol{x}_{3,j}^+})} - \frac{\mu^2}{2}L(N+1)\sum_i d_i 
    
    \\
    
    \le \sum\limits_{j = 1}^N \! {{f_{1,j}}({\boldsymbol{x}_{1,j}^*},{\boldsymbol{x}_{2,j}^*},{\boldsymbol{x}_{3,j}^*})}  - \frac{\mu^2}{2}L(N+1)\sum_i d_i

    \\
    \le F(\{{\boldsymbol{x}_{1,j}^*}\},\! \{{\boldsymbol{x}_{2,j}^*}\},\! \{{\boldsymbol{x}_{3,j}^*}\},\! \boldsymbol{z}_1^*,\! \boldsymbol{z}_2^*,\! \boldsymbol{z}_3^*) - \frac{\mu^2}{2}L(N+1)\sum_i d_i

   \vspace{2mm}  \\ \le F_{\mu}(\{{\boldsymbol{x}_{1,j}^*}\},\! \{{\boldsymbol{x}_{2,j}^*}\},\! \{{\boldsymbol{x}_{3,j}^*}\},\! \boldsymbol{z}_1^*,\! \boldsymbol{z}_2^*,\! \boldsymbol{z}_3^*)

  \vspace{2mm}  \\  = F_{\mu}^*
    .
\end{array}
\end{equation}
By combining Eq. (\ref{eq:5_10_87}) with the fact that $\frac{\mu^2}{2}L(N+1)\sum_i d_i$ is a constant, we can obtain that the Assumption \ref{assum:1} (i.e., $F_{\mu}^*$ is lower-bounded) is mild since the assumption that ${f_{1,j}}({\boldsymbol{x}_{1,j}^-},{\boldsymbol{x}_{2,j}^-},{\boldsymbol{x}_{3,j}^-})$ is lower-bounded is widely-used and mild \cite{liu2021conflict,liu2018signsgd,liu2022bome,fang2022communication,li2021distributed,liang2024novel,tang2020distributed,shaban2019truncated}.

According to the definition of $F(\{{\boldsymbol{x}_{1,j}}\},\! \{{\boldsymbol{x}_{2,j}}\},\! \{{\boldsymbol{x}_{3,j}}\},\! \boldsymbol{z}_1,\! \boldsymbol{z}_2,\! \boldsymbol{z}_3)$, i.e.,
\begin{equation}
\begin{array}{l}
\! F(\{{\boldsymbol{x}_{1,j}}\},\! \{{\boldsymbol{x}_{2,j}}\},\! \{{\boldsymbol{x}_{3,j}}\},\! \boldsymbol{z}_1,\! \boldsymbol{z}_2,\! \boldsymbol{z}_3) \!= \!\sum\limits_{j = 1}^N \! {{f_{1,j}}({\boldsymbol{x}_{1,j}},{\boldsymbol{x}_{2,j}},{\boldsymbol{x}_{3,j}}) + \phi_j ||\boldsymbol{x}_{1,j} \!-\! \boldsymbol{z}_1 ||^2} \vspace{1mm} \\

\qquad \qquad \qquad \qquad \qquad \qquad \qquad \; + \! \sum_l \! \lambda_l [\max\{h_l^{\rm{out}}( \{{\boldsymbol{x}_{2,j}}\},\! \{{\boldsymbol{x}_{3,j}}\},\! \boldsymbol{z}_1,\! \boldsymbol{z}_2,\! \boldsymbol{z}_3) \!-\! \varepsilon_{\rm{out}} \}]^2,
\end{array}
\end{equation}
we have that 1) term $\phi_j ||\boldsymbol{x}_{1,j} \!-\! \boldsymbol{z}_1 ||^2$ satisfies the $L$-smoothness because the domains of variables $\boldsymbol{x}_{1,j}$ and $\boldsymbol{z}_1$ are bounded; 2) term $\sum_l \! \lambda_l [\max\{h_l^{\rm{out}}( \{{\boldsymbol{x}_{2,j}}\},\! \{{\boldsymbol{x}_{3,j}}\},\! \boldsymbol{z}_1,\! \boldsymbol{z}_2,\! \boldsymbol{z}_3) \!-\! \varepsilon_{\rm{out}} \}]^2$ satisfies the $L$-smoothness because the domains of variables are bounded and there are at most $\lfloor \frac{T_1}{\mathcal{T}} \rfloor$ zeroth order cuts. Moreover, the assumption that ${f_{1,j}}({\boldsymbol{x}_{1,j}},{\boldsymbol{x}_{2,j}},{\boldsymbol{x}_{3,j}})$ satisfies the $L$-smoothness is mild and widely-used \cite{ji2021bilevel,gao2024decentralized,gao2022convergence,chen2023optimal,li2024communication,wu2024federated,huang2024nonconvex,jing2024asynchronous,chen2024optimal,xiao2023alternating,hong2023two}. Consequently, we can obtain that $F(\{{\boldsymbol{x}_{1,j}}\},\! \{{\boldsymbol{x}_{2,j}}\},\! \{{\boldsymbol{x}_{3,j}}\},\! \boldsymbol{z}_1,\! \boldsymbol{z}_2,\! \boldsymbol{z}_3)$ satisfies the $L$-smoothness, i.e., Assumption \ref{assum:2} is mild.

\section{Exterior Penalty Method}
\label{appendix:penalty}

Exterior penalty methods are widely-used when dealing with constrained optimization problems \cite{boyd2004convex,bertsekas2015convex}. In this work, the exterior penalty method is utilized based on the following key reasons. 1) The lower-level optimization problem often serves as a soft constraint to the upper-level optimization problem, as discussed in Sec. \ref{Cascaded} and Appendix \ref{appedix:phi}, which can be partially violated without rendering the optimization problem meaningless. We can flexibly control the importance in the upper-level and lower-level problems through adjusting the penalty parameters. For example, if the importance of the lower-level optimization problem is required to be high within the nested optimization problem, we can raise the penalty parameters. 2) The complexity of using the exterior penalty method is relatively lower. For example, if we utilize the gradient projection method, which is also widely-used in constrained optimization \cite{jiao2022asynchronous,xu2020unified}, we need to solve additional one constrained optimization problem with non-convex feasible regions at each iteration when performing projection, i.e.,
\begin{equation}
\label{eq:5_19_98}
    \begin{array}{l}
\min \sum\limits_{i = 1}^3 \sum\limits_{j = 1}^N ||\boldsymbol{x}_{i,j}^{t+1} - \boldsymbol{x}_{i,j}||^2 + \sum\limits_{i = 1}^3|| \boldsymbol{z}_{i}^{t+1} - \boldsymbol{z}_{i}||^2  \\

{\rm{s}}{\rm{.t}}{\rm{.}}\; {\boldsymbol{x}_{1,j}} = \boldsymbol{z}_1, \forall j =1,\cdots,N 

\\ \sum\limits_{i=2}^{3} \! \sum\limits_{j=1}^{N}\!{{{\boldsymbol{a}_{i,j,l}^{\rm{out}}}^{\top}\boldsymbol{x}_{i,j}^2} \!+\! {{\boldsymbol{b}_{i,j,l}^{\rm{out}}}^{\top}\boldsymbol{x}_{i,j}}}  \!+\!  \sum\limits_{i=1}^{3} \!{{\boldsymbol{c}_{i,l}^{\rm{out}}}^{\top}\boldsymbol{z}_{i}^2} \!+\! {{\boldsymbol{d}_{i,l}^{\rm{out}}}^{\top}\boldsymbol{z}_{i}} \!+\! e_{l}^{\rm{out}} \!\le\! \varepsilon_{\rm{out}}, \forall l 

\\ {\mathop{\rm var}} .\qquad  \{{\boldsymbol{x}_{1,j}}\},\{{\boldsymbol{x}_{2,j}}\},\{{\boldsymbol{x}_{3,j}}\}, \boldsymbol{z}_1, \boldsymbol{z}_2, \boldsymbol{z}_3,

\end{array}
\end{equation}
where $(\{\boldsymbol{x}_{i,j}^{t+1}\},\{\boldsymbol{z}_{i}^{t+1}\} )$ denotes the points in $(t+1)^{\rm{th}}$ iteration after performing zeroth order gradient descent. Thus, it is seen from Eq. (\ref{eq:5_19_98}) that the complexity of utilizing gradient projection descent method is higher than using the penalty method since it requires addressing the constrained non-convex optimization problem in Eq. (\ref{eq:5_19_98}) at each iteration. Likewise, utilizing the Frank-Wolfe based methods \cite{shen2019complexities,garber2015faster,zhang2020one,xian2021communication,wang2016parallel,balashov2020gradient} may also lead to relatively more computational complexity since it also needs to solve one additional constrained non-convex optimization problem, i.e.,
\begin{equation}
\label{eq:9_30_102}
    \begin{array}{l}
\min \sum\limits_{i = 1}^3 \sum\limits_{j = 1}^N \nabla_{{\boldsymbol{x}_{i,j}}} {f_{1,j}}({\boldsymbol{x}_{1,j}^{t+1}},{\boldsymbol{x}_{2,j}^{t+1}},{\boldsymbol{x}_{3,j}^{t+1}})^{\top}({\boldsymbol{x}_{i,j}} - {\boldsymbol{x}_{i,j}^{t+1}})   \\

{\rm{s}}{\rm{.t}}{\rm{.}}\; {\boldsymbol{x}_{1,j}} = \boldsymbol{z}_1, \forall j =1,\cdots,N 

\\ \sum\limits_{i=2}^{3} \! \sum\limits_{j=1}^{N}\!{{{\boldsymbol{a}_{i,j,l}^{\rm{out}}}^{\top}\boldsymbol{x}_{i,j}^2} \!+\! {{\boldsymbol{b}_{i,j,l}^{\rm{out}}}^{\top}\boldsymbol{x}_{i,j}}}  \!+\!  \sum\limits_{i=1}^{3} \!{{\boldsymbol{c}_{i,l}^{\rm{out}}}^{\top}\boldsymbol{z}_{i}^2} \!+\! {{\boldsymbol{d}_{i,l}^{\rm{out}}}^{\top}\boldsymbol{z}_{i}} \!+\! e_{l}^{\rm{out}} \!\le\! \varepsilon_{\rm{out}}, \forall l 

\\ {\mathop{\rm var}} .\qquad  \{{\boldsymbol{x}_{1,j}}\},\{{\boldsymbol{x}_{2,j}}\},\{{\boldsymbol{x}_{3,j}}\}, \boldsymbol{z}_1, \boldsymbol{z}_2, \boldsymbol{z}_3.

\end{array}
\end{equation}

Thus, as indicated by Eq. (\ref{eq:9_30_102}), the complexity of using the Frank-Wolfe based method is higher than that of the exterior penalty method, as it requires solving an additional constrained non-convex optimization problem in Eq. (\ref{eq:9_30_102}) at each iteration. Based on the aforementioned reasons, we chose to use the exterior penalty method in this work.

In addition, we demonstrate the close relationship between the original constrained optimization problem (P1) in Eq. (\ref{eq:4}) and the unconstrained optimization problem (P2) in Eq. (\ref{eq:5}) in this work. That is, 1) the optimal solution to P2 is also a feasible solution to the relaxed original problem P1; 2) the gap between the optimal objective value by utilizing the exterior penalty method ( i.e., $\sum_{j = 1}^N \! {f_{1,j}}({\boldsymbol{x}_{1,j}^*},{\boldsymbol{x}_{2,j}^*},{\boldsymbol{x}_{3,j}^*})$ in P2) and the optimal objective value in original problem P1 (i.e., $\sum_{j = 1}^N \!{f_{1,j}}(\{{\overline{\boldsymbol{x}}_{1,j}}\},\! \{{\overline{\boldsymbol{x}}_{2,j}}\},\! \{{\overline{\boldsymbol{x}}_{3,j}}\})$) will continuously decrease with penalty parameters increased. To enhance the readability of this discussion, the constrained optimization problem and unconstrained optimization problem are presented as follows.

Constrained cascaded polynomial approximation problem (P1):
\begin{equation}
    \begin{array}{l}
\min \sum\limits_{j = 1}^N {{f_{1,j}}({\boldsymbol{x}_{1,j}},{\boldsymbol{x}_{2,j}},{\boldsymbol{x}_{3,j}})} \\

{\rm{s}}{\rm{.t}}{\rm{.}}\; {\boldsymbol{x}_{1,j}} = \boldsymbol{z}_1, \forall j =1,\cdots,N 

\\ \sum\limits_{i=2}^{3} \! \sum\limits_{j=1}^{N}\!{{{\boldsymbol{a}_{i,j,l}^{\rm{out}}}^{\top}\boldsymbol{x}_{i,j}^2} \!+\! {{\boldsymbol{b}_{i,j,l}^{\rm{out}}}^{\top}\boldsymbol{x}_{i,j}}}  \!+\!  \sum\limits_{i=1}^{3} \!{{\boldsymbol{c}_{i,l}^{\rm{out}}}^{\top}\boldsymbol{z}_{i}^2} \!+\! {{\boldsymbol{d}_{i,l}^{\rm{out}}}^{\top}\boldsymbol{z}_{i}} \!+\! e_{l}^{\rm{out}} \!\le\! \varepsilon_{\rm{out}}, \forall l 

\\ {\mathop{\rm var}} .\qquad  \{{\boldsymbol{x}_{1,j}}\},\{{\boldsymbol{x}_{2,j}}\},\{{\boldsymbol{x}_{3,j}}\}, \boldsymbol{z}_1, \boldsymbol{z}_2, \boldsymbol{z}_3.

\end{array}
\end{equation}

Unconstrained optimization problem based on exterior penalty method (P2):
\begin{equation}
\label{eq:5_14_94}
\begin{array}{l}
\! \min F(\{{\boldsymbol{x}_{1,j}}\},\! \{{\boldsymbol{x}_{2,j}}\},\! \{{\boldsymbol{x}_{3,j}}\},\! \boldsymbol{z}_1,\! \boldsymbol{z}_2,\! \boldsymbol{z}_3) \!:= \!\sum\limits_{j = 1}^N \! {{f_{1,j}}({\boldsymbol{x}_{1,j}},{\boldsymbol{x}_{2,j}},{\boldsymbol{x}_{3,j}}) + \phi_j ||\boldsymbol{x}_{1,j} \!-\! \boldsymbol{z}_1 ||^2} \vspace{1mm} \\

\qquad \qquad \qquad \qquad \qquad \qquad \qquad \qquad \, + \! \sum_l \! \lambda_l [\max\{h_l^{\rm{out}}( \{{\boldsymbol{x}_{2,j}}\},\! \{{\boldsymbol{x}_{3,j}}\},\! \boldsymbol{z}_1, \boldsymbol{z}_2, \boldsymbol{z}_3) \!-\! \varepsilon_{\rm{out}},0 \}]^2,

\\ {\mathop{\rm var}} . \qquad \qquad \qquad  \{{\boldsymbol{x}_{1,j}}\},\{{\boldsymbol{x}_{2,j}}\},\{{\boldsymbol{x}_{3,j}}\}, \boldsymbol{z}_1, \boldsymbol{z}_2, \boldsymbol{z}_3,
\end{array}
\end{equation}

where $h_l^{\rm{out}}( \{{\boldsymbol{x}_{2,j}}\}, \{{\boldsymbol{x}_{3,j}}\}, \boldsymbol{z}_1, \boldsymbol{z}_2, \boldsymbol{z}_3) = \sum\limits_{i=2}^{3} \! \sum\limits_{j=1}^{N}\!{{{\boldsymbol{a}_{i,j,l}^{\rm{out}}}^{\top}\boldsymbol{x}_{i,j}^2} \!+\! {{\boldsymbol{b}_{i,j,l}^{\rm{out}}}^{\top}\boldsymbol{x}_{i,j}}}  \!+\!  \sum\limits_{i=1}^{3} \!{{\boldsymbol{c}_{i,l}^{\rm{out}}}^{\top}\boldsymbol{z}_{i}^2} \!+\! {{\boldsymbol{d}_{i,l}^{\rm{out}}}^{\top}\boldsymbol{z}_{i}} \!+\! e_{l}^{\rm{out}}$. We first show that the optimal solution to P2 is also a feasible solution to the relaxed original problem P1, and this relaxation will be gradually tightened with penalty parameters increased. Let $(\{{\boldsymbol{x}_{1,j}^*}\},\! \{{\boldsymbol{x}_{2,j}^*}\},\! \{{\boldsymbol{x}_{3,j}^*}\},\! \boldsymbol{z}_1^*,\! \boldsymbol{z}_2^*,\! \boldsymbol{z}_3^*)$ denote the optimal solution to P2 in Eq. (\ref{eq:5_14_94}). For any point $(\{{\boldsymbol{x}_{1,j}^-}\}, \{{\boldsymbol{x}_{2,j}^-}\}, \{{\boldsymbol{x}_{3,j}^-}\}, \boldsymbol{z}_1^-, \boldsymbol{z}_2^-, \boldsymbol{z}_3^-)$ satisfies $h_l^{\rm{out}}(\{{\boldsymbol{x}_{1,j}^-}\}, \{{\boldsymbol{x}_{2,j}^-}\}, \{{\boldsymbol{x}_{3,j}^-}\}, \boldsymbol{z}_1^-, \boldsymbol{z}_2^-, \boldsymbol{z}_3^-) \le \varepsilon_{\rm{out}}, \forall l$ and $\boldsymbol{x}_{1,j} - \boldsymbol{z}_1 = 0, \forall j$, since it is also the feasible solution to P2, we have that,
\begin{equation}
\begin{array}{l}
   \sum\limits_{j = 1}^N \! {{f_{1,j}}({\boldsymbol{x}_{1,j}^*},{\boldsymbol{x}_{2,j}^*},{\boldsymbol{x}_{3,j}^*})\! +\! \phi_j ||\boldsymbol{x}_{1,j}^* \!-\! \boldsymbol{z}_1^* ||^2} 
  \vspace{1mm} \\ \!+ \! \sum_l \! \lambda_l [\max\{h_l^{\rm{out}}( \{{\boldsymbol{x}_{2,j}^*}\}, \{{\boldsymbol{x}_{3,j}^*}\}, \boldsymbol{z}_1^*, \boldsymbol{z}_2^*, \boldsymbol{z}_3^*) \!-\! \varepsilon_{\rm{out}},0 \}]^2

  \vspace{1mm} \\
   
  \le \sum\limits_{j = 1}^N \! {{f_{1,j}}({\boldsymbol{x}_{1,j}^-},{\boldsymbol{x}_{2,j}^-},{\boldsymbol{x}_{3,j}^-})\! +\! \phi_j ||\boldsymbol{x}_{1,j}^- \!-\! \boldsymbol{z}_1^- ||^2} 
  \vspace{1mm} \\ \!+ \! \sum_l \! \lambda_l [\max\{h_l^{\rm{out}}( \{{\boldsymbol{x}_{2,j}^-}\}, \{{\boldsymbol{x}_{3,j}^-}\}, \boldsymbol{z}_1^-, \boldsymbol{z}_2^-, \boldsymbol{z}_3^-) \!-\! \varepsilon_{\rm{out}},0 \}]^2.
\end{array}
\end{equation}

According to \cite{shen2024principled}, let $C=2\max |f_{1,j}|$, we can obtain that,
\begin{equation}
\label{eq:5_14_96}
\begin{array}{l}  
\sum\limits_{j = 1}^N \! { \phi_j ||\boldsymbol{x}_{1,j}^* \!-\! \boldsymbol{z}_1^* ||^2}  \!+ \! \sum_{l} \! \lambda_l [\max\{h_l^{\rm{out}}( \{{\boldsymbol{x}_{2,j}^*}\}, \{{\boldsymbol{x}_{3,j}^*}\}, \boldsymbol{z}_1^*, \boldsymbol{z}_2^*, \boldsymbol{z}_3^*) \!-\! \varepsilon_{\rm{out}},0 \}]^2 
 \\
   
 \le  \sum\limits_{j = 1}^N  {{f_{1,j}}({\boldsymbol{x}_{1,j}^-},{\boldsymbol{x}_{2,j}^-},{\boldsymbol{x}_{3,j}^-}) -\sum\limits_{j = 1}^N {f_{1,j}}({\boldsymbol{x}_{1,j}^*},{\boldsymbol{x}_{2,j}^*},{\boldsymbol{x}_{3,j}^*})}  
  
  \vspace{1mm} \\
  
  \le NC.
 
\end{array}
\end{equation}

Because of $||\boldsymbol{x}_{1,j}^* \!-\! \boldsymbol{z}_1^* ||^2 \ge 0$ and $[\max\{h_l^{\rm{out}}( \{{\boldsymbol{x}_{2,j}^*}\}, \{{\boldsymbol{x}_{3,j}^*}\}, \boldsymbol{z}_1^*, \boldsymbol{z}_2^*, \boldsymbol{z}_3^*) \!-\! \varepsilon_{\rm{out}},0 \}]^2 \ge 0, \forall l$ and according to Eq. (\ref{eq:5_14_96}), we can obtain that,
\begin{equation}
\label{eq:5_14_97}
    ||\boldsymbol{x}_{1,j}^* \!-\! \boldsymbol{z}_1^* ||^2 \le \frac{NC}{\phi_j }, \forall j, 
\end{equation}
\begin{equation}
\label{eq:5_14_98}
    h_l^{\rm{out}}( \{{\boldsymbol{x}_{2,j}^*}\}, \{{\boldsymbol{x}_{3,j}^*}\}, \boldsymbol{z}_1^*, \boldsymbol{z}_2^*, \boldsymbol{z}_3^*) - \varepsilon_{\rm{out}} \le \sqrt{\frac{NC}{ \lambda_l}}, \forall l. 
\end{equation}

According to Eq. (\ref{eq:5_14_97}) and Eq. (\ref{eq:5_14_98}), we can conclude that the optimal solution $(\{{\boldsymbol{x}_{1,j}^*}\},\! \{{\boldsymbol{x}_{2,j}^*}\},\! \{{\boldsymbol{x}_{3,j}^*}\},\! \boldsymbol{z}_1^*,\! \boldsymbol{z}_2^*,\! \boldsymbol{z}_3^*)$ to P2 is a feasible solution to the relaxed problem of the original constrained problem P1, that is,
\begin{equation}
\label{eq:5_14_99}
    \begin{array}{l}
\min \sum\limits_{j = 1}^N {{f_{1,j}}({\boldsymbol{x}_{1,j}},{\boldsymbol{x}_{2,j}},{\boldsymbol{x}_{3,j}})} \\

{\rm{s}}{\rm{.t}}{\rm{.}}\; ||{\boldsymbol{x}_{1,j}} - \boldsymbol{z}_1||^2 \le \frac{NC}{\phi_j }, \forall j =1,\cdots,N 

\\  h_l^{\rm{out}}( \{{\boldsymbol{x}_{2,j}^*}\}, \{{\boldsymbol{x}_{3,j}^*}\}, \boldsymbol{z}_1^*, \boldsymbol{z}_2^*, \boldsymbol{z}_3^*)   \le \varepsilon_{\rm{out}} + \sqrt{\frac{NC}{ \lambda_l}}, \forall l

\\ {\mathop{\rm var}} .\qquad  \{{\boldsymbol{x}_{1,j}}\},\{{\boldsymbol{x}_{2,j}}\},\{{\boldsymbol{x}_{3,j}}\}, \boldsymbol{z}_1, \boldsymbol{z}_2, \boldsymbol{z}_3.

\end{array}
\end{equation}

Let $(\{{\overline{\boldsymbol{x}}_{1,j}}\},\! \{{\overline{\boldsymbol{x}}_{2,j}}\},\! \{{\overline{\boldsymbol{x}}_{3,j}}\},\! \overline{\boldsymbol{z}}_1,\! \overline{\boldsymbol{z}}_2,\! \overline{\boldsymbol{z}}_3)$ and $(\{{\underline{\boldsymbol{x}}_{1,j}}\},\! \{{\underline{\boldsymbol{x}}_{2,j}}\},\! \{{\underline{\boldsymbol{x}}_{3,j}}\},\! \underline{\boldsymbol{z}}_1,\! \underline{\boldsymbol{z}}_2,\! \underline{\boldsymbol{z}}_3)$ respectively denote the optimal solutions to P1 and the relaxed problem of P1 (i.e., Eq. (\ref{eq:5_14_99})), and let gap
\begin{equation}
\label{eq:5_14_100}
    \beta(\{\phi_j\},\{\lambda_l\}) = \sum\limits_{j = 1}^N {{f_{1,j}}(\{{\overline{\boldsymbol{x}}_{1,j}}\},\! \{{\overline{\boldsymbol{x}}_{2,j}}\},\! \{{\overline{\boldsymbol{x}}_{3,j}}\})} 
 - \sum\limits_{j = 1}^N {{f_{1,j}}(\{{\underline{\boldsymbol{x}}_{1,j}}\},\! \{{\underline{\boldsymbol{x}}_{2,j}}\},\! \{{\underline{\boldsymbol{x}}_{3,j}}\})}. 
\end{equation}

It is seen from Eq. (\ref{eq:5_14_99}) that this relaxation will be tightened with penalty parameter $\phi_j, \lambda_l, \forall j, \forall l$ increased. Combining with Eq. (\ref{eq:5_14_100}), we can obtain that $\beta(\{\phi_j\},\{\lambda_l\}) \ge 0$ will decrease when $\phi_j, \lambda_l, \forall j, \forall l$ increase. Next, we will demonstrate the gap between the optimal objective value by utilizing the exterior penalty method ( i.e., $\sum_{j = 1}^N \! {f_{1,j}}({\boldsymbol{x}_{1,j}^*},{\boldsymbol{x}_{2,j}^*},{\boldsymbol{x}_{3,j}^*})$ in P2) and the optimal objective value in original problem P1 (i.e., $\sum_{j = 1}^N \!{f_{1,j}}(\{{\overline{\boldsymbol{x}}_{1,j}}\},\! \{{\overline{\boldsymbol{x}}_{2,j}}\},\! \{{\overline{\boldsymbol{x}}_{3,j}}\})$) will continuously decrease with $\phi_j, \lambda_l, \forall j, \forall l$ increased. 

Because $(\{{\overline{\boldsymbol{x}}_{1,j}}\},\! \{{\overline{\boldsymbol{x}}_{2,j}}\},\! \{{\overline{\boldsymbol{x}}_{3,j}}\},\! \overline{\boldsymbol{z}}_1,\! \overline{\boldsymbol{z}}_2,\! \overline{\boldsymbol{z}}_3)$ is also the feasible solution to P2, and according to $\sum_j \phi_j ||\overline{\boldsymbol{x}}_{1,j} -\overline{\boldsymbol{z}}_1 ||^2 = 0$,  $ \sum_l \! \lambda_l [\max\{h_l^{\rm{out}}( \{{\overline{\boldsymbol{x}}_{2,j}}\}, \{{\overline{\boldsymbol{x}}_{3,j}}\}, \overline{\boldsymbol{z}}_1, \overline{\boldsymbol{z}}_2, \overline{\boldsymbol{z}}_3) \!-\! \varepsilon_{\rm{out}},0 \}]^2 = 0$, we have that,
\begin{equation}
\label{eq:5_14_101}
\begin{array}{l}
   \sum\limits_{j = 1}^N \! {f_{1,j}}({\boldsymbol{x}_{1,j}^*},{\boldsymbol{x}_{2,j}^*},{\boldsymbol{x}_{3,j}^*}) -\sum\limits_{j = 1}^N \!{f_{1,j}}(\{{\overline{\boldsymbol{x}}_{1,j}}\},\! \{{\overline{\boldsymbol{x}}_{2,j}}\},\! \{{\overline{\boldsymbol{x}}_{3,j}}\}) 
  \vspace{1mm} \\ 
  
  \le - \sum\limits_{j = 1}^N    \phi_j ||\boldsymbol{x}_{1,j}^* \!-\! \boldsymbol{z}_1^* ||^2  - \sum_l \! \lambda_l [\max\{h_l^{\rm{out}}( \{{\boldsymbol{x}_{2,j}^*}\}, \{{\boldsymbol{x}_{3,j}^*}\}, \boldsymbol{z}_1^*, \boldsymbol{z}_2^*, \boldsymbol{z}_3^*) \!-\! \varepsilon_{\rm{out}},0 \}]^2

  \vspace{1mm} \\
   
  \le 0.
\end{array}
\end{equation}

According to  $(\{{\boldsymbol{x}_{1,j}^*}\},\! \{{\boldsymbol{x}_{2,j}^*}\},\! \{{\boldsymbol{x}_{3,j}^*}\},\! \boldsymbol{z}_1^*,\! \boldsymbol{z}_2^*,\! \boldsymbol{z}_3^*)$ is a feasible solution to problem in Eq. (\ref{eq:5_14_99}), we can obtain that,
\begin{equation}
\label{eq:5_14_102}
    \sum\limits_{j = 1}^N \! {f_{1,j}}({\boldsymbol{x}_{1,j}^*},{\boldsymbol{x}_{2,j}^*},{\boldsymbol{x}_{3,j}^*}) \ge \sum\limits_{j = 1}^N \! {f_{1,j}}(\{{\underline{\boldsymbol{x}}_{1,j}}\},\! \{{\underline{\boldsymbol{x}}_{2,j}}\},\! \{{\underline{\boldsymbol{x}}_{3,j}}\}).
\end{equation}

By combining Eq. (\ref{eq:5_14_102}) with Eq. (\ref{eq:5_14_100}), we can obtain that,
\begin{equation}
\label{eq:5_14_103}
\begin{array}{l}
\sum\limits_{j = 1}^N \!{f_{1,j}}(\{{\overline{\boldsymbol{x}}_{1,j}}\},\! \{{\overline{\boldsymbol{x}}_{2,j}}\},\! \{{\overline{\boldsymbol{x}}_{3,j}}\}) - \sum\limits_{j = 1}^N \! {f_{1,j}}({\boldsymbol{x}_{1,j}^*},{\boldsymbol{x}_{2,j}^*},{\boldsymbol{x}_{3,j}^*})
  \vspace{1mm} \\ 

  \le \sum\limits_{j = 1}^N \!{f_{1,j}}(\{{\overline{\boldsymbol{x}}_{1,j}}\},\! \{{\overline{\boldsymbol{x}}_{2,j}}\},\! \{{\overline{\boldsymbol{x}}_{3,j}}\}) - \sum\limits_{j = 1}^N \! {f_{1,j}}(\{{\underline{\boldsymbol{x}}_{1,j}}\},\! \{{\underline{\boldsymbol{x}}_{2,j}}\},\! \{{\underline{\boldsymbol{x}}_{3,j}}\})

    \vspace{1mm} \\ 

    = \beta(\{\phi_j\},\{\lambda_l\}).
\end{array}
\end{equation}

By combining Eq. (\ref{eq:5_14_103}) with Eq. (\ref{eq:5_14_101}), we can obtain that,
\begin{equation}
\label{eq:5_14_104_1}
 -\beta(\{\phi_j\},\{\lambda_l\}) \le  \sum\limits_{j = 1}^N \! {f_{1,j}}({\boldsymbol{x}_{1,j}^*},{\boldsymbol{x}_{2,j}^*},{\boldsymbol{x}_{3,j}^*}) -\sum\limits_{j = 1}^N \!{f_{1,j}}(\{{\overline{\boldsymbol{x}}_{1,j}}\},\! \{{\overline{\boldsymbol{x}}_{2,j}}\},\! \{{\overline{\boldsymbol{x}}_{3,j}}\}) \le 0.
\end{equation}

Based on Eq. (\ref{eq:5_14_104_1}) and $ \beta(\{\phi_j\},\{\lambda_l\}) \ge 0$, we can get that, 
\begin{equation}
\label{eq:5_14_107}
  | \sum\limits_{j = 1}^N \! {f_{1,j}}({\boldsymbol{x}_{1,j}^*},{\boldsymbol{x}_{2,j}^*},{\boldsymbol{x}_{3,j}^*}) -\sum\limits_{j = 1}^N \!{f_{1,j}}(\{{\overline{\boldsymbol{x}}_{1,j}}\},\! \{{\overline{\boldsymbol{x}}_{2,j}}\},\! \{{\overline{\boldsymbol{x}}_{3,j}}\}) | \le \beta(\{\phi_j\},\{\lambda_l\}).
\end{equation}

By combining Eq. (\ref{eq:5_14_107}) with Eq. (\ref{eq:5_14_99}) and Eq. (\ref{eq:5_14_100}), we can conclude the gap between the optimal objective value by utilizing the exterior penalty method (i.e., $\sum_{j = 1}^N \! {f_{1,j}}({\boldsymbol{x}_{1,j}^*},{\boldsymbol{x}_{2,j}^*},{\boldsymbol{x}_{3,j}^*})$ in P2) and the optimal objective value in original problem P1 (i.e., $\sum_{j = 1}^N \!{f_{1,j}}(\{{\overline{\boldsymbol{x}}_{1,j}}\},\! \{{\overline{\boldsymbol{x}}_{2,j}}\},\! \{{\overline{\boldsymbol{x}}_{3,j}}\})$) is bounded and will decrease with penalty parameter $\phi_j, \lambda_l, \forall j, \forall l$ increased.

\renewcommand\arraystretch{1.5}
\renewcommand\tabcolsep{10pt}
\begin{table*}[t]
\centering
\renewcommand{\thetable}{\arabic{table}}
\caption{Comparisons between the proposed DTZO with the state-of-the-art TLL methods based on the applicability to different TLL problems. $\checkmark$ represents that the method can be applied to this TLL problem. The proposed DTZO is versatile and can be adapted to a wide range of TLL problems. We use ZOC as an abbreviation for zeroth order constraints.}
{
\scalebox{1.}{
\begin{tabular}{l|ccc|c}
\toprule
    & Betty & Hypergradient & AFTO & \textbf{DTZO} \\ \hline
 
Non-distributed TLL without ZOC  & \checkmark & \checkmark & \checkmark & \checkmark  \\

Distributed TLL without ZOC &  &  & \checkmark & \checkmark \\

TLL with partial ZOC   &  & & & \checkmark\\

TLL with level-wise ZOC     &  & & & \checkmark
  \\
\bottomrule  
\end{tabular}}
\label{tab:TLL difference}}
\end{table*}

\section{TLL with Partial Zeroth Order Constraints}
\label{appendix:partial zeroth order}
In this work, TLL with \textit{level-wise} zeroth order constraints is considered, where first order information at \textit{each level} is unavailable. In addition, it is worth mentioning that the proposed framework is versatile and can be adapted to a wide range of TLL problems with partial zeroth order constraints, i.e., grey-box TLL, through slight adjustments. The reason we refer to it as grey-box TLL is that the first order information for some levels in TLL is available, while for others it is not \cite{huang2024enhancing,beykal2020domino,astudillo2021thinking,bajaj2018trust}. To further show the superiority of the proposed DTZO, we compare it with the state-of-the-art TLL methods (i.e., Betty \cite{choe2022betty}, Hypergradient based method \cite{sato2021gradient}, and AFTO \cite{jiao2024provably}) based on their applicability to TLL problems in Table \ref{tab:TLL difference}. In DTZO, the zeroth order cut takes center stage, driving the construction of cascaded polynomial approximations without the need for gradients or sub-gradients. Notably, zeroth order cut is not only the backbone of DTZO but also opens the door to tackling grey-box TLL problems, seamlessly handling nested functions that combine both black-box and white-box elements. Discussions are provided as follows.

\subsection{TLL with second and third-level zeroth order constraints}

In this situation, the first order information at the first-level in TLL problems is accessible. Thus, we can use the exact gradients to replace the zeroth order gradient estimator, i.e., Eq. (\ref{eq:5_6_15})-(\ref{eq:5_6_18}) can be replaced by,
\begin{equation}
\label{eq:10_1_117}
  {\boldsymbol{x}_{1,j}^{t+1}} = {\boldsymbol{x}_{1,j}^{t}} - \eta_{\boldsymbol{x}_1}\left( \nabla_{\boldsymbol{x}_{1,j}} {f_{1,j}}({\boldsymbol{x}_{1,j}^t},{\boldsymbol{x}_{2,j}^t},{\boldsymbol{x}_{3,j}^t}) +  2\phi_j (\boldsymbol{x}_{1,j}^t - \boldsymbol{z}_1^t)\right),
\end{equation}
\begin{equation}
\label{eq:10_1_118}
  {\boldsymbol{x}_{2,j}^{t+1}} = {\boldsymbol{x}_{2,j}^{t}} - \eta_{\boldsymbol{x}_2}  \nabla_{\boldsymbol{x}_{2,j}} {f_{1,j}}({\boldsymbol{x}_{1,j}^t},{\boldsymbol{x}_{2,j}^t},{\boldsymbol{x}_{3,j}^t})-\eta_{\boldsymbol{x}_2} \nabla_{\boldsymbol{x}_{2,j}}  o( \{{\boldsymbol{x}_{2,j}^t}\},\! \{{\boldsymbol{x}_{3,j}^t}\},\! \boldsymbol{z}_1^t, \boldsymbol{z}_2^t, \boldsymbol{z}_3^t)  ,
\end{equation}
\begin{equation}
\label{eq:10_1_119}
  {\boldsymbol{x}_{3,j}^{t+1}} = {\boldsymbol{x}_{3,j}^{t}} - \eta_{\boldsymbol{x}_3}  \nabla_{\boldsymbol{x}_{3,j}} {f_{1,j}}({\boldsymbol{x}_{1,j}^t},{\boldsymbol{x}_{2,j}^t},{\boldsymbol{x}_{3,j}^t})-\eta_{\boldsymbol{x}_3} \nabla_{\boldsymbol{x}_{3,j}}  o( \{{\boldsymbol{x}_{2,j}^t}\},\! \{{\boldsymbol{x}_{3,j}^t}\},\! \boldsymbol{z}_1^t, \boldsymbol{z}_2^t, \boldsymbol{z}_3^t) .
\end{equation}

By using the gradient descent steps in Eq. (\ref{eq:10_1_117})-(\ref{eq:10_1_119}), the TLL problems with second and third-level zeroth order constraints can be effectively by the proposed framework.

\subsection{TLL with first and third-level zeroth order constraints}

In this situation, the first order information at the second-level in TLL problems is available. Thus, we can use the first order information to generate outer layer cutting plane, e.g., $\rho$-cut \cite{jiao2024provably}. By combining the outer layer first order cutting plane with the inner layer zeroth order cut, the proposed framework is capable of constructing the cascaded polynomial approximation. The generated outer layer $\rho$-cut can be expressed as,
\begin{equation}
\label{eq:10_1_120}
\begin{array}{l}
     \nabla \phi_{\rm{out}}{(\{{\boldsymbol{x}_{2,j}^t}\},\!\{{\boldsymbol{x}_{3,j}^t}\},\! \boldsymbol{z}_1^t, \boldsymbol{z}_2^t, \boldsymbol{z}_3^t)^{\top}}\left( {\left[ \begin{array}{l}
\!\{{\boldsymbol{x}_{2,j}}\}\\
\!\{{\boldsymbol{x}_{3,j}}\}\\
\boldsymbol{z}_1\\
\boldsymbol{z}_2\\
\boldsymbol{z}_3
\end{array} \right] \!-\! \left[ \begin{array}{l}
\!\{{\boldsymbol{x}_{2,j}^t}\}\\
\!\{{\boldsymbol{x}_{3,j}^t}\}\\
\boldsymbol{z}_1^t\\
\boldsymbol{z}_2^t\\
\boldsymbol{z}_3^t
\end{array} \right]} \right) \\
 + 
 \phi_{\rm{out}}(\{{\boldsymbol{x}_{2,j}^t}\},\! \{{\boldsymbol{x}_{3,j}^t}\},\! \boldsymbol{z}_1^t, \boldsymbol{z}_2^t, \boldsymbol{z}_3^t) \vspace{1mm} \\

\le \varepsilon_{\rm{out}} \!+\!  \rho\left(a_1 + (N+1)(a_2 + a_3)  + \sum_{i=2}^3 \sum_{j=1}^N ||{\boldsymbol{x}_{i,j}^t}||^2 + \sum_{i=1}^3 ||\boldsymbol{z}_i^t||^2  \right).
\end{array}
\end{equation}

In Eq. (\ref{eq:10_1_120}), $\rho>0$ is a parameter in $\rho$-weakly convex function, and $a_i, i=1,2,3$ is the boundness of variable $\boldsymbol{x}_{i,j}, \boldsymbol{z}_i$, as discussed in \cite{jiao2024provably}. By using the outer layer first order cutting plane, the TLL problems with first and third-level zeroth order constraints can be addressed by the proposed framework.

\subsection{TLL with first and second-level zeroth order constraints}

In this situation, the first order information at the third-level in TLL problems is accessible. Similarly, we can utilize the first order information to generate the inner layer cutting plane, e.g., $\rho$-cut. Through combining the inner layer first order cutting plane with the outer layer zeroth order cut, the proposed framework is capable of constructing the cascaded polynomial approximation. The generated inner layer $\rho$-cut can be expressed as,
\begin{equation}
\label{eq:10_1_121}
\begin{array}{l}
      \nabla \phi_{\rm{in}}{( \{{\boldsymbol{x}_{3,j}^t}\},\! \boldsymbol{z}_1^t, \boldsymbol{z}_2^t, \boldsymbol{z}_3^t)^{\top}}\left( {\left[ \begin{array}{l}
\{{\boldsymbol{x}_{3,j}}\}\\
\boldsymbol{z}_1\\
\boldsymbol{z}_2\\
\boldsymbol{z}_3
\end{array} \right] - \left[ \begin{array}{l}
\{{\boldsymbol{x}_{3,j}^t}\}\\
\boldsymbol{z}_1^t\\
\boldsymbol{z}_2^t\\
\boldsymbol{z}_3^t
\end{array} \right]} \right) 

+ \phi_{\rm{in}}( \{{\boldsymbol{x}_{3,j}^t}\},\! \boldsymbol{z}_1^t, \boldsymbol{z}_2^t, \boldsymbol{z}_3^t)

\vspace{1mm} \\

\le \varepsilon_{\rm{in}} + \rho \left(  (N+1)a_1  + a_2 + a_3 + \sum_{j=1}^N || \boldsymbol{x}_{3,j}^t||^2 + \sum_{i=1}^3 ||\boldsymbol{z}_i^t||^2 \right).
\end{array}
\end{equation}

By using the inner layer first order cutting plane in Eq. (\ref{eq:10_1_121}), the TLL problems with second and third-level zeroth order constraints can be addressed by the proposed framework.

{{\section{Discussions}
\label{appendix:discussion}
\subsection{Cutting Plane Method}

Cutting plane method, also called polyhedral approximation \cite{bertsekas2015convex}, is widely used in convex optimization \cite{franc2011cutting,boyd2007localization} and distributed optimization \cite{burger2013polyhedral,yang2014distributed}. The rationale behind cutting plane method is to use the intersection of a finite number of half-spaces (e.g., $P=\{x|a_l^Tx\le b_l, l=1,\cdots,L\}$, where $\{x|a_l^Tx\le b_l\}$ represent a half-space \cite{boyd2004convex}) to approximate the feasible region of the original optimization problem (e.g., $x \in \mathcal{X}$) . The approximation can be gradually refined by generating additional half-spaces \cite{bertsekas2015convex}. Recently, cutting plane methods have proven effective in tackling distributed multilevel optimization problems. By leveraging these methods, such problems can be transformed into decomposable optimization problems, which greatly simplifies the design of distributed algorithms for nested optimization, as discussed in \cite{jiao2022asynchronous,jiao2024provably}.  In \cite{jiao2022asynchronous}, cutting plane methods are applied to solve bilevel optimization problems within a distributed framework. Likewise, \cite{chen2024robust} utilize the cutting plane method to tackle distributed bilevel optimization challenges in downlink multi-cell systems. Building on this, \cite{jiao2024provably} further extend the approach to address distributed trilevel optimization problems. However, existing cutting plane methods for multilevel optimization rely on the first-order information to generate cutting planes, which are not available in zeroth-order optimization. In this work, we propose a framework capable of generating zeroth-order cuts for multilevel optimization problems $\textbf{without}$ the use of first-order information.

\subsection{The Choice of Gradient Estimator}
It is worth noting that the proposed framework is versatile, allowing for the integration of various gradient estimators. For instance, the mini-batch sampling-based gradient estimator \cite{liu2020primer,duchi2015optimal} can be employed to replace the two-point gradient estimator, reducing variance. Specifically, with mini-batch sampling, Eq. (\ref{eq:5_6_9}), (\ref{eq:5_6_11}) (\ref{eq:5_6_18}), (\ref{eq:5_6_19}), and (\ref{eq:5_6_20}) can be replaced by the following multi-point gradient estimators.
\begin{equation}
\begin{array}{l}
     {G_\mu^{\rm{in}} }( \{{\boldsymbol{x}_{3,j}^t}\}, \boldsymbol{z}_1^t, {\boldsymbol{z}_2^t}', \boldsymbol{z}_3^t) \\
  \!   = \! \frac{1}{{\mu}} \sum\limits_{p=1}^b  [\phi_{\rm{in}}( \{{\boldsymbol{x}_{3,j}^t} \!+ \! \mu \boldsymbol{\mu}_{x_{3,j}}^p\}, \boldsymbol{z}_1^t \!+\! \mu \boldsymbol{\mu}_{z_1}^p, {\boldsymbol{z}_2^t}' \!+\! \mu \boldsymbol{\mu}_{z_2}^p, \boldsymbol{z}_3^t \!+\! \mu \boldsymbol{\mu}_{z_3}^p)   -    \phi_{\rm{in}}( \{{\boldsymbol{x}_{3,j}^t}\}, \boldsymbol{z}_1^t, {\boldsymbol{z}_2^t}', \boldsymbol{z}_3^t)  \boldsymbol{\mu}^{{\rm{in}},p}],
\end{array}
\end{equation}
\begin{equation}
\begin{array}{l}
    {G_\mu^{\rm{out}} }(\{{\boldsymbol{x}_{2,j}^t}\},\!\{{\boldsymbol{x}_{3,j}^t}\},\! \boldsymbol{z}_1^t, \boldsymbol{z}_2^t, \boldsymbol{z}_3^t) \\
    = \frac{1}{\mu} \sum\limits_{p=1}^b  [ \phi_{\rm{out}}(\{{\boldsymbol{x}_{2,j}^t} + \mu \boldsymbol{\mu}_{x_{2,j}}^p\}, \{{\boldsymbol{x}_{3,j}^t} + \mu \boldsymbol{\mu}_{x_{3,j}}^p\}, \boldsymbol{z}_1^t + \mu \boldsymbol{\mu}_{z_1}^p, \boldsymbol{z}_2^t + \mu \boldsymbol{\mu}_{z_2}^p, \boldsymbol{z}_3^t + \mu \boldsymbol{\mu}_{z_3}^p) \\
    \qquad \qquad -   \phi_{\rm{out}}(\{{\boldsymbol{x}_{2,j}^t}\}, \{{\boldsymbol{x}_{3,j}^t}\}, \boldsymbol{z}_1^t, \boldsymbol{z}_2^t, \boldsymbol{z}_3^t)\boldsymbol{\mu}^{{\rm{out}},p}],
\end{array}
\end{equation}
\begin{equation}
\begin{array}{l}
  G_{\boldsymbol{x}_{1,j}}(\{{\boldsymbol{x}_{1,j}^t}\}, \{{\boldsymbol{x}_{2,j}^t}\}, \{{\boldsymbol{x}_{3,j}^t}\}, \boldsymbol{z}_1^t, \boldsymbol{z}_2^t, \boldsymbol{z}_3^t)
   \\   =\frac{1}{{\mu }} \sum\limits_{p=1}^b  [{{{f_{1,j}}({\boldsymbol{x}_{1,j}^t} + \mu {\boldsymbol{u}_{k,1}^p},{\boldsymbol{x}_{2,j}^t},{\boldsymbol{x}_{3,j}^t}) -{f_{1,j}}({\boldsymbol{x}_{1,j}^t},{\boldsymbol{x}_{2,j}^t},{\boldsymbol{x}_{3,j}^t})}} {\boldsymbol{u}_{k,1}^p}]+  2\phi_j (\boldsymbol{x}_{1,j}^t - \boldsymbol{z}_1^t),
\end{array}
\end{equation}
\begin{equation}
 \begin{array}{l}
  G_{\boldsymbol{x}_{2,j}}(\{{\boldsymbol{x}_{1,j}^t}\}, \{{\boldsymbol{x}_{2,j}^t}\}, \{{\boldsymbol{x}_{3,j}^t}\}, \boldsymbol{z}_1^t, \boldsymbol{z}_2^t, \boldsymbol{z}_3^t) = \nabla_{\boldsymbol{x}_{2,j}}  o( \{{\boldsymbol{x}_{2,j}^t}\},\! \{{\boldsymbol{x}_{3,j}^t}\},\! \boldsymbol{z}_1^t, \boldsymbol{z}_2^t, \boldsymbol{z}_3^t) \\
   \qquad \qquad \qquad \qquad +  \frac{1}{{\mu }} \sum\limits_{p=1}^b  [{{{f_{1,j}}({\boldsymbol{x}_{1,j}^t},{\boldsymbol{x}_{2,j}^t} + \mu {\boldsymbol{u}_{k,2}^p},{\boldsymbol{x}_{3,j}^t}) -{f_{1,j}}({\boldsymbol{x}_{1,j}^t},{\boldsymbol{x}_{2,j}^t},{\boldsymbol{x}_{3,j}^t})}} {\boldsymbol{u}_{k,2}^p}],
\end{array}
\end{equation}
\begin{equation}
\begin{array}{l}
  G_{\boldsymbol{x}_{3,j}}(\{{\boldsymbol{x}_{1,j}^t}\}, \{{\boldsymbol{x}_{2,j}^t}\}, \{{\boldsymbol{x}_{3,j}^t}\}, \boldsymbol{z}_1^t, \boldsymbol{z}_2^t, \boldsymbol{z}_3^t) =  \nabla_{\boldsymbol{x}_{3,j}} o( \{{\boldsymbol{x}_{2,j}^t}\},\! \{{\boldsymbol{x}_{3,j}^t}\},\! \boldsymbol{z}_1^t, \boldsymbol{z}_2^t, \boldsymbol{z}_3^t) \\
  \qquad \qquad \qquad \qquad  + \frac{1}{{\mu }} \sum\limits_{p=1}^b  [ {{{f_{1,j}}({\boldsymbol{x}_{1,j}^t},{\boldsymbol{x}_{2,j}^t},{\boldsymbol{x}_{3,j}^t} + \mu {\boldsymbol{u}_{k,3}^p}) -{f_{1,j}}({\boldsymbol{x}_{1,j}^t},{\boldsymbol{x}_{2,j}^t},{\boldsymbol{x}_{3,j}^t})}} {\boldsymbol{u}_{k,3}^p}],
\end{array}
\end{equation}
where $\boldsymbol{\mu}^{{\rm{in}},p}=[\{\boldsymbol{\mu}_{x_{3,j}}^p\}, \boldsymbol{\mu}_{z_1}^p, \boldsymbol{\mu}_{z_2}^p, \boldsymbol{\mu}_{z_3}^p]$, $\boldsymbol{\mu}^{{\rm{out}},p}=[\{\boldsymbol{\mu}_{x_{2,j}}^p\},\{\boldsymbol{\mu}_{x_{3,j}}^p\}, \boldsymbol{\mu}_{z_1}^p, \boldsymbol{\mu}_{z_2}^p, \boldsymbol{\mu}_{z_3}^p]$,  $\boldsymbol{u}_{k,1}^p$, $ \boldsymbol{u}_{k,2}^p$, $ \boldsymbol{u}_{k,3}^p, p=1,\cdots b$ are drawn from $\mathcal{N}(0, {\bf{I}})$, and $b$ represents the number of samples used in the multi-point gradient estimator.
}}

\section{Future Work}
\label{appendix:limitation}

This study is the first work that considers how to address the trilevel zeroth order optimization problems. The proposed framework is not only capable of addressing the single-level and bilevel zeroth order learning problems but can also be applied to a broad class of TLL problems, e.g., TLL with partial zeroth order constraints. However, higher-level nested learning problems, specifically those with more than three levels, are not considered in this work and will be addressed in future research.

\end{document}